# GPEX, A Framework For Interpreting Artificial Neural Networks

**Amir Akbarnejad**
Department of Computing Science
University of Alberta
Edmonton, AB, Canada
ah8@ualberta.ca

**Gilbert Bigras**
Department of Laboratory Medicine and Pathology
University of Alberta
Edmonton, AB, Canada
Gilbert.Bigras@albertaprecisionlabs.ca

**Nilanjan Ray**
Department of Computing Science
University of Alberta
Edmonton, AB, Canada
nray1@ualberta.ca

## Abstract

The analogy between Gaussian processes (GPs) and deep artificial neural networks (ANNs) has received a lot of interest, and has shown promise to unbox the blackbox of deep ANNs. Existing theoretical works put strict assumptions on the ANN (e.g. requiring all intermediate layers to be wide, or using specific activation functions). Accommodating those theoretical assumptions is hard in recent deep architectures, and those theoretical conditions need refinement as new deep architectures emerge. In this paper we derive an evidence lower-bound that encourages the GP's posterior to match the ANN's output without any requirement on the ANN. Using our method we find out that on 5 datasets, only a subset of those theoretical assumptions are sufficient. Indeed, in our experiments we used a normal ResNet-18 or feed-forward backbone with a single wide layer in the end. One limitation of training GPs is the lack of scalability with respect to the number of inducing points. We use novel computational techniques that allow us to train GPs with hundreds of thousands of inducing points and with GPU acceleration. As shown in our experiments, doing so has been essential to get a close match between the GPs and the ANNs on 5 datasets. We implement our method as a publicly available tool called GPEX: https://github.com/amirakbarnejad/gpex. On 5 datasets (4 image datasets, and 1 biological dataset) and ANNs with 2 types of functionality (classifier or attention-mechanism) we were able to find GPs whose outputs closely match those of the corresponding ANNs. After matching the GPs to the ANNs, we used the GPs' kernel functions to explain the ANNs' decisions. We provide more than 200 explanations (around 30 explanations in the paper and the rest in the supplementary) which are highly interpretable by humans and show the ability of the obtained GPs to unbox the ANNs' decisions.

## 1 Introduction

Artificial neural networks (ANNs) are widely adopted in machine learning. Despite their benefits, ANNs are known to be black-box to humans, meaning that their inner mechanism for making predictions is not necessarily interpretable/explainable to humans. ANN's black-box property impedes its deployment in safety-critical applications like medical imaging or autonomous driving, and makes them hard-to-troubleshoot for machine learning researchers.

Attribution-based explanation methods like LIME[31], SHAP[21] and most gradient-based explanation methods like DeepLIFT [5] presume a linear surrogate model. Given a test instance $\boldsymbol{x_{test}}$, this simpler surrogate model is encouraged to have the same output "locally" around $\boldsymbol{x_{test}}$. Because of this "local assumptions", explanations from these methods might be unreliable, and can be easily manipulated by an adversary model [11][28]. Moreover, these models may produce discordant explanations for a fixed model and test instance [16].

Considering Gaussian processes (GPs) [26] as the explainer model is beneficial, because: 1. Gaussian processes are highly interpretable. 2. Researchers have long known that GP's posterior has the potential to match an ANN's output "globally". More precisely, given an ANN and some requirements on it [24][8], there might exist a GP whose posterior matches the ANN's output all over the input-space $\mathbb{X}$ (as opposed to the local explanation models for which the match happens only locally around a test instance $\boldsymbol{x_{test}} \in \mathbb{X}$). Not many explainer models can globally match the ANN's output. Among gradient-based methods, with the best of our knowledge only Integrated Gradients [33] has a weak sense of ANN's global behaviour over the input space. Having some conditions on an ANN, representer point selection [37] finds a "globally faithfull" explainer model that, similar to GPs, works with a kernel function. As we will elaborate upon in Sec. 4.3 and Sec. S6 of the supplementary, the GP's kernel that we find in this paper is superior due to a technical point in the formulation of representer point selection [37]. All in all, using GPs to explain ANNs is quite promissing and has advantages over other approaches to explain ANNs.

The contributions of this paper are as follows:

- Theoretical results on ANN-GP analogy impose some restrictions on ANNs under which the ANN will be equivalent to a GP. These conditions are too restrictive for recently used deep architectures. Moreover, those theoretical conditions need refinement as new deep architectures emerge. In this paper we derive an ELBO for training GPs which encourages GP's posterior to match ANN's output. Our formulation and method doesn't impose any restriction on the ANN and the method used to train it.
- Using our method, we empirically show that on 5 datasets (4 image datasets, and 1 biological dataset) and ANNs with 2 types of functionality (classifier or attention-mechanism) the ANN needs to fulfill only a subset of those theoretical conditions. Indeed, in our experiments we used a normal ResNet-18 or feed-forward backbone with a single wide layer in the end.
- Scalability is a major issue in training GPs. To address this issue, we adopted computational techniques recently used for fast spectral clustering [13] as well as a novel method to learn the GPs using mini-batches of inducing points and training instances. These computational techniques allow us to train GPs with hundreds of thousands of inducing points. According to our analysis, increasing the inducing points has been essential to get a good match between the trained GPs and ANNs. Indeed, without many inducing points GPs posterior cannot be a complex function (a function with many ups and downs [36]) and fails to match ANNs' output.
- With the best of our knowledge, our work is the first method that performs knowledge distillation between GPs and ANNs.
- We implement our method as a public python library called GPEX (Gaussian Processes for EXplaining ANNs). GPEX takes in an arbitrary PyTorch module, and replaces any ANN submodule of choice by GPs. Our package makes use of GPU-accelaration, and enables effortless application of GPs without getting users involved in details of the inference procedure. GPEX can be used by machine learning researchers to interpret/troubleshoot their artificial neural networks. Moreover, GPEX can be used by researchers working on the theoretical side of ANN-GP analogy to empirically test their hypotheses.

# 2 Proposed Method

## 2.1 Notation

In this article the function $g(.)$ always denotes an ANN. The kernel of a Gaussian process is denoted by the double-input function $\mathcal{K}(.,.)$. We assume the kernel similarity between two instances $\boldsymbol{x_i}$ and $\boldsymbol{x_j}$ is equal to $f(\boldsymbol{x_i})^T f(\boldsymbol{x_j})$, where $f(.)$ maps the input-space to the kernel space. In this paper $\boldsymbol{u}$ (resp. $v$) denotes a vector in the kernel-space (resp. the posterior mean) of a GP. In some sense $\boldsymbol{u}$ and $v$ denote

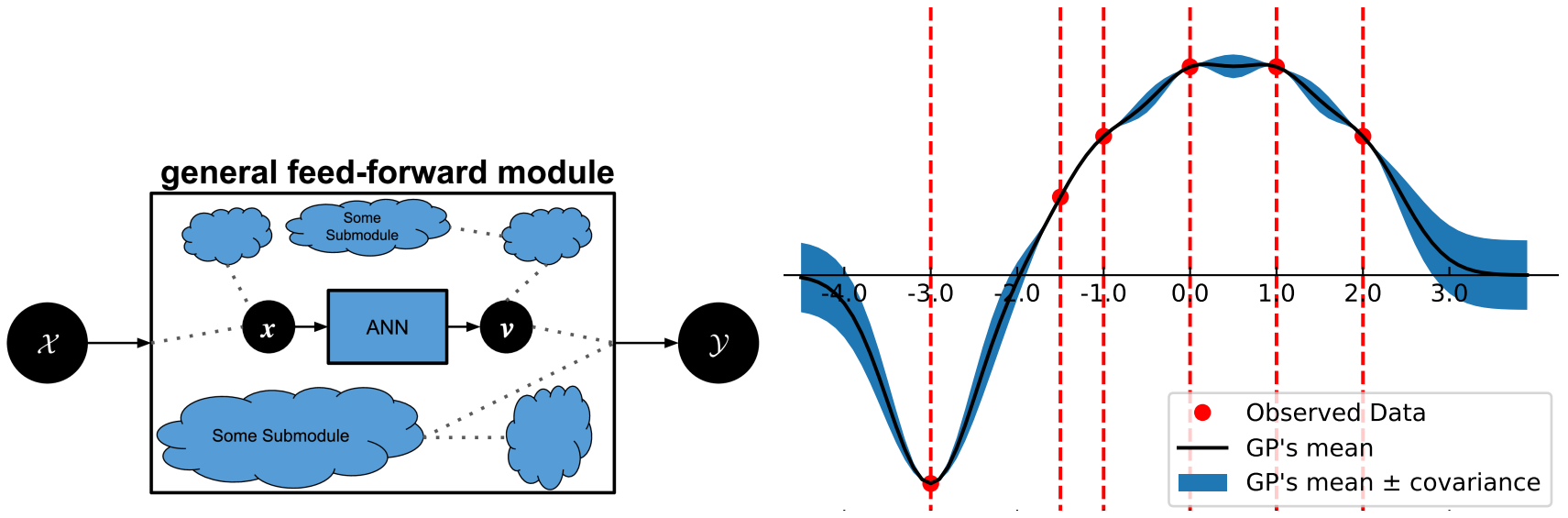

a b

Figure 1: a) A general feed-forward pipeline, with an ANN sub-module to be explained by GPEX. b) Typical behaviour of Guassian process posterior given a set of observed values.

the input and the output of a GP, respectively. We have that: $\mathcal{K}(\boldsymbol{x_i}, \boldsymbol{x_j}) = f(\boldsymbol{x_i})^T f(\boldsymbol{x_j}) = \boldsymbol{u_i}^T \boldsymbol{u_j}$. The number of GPs is equal to the number of the outputs from the ANN. In other words, we consider one GP per scalar output-head from the ANN. We use index $\ell$ to specify the $\ell$-th GP as follows: $\mathcal{K}_\ell(\boldsymbol{x_i}, \boldsymbol{x_j}) = f_\ell(\boldsymbol{x_i})^T f_\ell(\boldsymbol{x_j}) = \boldsymbol{u_i}^{(\ell)T} \boldsymbol{u_j}^{(\ell)}$. We parameterize the $\ell$-th GP by a set of $M$ inducing points $\{(\tilde{\boldsymbol{u}}_m^{(\ell)}, \tilde{v}_m^{(\ell)})\}_{m=1}^M$. The tilde in $(\tilde{\boldsymbol{u}}_m^{(\ell)}, \tilde{v}_m^{(\ell)})$ indicates that $\tilde{\boldsymbol{u}}$ is one of the $M$ inducing points in the kernel space. However, $\boldsymbol{u}$ (without tilde) can be an arbitrary point in the continuous kernel-space.

### 2.2 The Proposed Framework

To make our framework as general as possible, we consider a general feed-forward pipeline that contains an ANN as a submodule. In Fig. 1a the bigger square illustrates the general module. The input-output of the general pipeline are denoted in Fig. 1a by $\mathcal{X}$ and $\mathcal{Y}$. The general pipeline has at least one ANN submodule to be explained by GPEX. Fig. 1a illustrates this ANN by the small blue rectangle within the general pipeline. The input-output of the ANN are denoted in Fig. 1a by $\boldsymbol{x}$ and $\boldsymbol{v}$. Note that $\mathcal{X}$ and $\mathcal{Y}$ can be anything, including without any limitation, a set of vectors, labels, and meta-information. However, input-output of the ANN (i.e. $\boldsymbol{x}$ and $\boldsymbol{y}$) are required to be in tensor format. The exact requirements are provided in the online documentation for GPEX. Moreover, the general module can have other arbitrary submodules, which are depicted by the blue clouds. The relations between the submodules, as illustrated by the dotted-lines in Fig. 1a, can also be quite general. Our probabilistic formulation only needs access to the conditional distributions $p(\boldsymbol{x}|\mathcal{X})$ and $p(\mathcal{Y}|\boldsymbol{x}, \mathcal{X})$. Similarly, the proposed GPEX is completely agnostic about the general pipeline and it only requires the ANN's input-output to be in the tensor format. Given a PyTorch module, the proposed GPEX tool automatically grabs the distributions $p(\boldsymbol{x}|\mathcal{X})$ and $p(\mathcal{Y}|\boldsymbol{x}, \mathcal{X})$ from the main module it is given.

The inducing points $\{\tilde{\boldsymbol{u}}_m^{(\ell)}, \tilde{v}_m^{\ell}\}_{m=1}^M$ parameterize the $\ell$-th GP. Note that $\tilde{\boldsymbol{u}}_m^{(\ell)} = f_\ell(\tilde{\boldsymbol{x}}_m)$. A feature point like $\boldsymbol{x}$ is first mapped to the kernel-space as $\boldsymbol{u}^{(\ell)} = f_\ell(\boldsymbol{x})$. Note that the kernel functions $\{f_\ell(.)\}_{\ell=1}^L$ are implemented as separate neural networks, or for the sake of efficiency as a single neural network backbone with $L$ different heads. Afterwards, the GP's posterior on $\boldsymbol{x}$ depends on the kernel similarities between $\boldsymbol{u}^{(\ell)}$ and the inducing points $\{\tilde{\boldsymbol{u}}_m^{(\ell)}\}_{m=1}^M$. More precisely, the posterior of the $\ell$-th GP on $\boldsymbol{x}$ is a random variable $v^{(\ell)}$ whose distribution is as follows [26]:

$$p(v^{(\ell)}|\boldsymbol{u}^{(\ell)}, \tilde{\mathbf{u}}_{1:M}^{(\ell)}, \tilde{v}_{1:M}^{(\ell)}) = \mathcal{N}\Big(v^{(\ell)} \ ; \ \mu_v(\boldsymbol{u}^{(\ell)}, \tilde{\mathbf{u}}_{1:M}^{(\ell)}, \tilde{v}_{1:M}^{(\ell)}), \ cov_v(\boldsymbol{u}^{(\ell)}, \tilde{\mathbf{u}}_{1:M}^{(\ell)}, \tilde{v}_{1:M}^{(\ell)})\Big), \quad (1)$$

where $\mu_v(.,.,.)$ and $cov_v(.,.,.)$ are the mean and covariance of a GP's posterior computed as:

$$\mu_v(\boldsymbol{u}^{(\ell)}, \tilde{\mathbf{u}}_{1:M}^{(\ell)}, \tilde{v}_{1:M}^{(\ell)}) = \mathcal{K}(\boldsymbol{u}^{(\ell)}, \tilde{\mathbf{u}}_{1:M}^{(\ell)}) \big[\mathcal{K}(\tilde{\mathbf{u}}_{1:M}^{(\ell)}, \tilde{\mathbf{u}}_{1:M}^{(\ell)}) + \sigma_{gp}^2 \mathbf{I}_{M\times M}\big]^{-1} \tilde{v}_{1:M}^{(\ell)} \quad (2)$$

and

$$cov_v(\boldsymbol{u}^{(\ell)}, \tilde{\mathbf{u}}_{1:M}^{(\ell)}, \tilde{v}_{1:M}^{(\ell)}) = \mathcal{K}(\boldsymbol{u}^{(\ell)}, \boldsymbol{u}^{(\ell)}) - \mathcal{K}(\boldsymbol{u}^{(\ell)}, \tilde{\mathbf{u}}_{1:M}^{(\ell)}) \times \big[\mathcal{K}(\tilde{\mathbf{u}}_{1:M}^{(\ell)}, \tilde{\mathbf{u}}_{1:M}^{(\ell)}) + \sigma_{gp}^2 \mathbf{I}_{M\times M}\big]^{-1} \mathcal{K}(\tilde{\mathbf{u}}_{1:M}^{(\ell)}, \boldsymbol{u}^{(\ell)}). \quad (3)$$

As the variables $\{v_m^{(\ell)}\}_{m=1}^M$ and $v$ are latent or hidden, we train the model parameters by optimizing a variational lower-bound. We consider the following variational distributions:

$$q_1(v^{(\ell)} \ |\boldsymbol{x}) = \mathcal{N}(v^{(\ell)} \ ; \ g_\ell(\boldsymbol{x}) \ , \ \sigma_g^2), \qquad q_2(\tilde{v}_m^{(\ell)}) = \mathcal{N}(\tilde{v}_m^{(\ell)} \ ; \varphi_m^{(\ell)}, \sigma_\varphi^2). \quad (4)$$

In Eq. 4, the function $g_\ell(.)$ is the $\ell$-th output from the ANN. Note that as the set of hidden variables $\{\tilde{v}_m^{(\ell)}\}_{m=1}^M$ is finite, we have parameterized their variational distribution by a finite set of numbers $\{\varphi_m^{(\ell)}\}_{m=1}^M$. However, as the variables $\boldsymbol{x}$ can vary arbitrarily in the feature space, the variable $\boldsymbol{u}^{(\ell)}$ varies arbitrarily in the kernel space. Therefore, the set of values $v^{(\ell)}$ may be infinite. Accordingly, the variational distribution for $v^{(\ell)}$ is conditioned on $\boldsymbol{x}$ and is parameterized by the ANN $g(.)$.

### 2.3 The Derived Evidence Lower-Bound (ELBO)

Due to space limitation, the derivation of the lower-bound is moved to Sec. S1 of the supplementary material. In this section we only introduce the derived ELBO and discuss how it relates the GP, the ANN and the training cost of the main module in an intuitive way. The ELBO terms containing the GP parameters (i.e. the parameters of the kernel function $f(.)$) is denoted by $\mathcal{L}_{gp}$. According to Eq. S9 of the supplementary material $\mathcal{L}_{gp}$ is as follows:

$$\mathcal{L}_{gp} = -\frac{1}{2}\,\mathbb{E}_{\sim q}\Big[\sum_{\ell=1}^{L}\frac{(\mu_v(\boldsymbol{u}^{(\ell)},\tilde{\mathbf{u}}_{1:M}^{(\ell)},\tilde{v}_{1:M}^{(\ell)}) - g_\ell(\boldsymbol{x}))^2 + \sigma_g^2}{cov_v(\boldsymbol{u}^{(\ell)},\tilde{\mathbf{u}}_{1:M}^{(\ell)},\tilde{v}_{1:M}^{(\ell)})}\Big] - \frac{1}{2}\,\mathbb{E}_{\sim q}\Big[\sum_{\ell=1}^{L}\log\big(\frac{cov_v(\boldsymbol{u}^{(\ell)},\tilde{\mathbf{u}}_{1:M}^{(\ell)},\tilde{v}_{1:M}^{(\ell)})}{\sigma_g^2}\big)\Big] + \big(\text{const.}\big), \tag{5}$$

where $q(.)$ is the variational distribution that factorizes to the $q_1(.)$ and $q_2(.)$ distributions defined in Eq. 4. In the first term of Eq. 5, the numerator encourages the GP and the ANN to have the same output. More precisely, for a feature point $\boldsymbol{x}$ we can compute the corresponding point in the kernel space as $\boldsymbol{u}^{(\ell)} = f_\ell(\boldsymbol{x})$ and then compute the GP's posterior mean based on kernel similarities between $\boldsymbol{u}$ and the inducing points to get the GP's mean $\mu_v$. In Eq. 5 the GP's mean $\mu_v$ is encouraged to match the ANN's output $g_\ell(\boldsymbol{x})$. In Eq. 5, because of the denominator of the first term, the ANN-GP similarity is not encouraged uniformly over the feature-space. Wherever the GP's uncertainty is low, the term $cov_v(\boldsymbol{u}^{(\ell)},\tilde{\mathbf{u}}_{1:M}^{(\ell)},\tilde{v}_{1:M}^{(\ell)})$ in the denominator becomes small. Therefore, the GP's mean is highly encouraged to match the ANN's output. On the other hand, in regions where the GP's uncertainty is high, the GP-ANN analogy is less encouraged. This formulation is quite intuitive according to the behaviour of Gaussian processes. Fig. 1b illustrates the posterior of a GP with radial-basis kernel for a given set of observations. In regions like $[3,\infty)$ and $(-\infty,-4]$ there are no nearby observed data. Therefore, in these regions the GP is highly uncertain and the blue uncertainty margin is thick in such regions. Intuitively, our derived ELBO in Eq. 5 encourages the GP-ANN analogy only when GP's uncertainty is low and gives less importance to regions similar to $[3,\infty)$ and $(-\infty,-4]$ in Fig. 1b. Note that this formulation makes no difference for the ANN as ANNs are known to be global approximators. However, this formulation makes a difference when training the GP, because the GP is not required to match the ANN in regions where there are no similar training instances. The ELBO terms containing the ANN parameters is denoted by $\mathcal{L}_{ann}$. According to Sec. S1.2 of the supplementary material, $\mathcal{L}_{ann}$ is as follows:

$$\mathcal{L}_{ann} = -\frac{1}{2}\,\mathbb{E}_{\sim q}\Big[\sum_{\ell=1}^{L}\frac{(\mu_v(\boldsymbol{u}^{(\ell)},\tilde{\mathbf{u}}_{1:M}^{(\ell)},\tilde{v}_{1:M}^{(\ell)}) - g_\ell(\boldsymbol{x}))^2}{cov_v(\boldsymbol{u}^{(\ell)},\tilde{\mathbf{u}}_{1:M}^{(\ell)},\tilde{v}_{1:M}^{(\ell)})}\Big] + \mathbb{E}_{\sim q}\big[\log p(\mathcal{Y}|\boldsymbol{y},\mathcal{X})\big]. \tag{6}$$

In the above objective the first term encourages the ANN to have the same output as the GP. Similar to the objective of Eq. 5, the denominator of the first term gives more weight to ANN-GP analogy when GP's uncertainty is low. In the right-hand-side of Eq. 6, the second term is the likelihood of the pipeline's output(s), i.e. $\mathcal{Y}$ in Fig. 1a. This term can be, e.g., the cross-entropy loss when $\mathcal{Y}$ contains class scores in a classification problem, or the mean-squared error when $\mathcal{Y}$ is the predicted value for a regression problem, or a combination of those costs in a multi-task setting.

# 3 Algorithm

We consider a separate Gaussian process for each output head of an ANN. In other words, given an ANN we have as many GPs as the number of the ANN's output heads. To explain an ANN, we find the explainer GPs by optimizing the objective in Eq. 5 w.r.t. to the kernel mappings $\{f_\ell(.)\}_{\ell=1}^L$. To do so, we need to have $\mu_v$ which in turn means we need to have all kernel-space representations

**Algorithm 1** Method Optim_KernMappings

**Input:** Input instance $\boldsymbol{x}$ and inducing instance $\tilde{\boldsymbol{x}}$, list of matrices $\mathbf{U}$, list of vectors $\mathbf{V}$.
**Output:** Kernel-space mappings $[f_1(.), ..., f_L(.)]$.
*Initialisation* : $loss \leftarrow 0$.
1: $\mu$, $cov \leftarrow$ forward_GP($\mathbf{x}$, $\tilde{\boldsymbol{x}}$, $\mathbf{U}$, $\mathbf{V}$) //feed $\mathbf{x}$ to GPs, "forward_GP" is Alg.S1 in supplementary.
2: $\mu_{ann} \leftarrow g(\mathbf{x})$ //feed $\mathbf{x}$ to ANN.
3: **for** $\ell = 1$ to $L$ **do**
4: $\quad loss \leftarrow loss + \frac{(\mu[\ell]-\mu_{ann}[\ell])^2+\sigma_g^2}{cov[\ell]} + \log(cov[\ell])$. //Eq.5.
5: **end for**
6: $\boldsymbol{\delta} \leftarrow \frac{\partial\ loss}{\partial\ params\big([f_1(.), ..., f_L(.)]\big)}$.//the gradient of loss.
7: $params\big([f_1, ..., f_L]\big) \leftarrow params\big([f_1, ..., f_L]\big) - lr \times \boldsymbol{\delta}$ //update the parameters.
8: $lr \leftarrow$ updated learning rate
9: **return** $[f_1(.), ..., f_L(.)]$

**Algorithm 2** Method Explain_ANN

**Input:** Training dataset $ds_train$, and the inducing dataset $ds_inducing$.
**Output:** Updated kernel-space mappings $[f_1(.), ..., f_L(.)]$, and the other GP parameters $\mathbf{U}$ and $\mathbf{V}$.
*Initialisation* : $\mathbf{U}$, $\mathbf{V}$ $\leftarrow$ Init_GPparams(ds_inducing) //Alg.S3 in supplementary.
1: **for** $iter = 1$ to $max_iter$ **do**
2: $\quad \boldsymbol{x} \leftarrow randselect(ds_train)$.
3: $\quad \tilde{\boldsymbol{x}} \leftarrow randselect(ds_inducing)$
4: $\quad [f_1(.), ..., f_L(.)]$ $\leftarrow$ Optim_KernMapings($\boldsymbol{x}, \tilde{\boldsymbol{x}}, \mathbf{U}, \mathbf{V}$).
5: $\quad \tilde{\boldsymbol{x}} \leftarrow randselect(ds_inducing)$.
6: $\quad$ **for** $\ell = 1$ to $L$ **do**
7: $\quad\quad$ //update kernel-space representations.
8: $\quad\quad U[\ell][\tilde{\boldsymbol{x}}.index] \leftarrow f_\ell(\tilde{\boldsymbol{x}})$
9: $\quad$ **end for**
10: **end for**
11: **return** $[f_1(.), ..., f_L(.)]$, $\mathbf{U}$, $\mathbf{V}$

$\{\tilde{\boldsymbol{u}}_m^{(\ell)}\}_{m=1}^M$. However, it is computationally prohibitive to feed thousands of inducing instances to the kernel mappings as $\tilde{\boldsymbol{u}}_m^{(\ell)} = f_\ell(\tilde{\boldsymbol{x}}_{\boldsymbol{m}})$ for $m \in \{1, 2, ..., M\}$ in each gradient-descent iteration. On the other hand, as the kernel-space mappings $\{f_\ell(.)\}_{\ell=1}^L$ keep changing during training, we need to somehow track how the inducing points $\{\tilde{\boldsymbol{u}}_m^{(\ell)}\}_{m=1}^M$ change during training. To this end, we put the kernel-space representations of the inducing points in matrices denoted by $\mathbf{U}$. During training, these matrices are repeatedly updated by feeding mini-batches of inducing instances to the kernel-mappings.

Alg. 2 optimizes the objective of Eq. 5 w.r.t. the kernel mappings $\{f_\ell(.)\}_{\ell=1}^L$. First, a single training instance $\boldsymbol{x}$ and a single inducing point $\tilde{\boldsymbol{x}}$ are selected (line 2-3). Afterwards, the procedure of Alg. 1 is called to update the kernel mappings (line 4 of Alg. 2). To update the kernel-mappings, the GP posterior is computed via the matrices $\mathbf{U}$ (line 1 of Alg.1). The "forward_GP" procedure (called in line 1 of Alg. 1) is provided in Alg. S1 of the supplementary, and uses the matrices $\mathbf{U}$ to compute GP's posterior. Only the rows of $\mathbf{U}$ that correspond to the selected inducing point $\tilde{\boldsymbol{x}}$ are computed using the kernel-mappings, so that the gradient w.r.t. the kernel-mappings can be computed in the backward pass (lines 5-6 of Alg. S1 in the supplementary). Finally, the matrices $\mathbf{U}$ are updated (lines 6-8 of Alg. 2). Due to the lack of space, the routines "forward_GP" and "Init_GPparams" and more details are moved to Sec. S2 of the supplementary. Of course instead of a single training/inducing instance, we used a mini-batch of multiple training/inducing instances.

One difficulty of training GPs is the matrix inversion of Eqs.2 and 3, which has $\mathcal{O}(M^3)$ complexity using standard matrix inversion methods. To address this issue, we adopted computational techniques recently used for fast spectral clustering [13]. Let $\boldsymbol{A}$ be an arbitrary $M \times D$ matrix where $M >> D$. Moreover, let $\boldsymbol{b}$ be a $M$-dimensional vector and let $\sigma$ be a scalar. The computational techniques [13] allow us to efficiently compute: $(\mathbf{A}\mathbf{A}^T + \sigma^2 \boldsymbol{I}_{M\times M})^{-1}\ \boldsymbol{b}$. (Note the similar terms in the right hand side of Eqs. 2 and 3.) The idea is that $\mathbf{A}\mathbf{A}^T$ is of rank $D$. Therefore, from linear algebra it follows that $(\mathbf{A}\mathbf{A}^T + \sigma^2 \boldsymbol{I}_{M\times M})^{-1}$ has $M - D$ eigen-values all of which are equal to $\sigma^{-2}$. Therefore, in the space of those eigen-vectors, the transformation on $\boldsymbol{b}$ is simply a scaling by $\sigma^{-2}$. The details and more computational techniques are provided in Sec. S2.1 of the supplementary. These computational techniques allow us to efficiently compute the GP-posterior for hundreds of thousands of inducing points in each gradient descent iteration.

**A note on the used datasets in Alg. 2:** According to our analysis of Sec. S7 in the supplementary material, "ds_inducing" should be as large as possible so the GP posteriors can be flexible enough to match the ANNs. Therefore a good practice is to include all training instances (without data augmentation) in "ds_inducing". By doing so, the following issue arises. An instance from "ds_train" like $\boldsymbol{x}$ is an augmented version of an inducing instance $\tilde{\boldsymbol{x}}$. Because $\boldsymbol{x}$ and $\tilde{\boldsymbol{x}}$ are close, their kernel-space representations $f(\boldsymbol{x})$ and $f(\tilde{\boldsymbol{x}})$ also become close regardless of parameters of $f(.)$. Consequently, regardless of $f(.)$, GP's posterior mean will be roughly equal for both $\boldsymbol{x}$ and $\tilde{\boldsymbol{x}}$. Indeed, in this case Alg. 2 fails to find the kernel mappings $\{f_\ell(.)\}_{\ell=1}^{L}$. To avoid this issue, we sample $\boldsymbol{x}$ in line 2 of Alg. 2 as follows: $\boldsymbol{x}_1$ and $\boldsymbol{x}_2$ are randomly selected from "ds_train", and $\alpha \sim uniform(-1, 2)$, and $\boldsymbol{x} = \alpha\boldsymbol{x}_1 + (1-\alpha)\boldsymbol{x}_2$. The rest of Alg. 2 after line 2 is run as before.

# 4 Experiments

We conducted several experiments on four publicly available datasets: MNIST [9], Cifar10 [19], Kather [15], and DogsWolves [34]. For MNIST [9] and Cifar10 [19] we used the standard split to training and test sets provided by the datasets. For Kather [15] and DogsWolves [34] we randomly selected 70% and 80% of instances as our training set. The exact parameter settings for running Alg. 2 are elaborated upon in Sec. S5 of the supplementary. We trained the ANNs as usual rather than using Eq. 6, because our proposed GPEX should be applicable to ANNs which are trained as usual.

## 4.1 Measuring Faithfulness of GPs to ANNs

We trained a separate convolutional neural network (CNN) on each dataset to perform the classification task. For MNIST [9], Cifar10 [19], and Kather [15] we used a ResNet-18 [12] backbone followed by some fully connected layers. DogsWolves [34] is a relatively small dataset, and very deep architectures like ResNet [12] overfit to training set. Therefore, we used a convolutional backbone which is suggested in the dataset website [34]. For all datasets, we set the width (i.e. the number of neurons) of the second last fully-connected layer to 1024. Because according to theoretical results on GP-ANN analogy, the second last layer of ANN should be wide. We used an implementation of ResNet [12] which is publicly available online [2]. We trained the pipelines for 20, 200, 20, and 20 epochs on MNIST [9], Cifar10 [19], Kather [15], and DogsWolves [34], respectively. For Cifar10 [19], we used the exact optimizer suggested by [2]. For other datasets we used an Adam [17] optimizer with a learning-rate of 0.0001. The test accuracies of the models are equal to 99.56%, 95.43%, 96.80%, and 80.50% on MNIST [9], Cifar10 [19], Kather [15], and DogsWolves [34], respectively. We also applied our proposed GPEX to a state-of-the-art cell-embedding method called scArches [20]. We ran a tutorial notebook [32] and applied GPEX to the decoder whose job is to predict expression of some genes given scArches [20] cell embeddings. More details are provided in our public github repository (repository link is provided in page 1).

We explained each classifier ANN using our proposed GPEX framework (i.e. Alg.2). As discussed in Sec. 3, given an ANN we have as many kernel-spaces (and as many GPs) as the number of ANN's output heads. The exact parameter settings and practical considerations for training the GPs is elaborated upon in Sec. S5 of the supplementary material. To measure the faithfulness of GPs to ANNs, we compute the Pearson correlation coefficient for each ANN head and the mean of the corresponding GP posterior on unseen test instances. The results are provided in Fig. 3. In Fig. 3, the first five groups of bars (i.e. the groups labeled as Cifar10 (classifier), MNIST (classifier), Kather (classifier), DogsWolves (classifier), and scArches (classifier)) correspond to applying the proposed GPEX to the five classifier ANNs trained on the four datasets and scArches embeddings. According to Fig. 3, our trained GPs almost perfectly match the corresponding ANNs. Only for DogsWovles [34], as illustrated by the 4-th bar group in Fig. 3, the correlation coefficients are lower compared to other datasets. We hypothesize that this is because the DogsWolves dataset [34] has very few images. GP posterior mean can be changed only by moving the inducing points in the kernel-space.

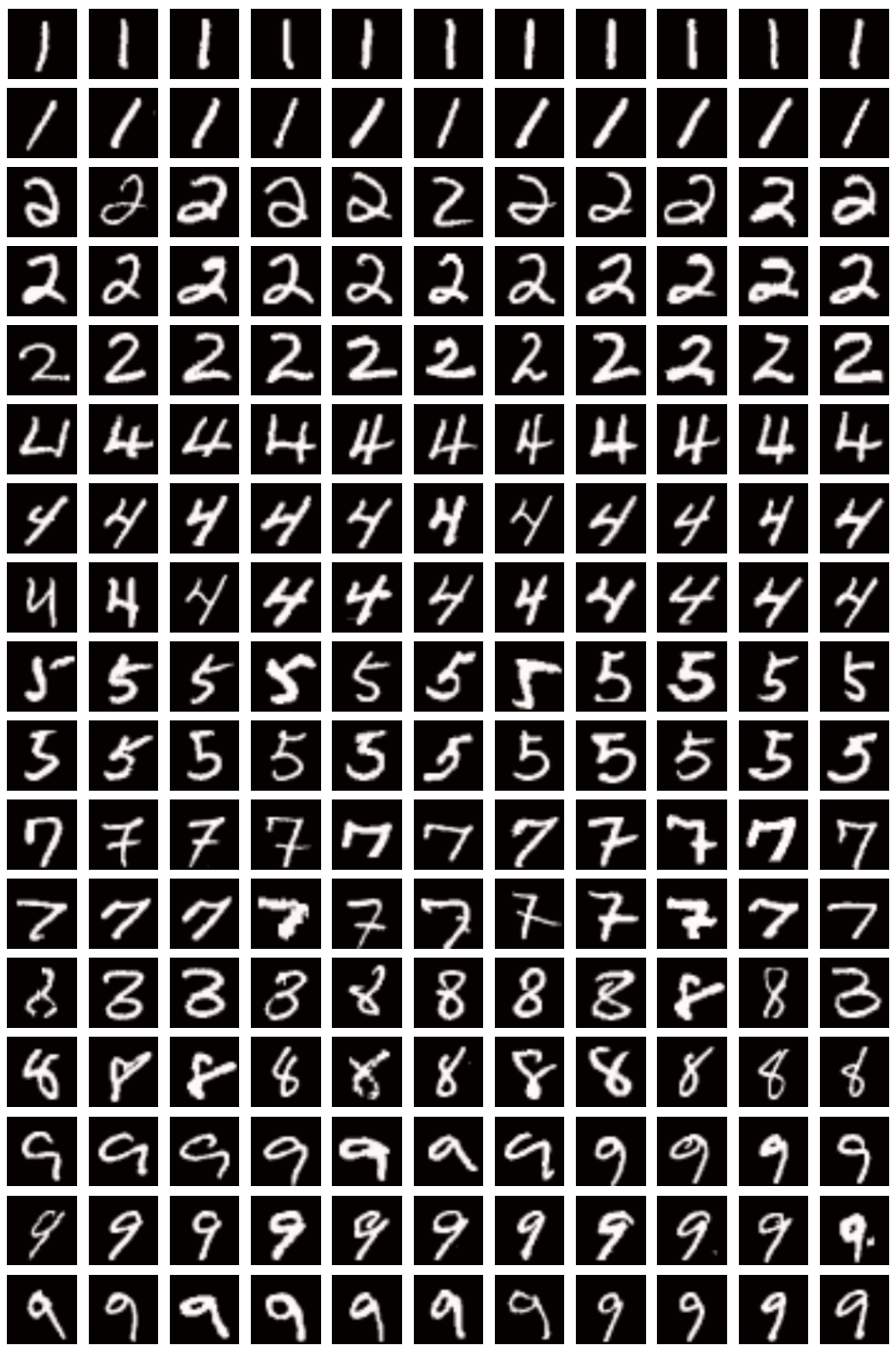

(a)

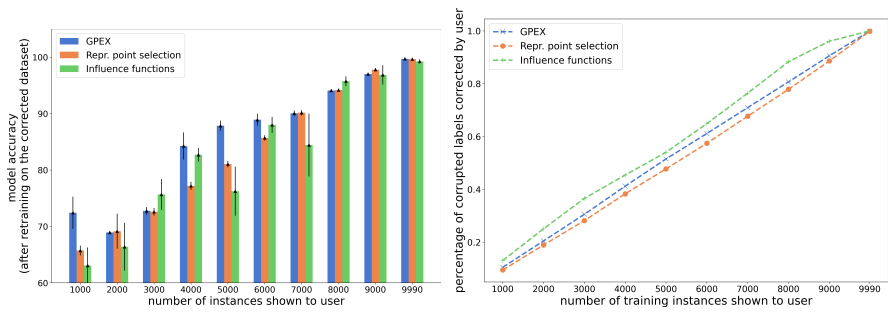

(b)

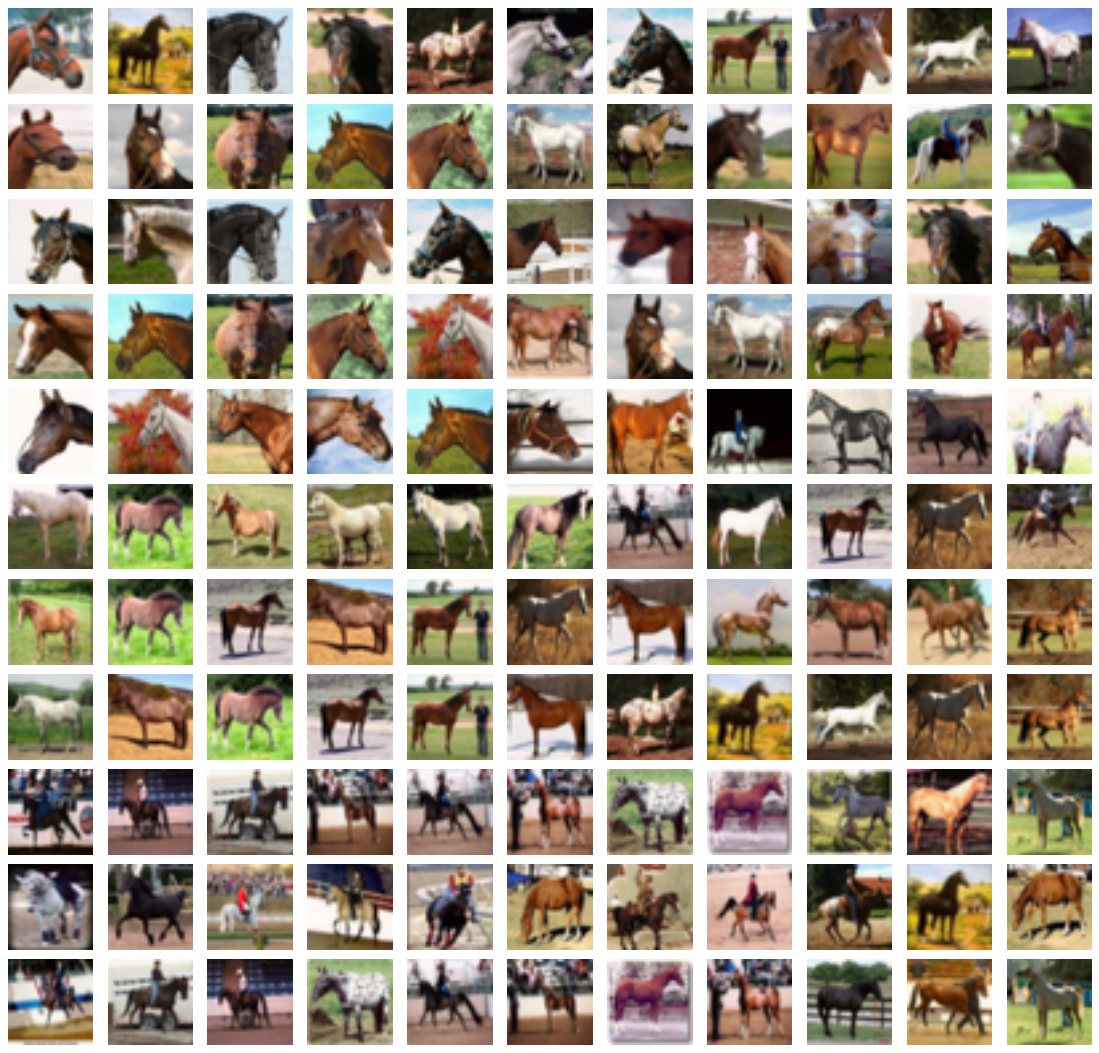

(c)

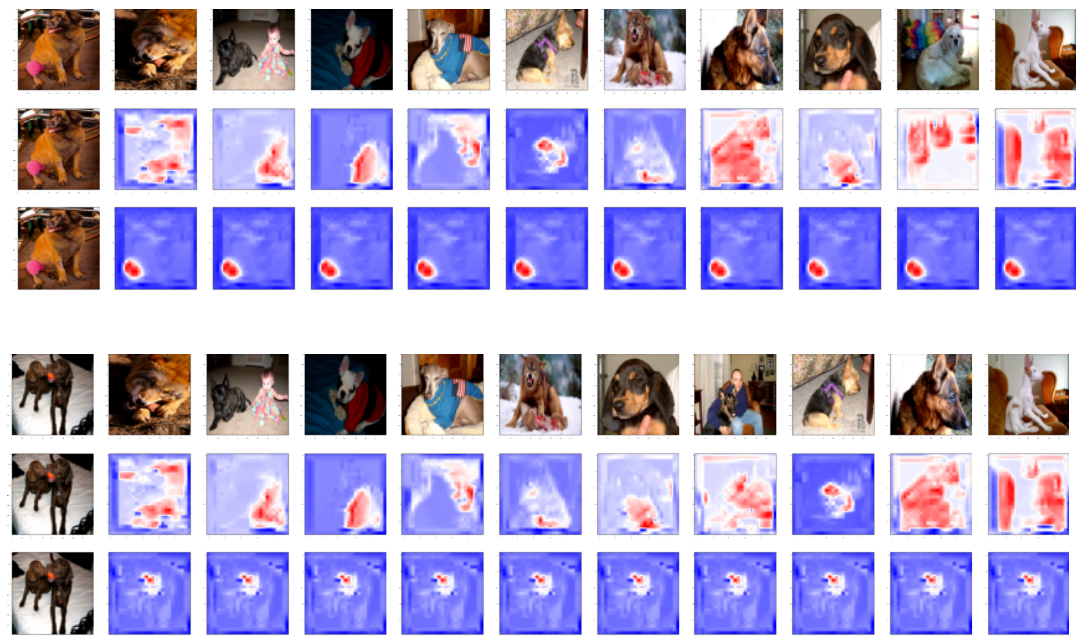

(d)

Figure 2: (a,c,d) Sample explanations for MNIST, Cifar10, and DogsWolves. In each row a test instance is shown in the first column, and the 10 nearest neighbours (in the kernel-space of the GP that corresponds to the output-head with maximum value at the test instance) is shown in columns 2-11. (b) Evaluating our proposed method, representer point selection [37], and influence functions [18] in dataset debugging task.

Therefore, when very few inducing points are available GP posterior mean is less flexible [36]. This is consistent with our parameter analysis in Sec. S7 of supplementary material.

In Fig. 1a we discussed that GPEX is not only able to explain a classifier ANN, but it can explain any ANN which is a subcomponent of any feed-forward pipeline. To evaluate this ability, we trained three classifiers with an attention mechanism [22]. Each classifier has two ResNet-18 [12] backbones: one extracts a volumetric map containing deep features, and the other produces a spatial attention mask. For each attention backbone, we set the width of the second last layer to 1024, followed by a linear layer and sigmoid activation. We applied our proposed GPEX (i.e. Alg.2) to each classifier, but this time the ANN to be explained (i.e. the box called "ANN" in Fig. 1a) is set to be the attention submodule. Note that each attention backbone produces a spatial attention mask of size $h$ by $w$. We think of each attention backbone as an ANN which has $h \times w$ output heads. We trained three classifier pipelines with attention mechanism on Cifar10 [19], MNIST [9], and Kather [15] with the same training procedure as previous part. In Fig. 3, 6-th, 7-th, and 8-th bar groups show the correlation coefficients between the attention backbones and the corresponding GPs on unseen test instances. According to Fig. 3, our proposed GPEX is able find GPs which are faithful to attention subcomponents of the classifier pipelines. Note that we didn't include all attention heads, because some pixels in attention masks are always off. In Sec. S3 of the supplementary material we have included more information and insights about the faithfulness of GPs to ANNs.

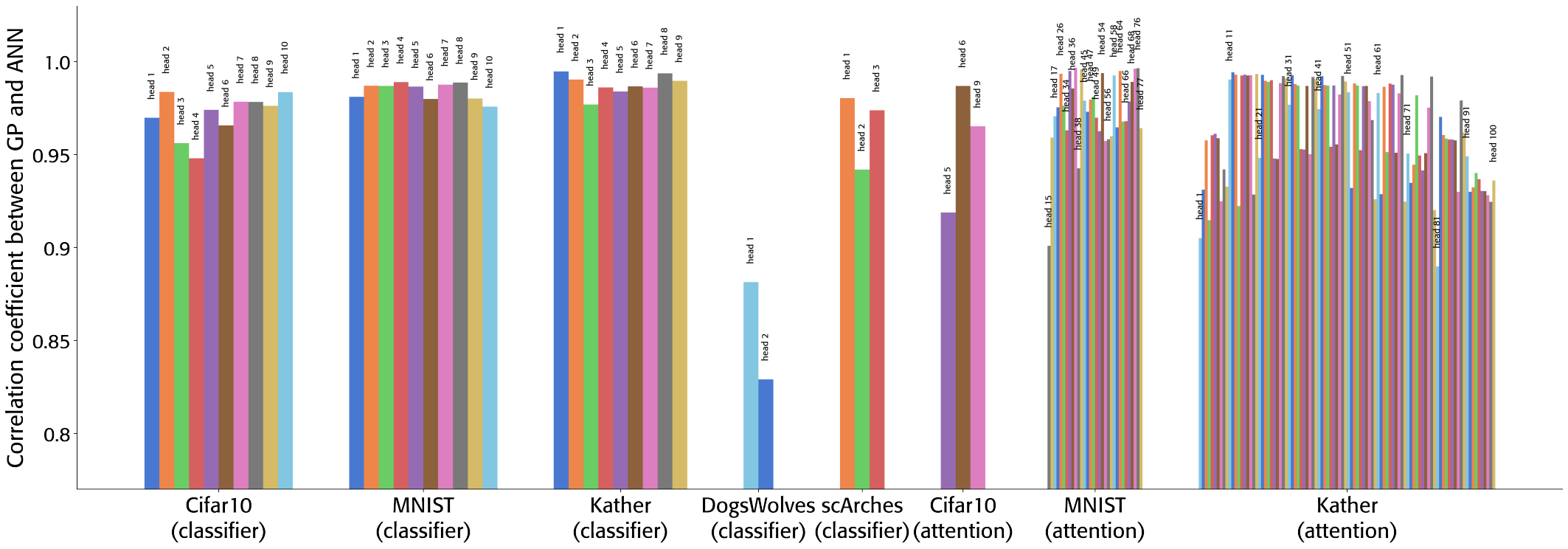

Figure 3: Faithfulness of GPs to ANNs measured by the Pearson correlation coefficient.

### 4.2 Explaining ANNs' Decisions

In Sec. 4.1 we trained four CNN classifiers on Cifar10 [19], MNIST [9], Kather [15], and DogsWolves [34], respectively. Afterwards, we applied our proposed explanation method to each CNN classifier. In this section, we are going to explain the decisions made by the classifiers via the obtained GPs found by Alg. 2. We explain the decision made for a test instance like $\boldsymbol{x}_{test}$ as follows. We consider the GP and the kernel-space that correspond to the ANN's head with maximum value (i.e. the ANN's head that relates to the predicted label). Consequently, among the instances in the inducing dataset, we find the 10 closest instances to $\boldsymbol{x}_{test}$, like $\{\boldsymbol{x}_{i1}, \boldsymbol{x}_{i2}, ..., \boldsymbol{x}_{i10}\}$. Intuitively the ANN has labeled $\boldsymbol{x}_{test}$ in that way because it has found $\boldsymbol{x}_{test}$ to be similar to $\{\boldsymbol{x}_{i1}, \boldsymbol{x}_{i2}, ..., \boldsymbol{x}_{i10}\}$.

For MNIST digit classification, some test instances and nearest neighbours in training set are shown in Fig. 2a. In this figure each row corresponds to a test instance. The first column depicts the test instance itself and columns 2 to 11 depict the 10 nearest neighbours. For example, in Fig. 2a the image in row3-col1 depicts a test instance $\boldsymbol{x}_{test}$ and the images in row3, cols2-11 depict the nearest neighbours $\{\boldsymbol{x}_{i1}, \boldsymbol{x}_{i2}, ..., \boldsymbol{x}_{i10}\}$. According to rows 1 and 2 of Fig. 2a, the classifier has labeled the two images as digit 1 because it has found 1 digits with similar inclinations in the training set (in Fig. 2a in row 1 all digits are vertical but in row 2 all digits are inclined). We see the model has also taken the inclination into account for the test instances of rows 7, 8, 15, 16, and 17 of Fig. 2a. In Fig. 2a, according to rows 3, 4, and 5 the test instances are classified as digit 2 because 2 digits with similar styles are found in the training set. We see the model has also taken the style into account for the test instances of rows 6, 7, 8, 9, 10, 12, 13, 14, 15, 16, and 17 of Fig. 2a. For instance, the test instance in row 6 of Fig. 2a is a 4 digit with a short stand and the two nearest neighbours are alike. Or for the test instances in rows 13, 14, and 15 of Fig. 2a the test instances have incomplete circles in the same way as their nearest neighbours. More explanations are provided in the supplementary material in Sec. S4.

Fig. 2c illustrates some sample explanations for Cifar10 [19]. Like before, each row corresponds to a test instance, the first column depicts the test instance itself and columns 2 to 11 depict the 10 nearest neighbours. In Fig. 2c, the test instances of rows 1, 2, 3, 4, and 5 are captured from horses' heads from closeby, and the nearest neighbours are alike. However, in rows 6, 7, 8, 9, 10, and 11 of Fig. 2c the test images are taken from faraway and the found similar training images are also taken from faraway. Intuitively, as the classifier is not aware of 3D geometry, it finds training images which are captured from the same distance. In rows 9, 10, and 11 of Fig. 2c, we see that the testing images contain riders. Similarly, the nearest neighbours also tend to have riders. Therefore, in rows 9, 10, and 11 of Fig. 2c the model has made use of the riders or other context information to classify the test instances as horse. More explanations are provided in the supplementary material in Sec. S4.

Besides finding the nearest neighbours, we provide CAM-like [38] explanations as to why $\boldsymbol{x}_{test}$ and an instance like $\boldsymbol{x}_{ij}, 1 \leq j \leq 10$ are considered similar by the model (according to the procedure of Sec. S2.2 in the supplementary material). Fig.2d illustrates some sample explanations for DogsWolves [34] dataset. In row 1 of Fig. 2d, the first column depicts the test instance itself and columns 2 to 11 depict the 10 nearest neighbours. The second and third rows highlight the pixels that contribute the most to the similarities. The second and third rows highlight the pixels of $\boldsymbol{x}_{test}$ and $\{\boldsymbol{x}_{i1}, \boldsymbol{x}_{i2}, ..., \boldsymbol{x}_{i10}\}$ respectively. According to row 3 of Fig. 2d, the pink object next to the dog's leg has contributed the most to the similarities. According to row 2 of Fig. 2d regions like the baby in column 3, the dog colar or costume in columns 4, 5, and 6, human finger in column 9, and the background in columns 10 and 11 have contributed the most to their similarity to the test instance. These are patterns that usually happen for dogs images. Indeed, since the training set has been small (1600 images), to detect dogs the model is making use of patterns that normally exist in indoor scenes and do not normally appear in wolves images. We see a similar pattern for the test instance in row 6 of Fig. 2d and also several explanations in Sec.S4 of the supplementary.

### 4.3 Comparing GPEX to Representer-Point Selection and Influence Functions

In Sec. S6 of the supplementary material we qualitatively compare GPEX explanations to those of representer point selection [37]. According to the experiments and detailed discussions of Sec. S6 in the supplementary, the GP's kernel that we find in this paper is superior due to a technical point in the formulation of representer point selection [37]. Besides the analysis of Sec. S6, we compared our proposed GPEX with representer point selection [37] and influence functions [18] in dataset debugging task. In these experiments we only selected images from Cifar10 [19] that are labeled as either automobile or horse. To corrupt the labels, we randomly selected 45% of training instances and changed their labels. Afterwards, we trained a classifier CNN with ResNet18 [12] backbone with the same training procedure explained in Sec. 4.1. In dataset debugging task, training instances are shown to a user in some order. After seeing an instance, the user checks the label of the instance and corrects it if needed. One can use explanation methods to bring the corrupted labels to the user's attention more quickly. Given an explanation method, we repeatedly select a test instance which is misclassified by the model. Afterwards, we show to the user the closest training instance (of course among the training instances which are not yet shown to the user). We repeat this process for test instances in turn until all training instances are shown to the user. We compared our proposed GPEX to representer point selection [37] and influence functions [18] in dataset debugging task. We used an implementation of influence functions [18] based on LiSSA [4] with 10 steps for each instance. The implementation is publicly available [1]. For representer point selection [37] we used the implementation by authors which is publicly available [3]. The result is shown in Fig. 2b. According to the plot on the left in Fig. 2b, when correcting the dataset by GPEX, the model accuracy becomes close to 90% after showing about 4000 instances to user. But when using representer point selection [37] or influence functions [18], this happens when the user has seen about 7000 training instances. With noisy labels model training becomes unstable. Therefore, in the plot on the left of Fig. 2b we repeat the training 5 times and we report the standard errors by the lines in top of the bars. According to the plot on the right of Fig. 2b, after showing a fixed number of training instances to the user, when using the proposed GPEX more corrupted labels are shown to the user. Indeed, GPEX brings the corrupted labels to the user's attention quicker than representer point selection [37] does. Interestingly, according to the plot in the right hand side of Fig. 2b influence functions [18] is quicker at spotting incorrect labels, but the instances found by our proposed method are more effective in increasing the accuracy quicker.

## 5 Related Work

The first theoretical connection between ANNs and GPs was that under some conditions, a random single-layer neural network converges to the mean of a Gaussian process [23] as the width of that single layer goes to infinity. This connection was later proven for ANNs with many layers [8], and for ANNs trained with gradient descent [14]. The theoretical requirements are usually too restrictive. For example, [8] requires all intermediate layers to be wide and also requires the dataset to be countable (so data-augmentations like color-jitter are not allowed). Or [24] requires the ANN to be trained with MSE loss and requires all intermediate layers to be wide. In this paper we do not presume any conditions on the ANN and simply distill knowledge from a neural network to some GPs. Of note, those theoretical conditions may facilitate knowledge distillation and improve the Pearson correlation coefficient between the ANNs and the GPs obtained by our method.

Scalability is a major issue when training GPs, and including a few inducing points may limit the flexibility of GP's posterior [36]. Here we review some previous methods to tackle the computational challenges of training GPs. SV-DKL [27] derives a lower-bound for training a GP with a deep kernel. In this method, a grid of inducing points are considered in the kernel-space (like the vectors $\{(\tilde{\boldsymbol{u}}_m^{(\ell)}, \tilde{v}_m^{(\ell)})\}_{m=1}^M$ with the notation of this paper). Afterwards, each input instance is firstly mapped to the kernel-space and the output is computed based on similarities to the grid points in the kernel-space. Since the GP posterior is computed via the grid points, SV-DKL [27] is scalable. But unfortunately the number of grid points cannot be increased to above 1000 even for Cifar10 [19] and with a RTX 3090 GPU. Therefore, this may limit the flexiblity of the GP's posterior [36].

A more recent framework called GPytorch [29] provides GPU acceleration. However, its computational complexity is quadratic in number of inducing points. Other approaches to improve scalability of GPs include: considering structured kernel matrices [7], kernel interpolation [35], and imposing grid-structure on including points [27]. Stack of Gaussian processes are shown to be connected to ANNs [10][30][24]. By stacking kernels, GP kernels work on intermediate representations and therefore are not necessarily interpretable to humans. But in our method the GPs' kernels work directly on the input-space itself. Knowledge distillation (KD) is closely related to this work. With the best of our knowledge and according to the authors, [6] is the first work that applies KD to GPs. But the distinction of our work is that we distill knowledge from ANN to GP, as opposed to the self-distillation of [6] that distills knowledge from a GP to another GP.

**Limitations and Outlook**: In this work we used Eq. 5 to distill knowledge from ANN to GP. One may use Eq. 6 to distill knowledge from GP to ANN in order to, e.g., transfer GP's good generalization to the ANN. Our method scales very well, and Alg. 2 runs without memory/computational issues even on imagenet with more than 1M inducing points (i.e. images) and Resnet-18 [12] when a few output-heads are selected, but on imagenet we failed to match the GPs to ANN in a 2-3 day runtime. The issue is that the $\mathbf{U}$ matrices have to be updated very often (the update of line 8 of Alg. 2) so that GPs' kernels are updated according to an accurate estimate of kernel-space representations. Otherwise the convergence may not happen especially for millions of inducing points and a small batch-size (as required for, e.g., CNNs). We used control-variate [25], but one may use more advanced heuristics [35] to achieve convergence for datasets like imagenet and with a reasonable computation time. In this paper we analyzed the effect of number of inducing points, the width of the second last layer, and the number of epochs for which the ANN is trained. One can use the proposed tool to answer other questions, like, is the GP kernel required to have more parameters than the ANN itself? May it so happen that a test instance is equally close to hundreds of training instances thereby limiting a human's ability to understand ANNs decision? Is the uncertainty provided by the GP correlated to the understandability of the explanations to humans or to the ANN's failures?

## 6 Acknowledgements

The experiments of this paper were enabled in part by the Digital Research Alliance of Canada. This work was supported in part by the NSERC Discovery Grant.

# Supplementary Material for GPEX, A Framework For Interpreting Artificial Neural Networks

Amir Akbarnejad, Gilbert Bigras, Nilanjan Ray

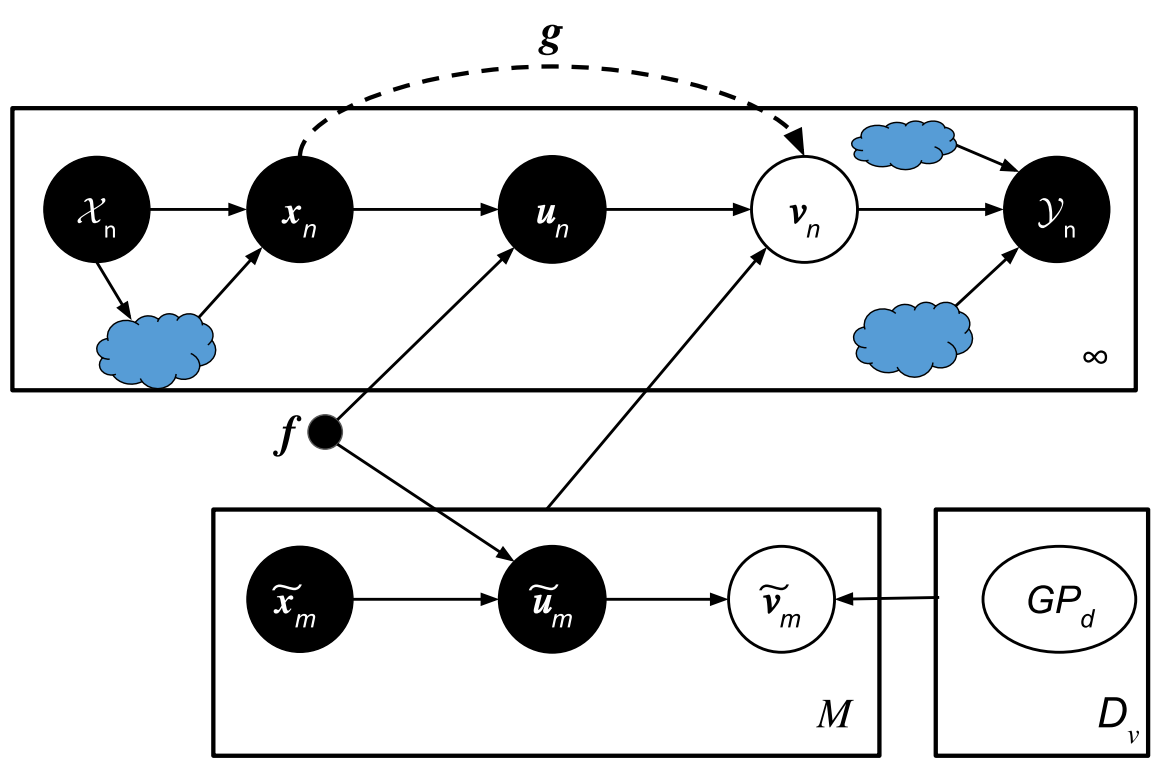

Fig. S1: The proposed framework as a probabilistic graphical model.

## S1 Deriving the Variational Lower-bound

In this section we derive the variational lower-bound introduced in Sec.2.3 of the main article. We firstly introduce Lemmas 1 and 2 as they appear in our derivations.

***Lemma 1.*** The KL-divergence between two normal distributions $\mathcal{N}_1(.\ ;\ \boldsymbol{\mu}_1, \boldsymbol{\Sigma}_1)$ and $\mathcal{N}_2(.\ ;\ \boldsymbol{\mu}_2, \boldsymbol{\Sigma}_2)$ can be computed as follows:

$$KL\Big(\mathcal{N}_1 \mid\mid \mathcal{N}_2\Big) = \frac{1}{2}\Big( \log(\frac{|\boldsymbol{\Sigma}_2|}{|\boldsymbol{\Sigma}_1|}) \ - \ D \ + \ trace\{\boldsymbol{\Sigma}_2^{-1}\boldsymbol{\Sigma}_1\} + (\boldsymbol{\mu}_2 - \boldsymbol{\mu}_1)^T \boldsymbol{\Sigma}_2^{-1}(\boldsymbol{\mu}_2 - \boldsymbol{\mu}_1)\Big). \blacksquare \tag{S1}$$

***Lemma 2.*** Let $p_1$ and $p_2$ be two normal distributions:

$$p_1(x) = \mathcal{N}(x\ ;\ \mu_1, \sigma_1^2),$$
$$p_2(x) = \mathcal{N}(x\ ;\ \mu_2, \sigma_2^2).$$

We have that

$$\mathbb{E}_{x\sim p_2}\big[\log\ p_1(x\ ;\ \mu_1, \sigma_1^2)\big] = -\frac{(\mu_1-\mu_2)^2+\sigma_2^2}{2\sigma_1^2} - \frac{1}{2}\log(\sigma_1^2) - \frac{1}{2}\log(2\pi). \blacksquare \tag{S2}$$

Fig.S1 illustrates the framework as a probabilistic graphical model. A general feed-forward pipeline takes in a set of input(s) $\mathcal{X}$ and produces a set of output(s) $\mathcal{Y}$. The general pipeline is required to have at least one ANN as a submodule. The ANN submodule is required to take in only one input $\boldsymbol{x}$ and to produce only one output $\boldsymbol{v}$, where $\boldsymbol{x}$ and $\boldsymbol{v}$ are tensors of arbitrary sizes. As illustrated in Fig.S1, the ANN's input $\boldsymbol{x}$ can depend arbitrarily on some other intermediate variables in the pipeline. This relation is modeled by the conditional distribution $p(\boldsymbol{x}_n|Parent(\boldsymbol{x}_n))$ where $Parent(\boldsymbol{x}_n)$ is the set of all variables which are connected to $\boldsymbol{x}_n$. Similarly, as illustrated in Fig.S1 the pipeline's output $\mathcal{Y}$ can arbitrarily depend on some intermediate variables in the pipeline. This relation is modeled by the conditional distribution $p(\mathcal{Y}_n|Parent(\mathcal{Y}_n))$. In Fig.S1 the lower boxes are the inducing points and other variables that determine the GPs' posterior. More precisely, in Fig.S1 $\{\tilde{\boldsymbol{x}}_m\}_{m=1}^M$ are some inducing points (e.g. some training images). Vectors in the kernel space are denoted by $\tilde{\boldsymbol{u}}$ and $\boldsymbol{u}$. Moreover, the observed values are denoted by $v$ and $\tilde{v}$. Informally, $\boldsymbol{u}$ and $v$ denote the input/output of the GPs. When referring to one of the $M$ inducing points a "tilde" is used (as $(\tilde{\boldsymbol{u}}, \tilde{v})$), however $(\boldsymbol{u}, v)$ corresponds to a point that can be anywhere in the kernel-space.

The inducing instances $\{\tilde{\boldsymbol{x}}_m\}_{m=1}^M$ are mapped to the kernel-spaces by the kernel mappings $\{f_1(.), ..., f_L(.)\}$. In Fig.S1 the variables $\{\tilde{\boldsymbol{u}}_m\}_{m=1}^M$ are the kernel-space representations of the inducing points $\{\tilde{x}_m\}_{m=1}^M$. Moreover, $\{\tilde{\boldsymbol{v}}_m\}_{m=1}^M$ are the GP's output values at the inducing points. Given an instance $\boldsymbol{x}_n$, it is firstly fed to the kernel mappings $\{f_1(.), ..., f_L(.)\}$ and the kernel-space representations $\boldsymbol{u}_n$ are obtained. Afterwards, the GPs' outputs on $\boldsymbol{u}_n$ depend on $\boldsymbol{u}_n$ as well as all other inducing points because the inducing points actually determine the GPs' posterior on all kernel-space points including $\boldsymbol{u}_n$. Therefore, in Fig.S1 the variable $\boldsymbol{v}_n$ is not only connected to $\boldsymbol{u}_n$ but it is also connected to the box at the bottom (i.e. all inducing points and other variables associated with them).

As usual, the variational lower-bound is equal to

$$\mathcal{L} = \mathbb{E}_{\sim q}\big[\log p(\text{all variables})\big] - \mathbb{E}_{\sim q}\big[\log q(\text{hidden variables})\big]. \tag{S3}$$

The likelihood of all variables in Eq.S3 factorizes as the product of conditional distributions of each variable given

its parents. Therefore

$$p(\text{all variables}) = \prod_{variable\ t} p(t|Parent(t)). \tag{S4}$$

In Eq.S4 only some conditional distributions appear in our derivations which are discussed at the following.

- The variable $\boldsymbol{x}_n$: the ANN's input $\boldsymbol{x}_n$ can depend arbitrarily on some other intermediate variables in the pipeline. In our derivations we leave this conditional distribution as $p(\boldsymbol{x}_n|Parent(\boldsymbol{x}_n))$.
- The variable $\boldsymbol{u}_n$: Given a training instance $\boldsymbol{x}_n$, the kernel-space representations $\boldsymbol{u}_n$ are deterministically obtained by feeding the instance to the kernel-mappings $[f_1(.), ..., f_L(.)]$.
- The variable $\boldsymbol{v}_n$: The ANN's output is required to depend only on the input, so
$$p(\boldsymbol{v}_n|Parent(\boldsymbol{v}_n)) = p(\boldsymbol{v}_n|\boldsymbol{u}_n, \boldsymbol{x}_n, \{\tilde{\boldsymbol{x}}_m, \tilde{\boldsymbol{u}}_m, \tilde{v}_m\}_{m=1}^M). \tag{S5}$$
The above distribution is actually the GPs' posterior at $\boldsymbol{u}_n$ (i.e. the normal distribution of Eq.1 of the main article).
- The variable $\tilde{\boldsymbol{x}}_m$: the inducing point $\tilde{\boldsymbol{x}}_m$ can depend arbitrarily on some other intermediate variables in the pipeline. In our derivations we leave this conditional distribution as $p(\tilde{\boldsymbol{x}}_m|Parent(\tilde{\boldsymbol{x}}_m))$.
- The variable $\tilde{\boldsymbol{u}}_m$: Given an inducing point $\tilde{\boldsymbol{x}}_m$, the kernel-space representations $\tilde{\boldsymbol{u}}_m$ are deterministically obtained by feeding the inducing point $\tilde{\boldsymbol{x}}_m$ to the kernel-mappings $[f_1(.), ..., f_L(.)]$.
- The variables $\hat{\boldsymbol{v}}_m$: Given the kernel-space representations $\{\tilde{\boldsymbol{u}}_m^{(\ell)}\}_{m=1}^M$, the variables $\{\tilde{v}_1^{(\ell)}, ..., \tilde{v}_M^{(\ell)}\}$ follow a $M$-dimensional Gaussian distribution with zero mean and a covariance matrix determined by the GP prior covariance among the variables $\{\tilde{\boldsymbol{u}}_m^{(\ell)}\}_{m=1}^M$.
- The variable $\mathcal{Y}_n$: the pipeline's output $\mathcal{Y}$ can arbitrarily depend on some intermediate variables in the pipeline. In our derivations we leave this conditional distribution as $p(\mathcal{Y}_n|Parent(\mathcal{Y}_n))$.

According to Eq.S4, the likelihood of all variables factorizes as

$$\begin{aligned}
p(\text{all variables}) &= \prod_{variable\ t} p(t|Parent(t)) \\
&= \big(\prod_n p(\boldsymbol{x}_n|Parent(\boldsymbol{x}_n))\big) \times \big(\prod_n p(\boldsymbol{u}_n|\boldsymbol{x}_n)\big)\times \\
&\big(\prod_n \prod_\ell p(v_n^{(\ell)}|\boldsymbol{u}_n, \boldsymbol{x}_n, \{\tilde{\boldsymbol{x}}_m, \tilde{\boldsymbol{u}}_m, \tilde{v}_m\}_{m=1}^M)\big)\times \\
&\big(\prod_m p(\tilde{\boldsymbol{x}}_m|Parent(\tilde{\boldsymbol{x}}_m))\big) \times \big(\prod_m p(\tilde{\boldsymbol{u}}_m|\tilde{\boldsymbol{x}}_m)\big)\times \\
&\big(\prod_\ell p(\tilde{\boldsymbol{v}}_{1:M}^{(\ell)}|\boldsymbol{0}, \mathcal{K}_{prior}(\tilde{\boldsymbol{u}}_{1:M}^{(\ell)}, \tilde{\boldsymbol{u}}_{1:M}^{(\ell)}))\big)\times \\
&\big(\prod_n p(\mathcal{Y}_n|Parent(\mathcal{Y}_n))\big) \times \big(\prod_{\text{other vars } t} p(t|Parent(t))\big).
\end{aligned} \tag{S6}$$

Now we derive the lower-bound $\mathcal{L}$ with respect to each parameter separately.

## S1.1 Deriving the Lower-bound With Respect to the Kernel-mappings

In the right-hand-side of Eq.S6 only the following terms are dependant on the kernel-mappings $[f_1(.), ..., f_L(.)]$:

$$\begin{aligned}
&\big[\prod_m p(\tilde{\boldsymbol{u}}_m|\tilde{\boldsymbol{x}}_m) \times \prod_\ell p(\tilde{v}_m^{(\ell)}|\boldsymbol{0}, \mathcal{K}_{prior}(\tilde{\boldsymbol{u}}_{1:M}^{(\ell)}, \tilde{\boldsymbol{u}}_{1:M}^{(\ell)}))\big]\times \\
&\big[\prod_n p(\boldsymbol{u}_n|\boldsymbol{x}_n) \times \prod_\ell p(v_n^{(\ell)}|\boldsymbol{u}_n, \boldsymbol{x}_n, \{\tilde{\boldsymbol{x}}_m, \tilde{\boldsymbol{u}}_m, \tilde{v}_m\}_{m=1}^M)\big].
\end{aligned} \tag{S7}$$

Note that in the above equation the terms $p(\tilde{\boldsymbol{u}}_m|\tilde{\boldsymbol{x}}_m)$ and $p(\boldsymbol{u}_n|\boldsymbol{x}_n)$ are equal to 1 because $\tilde{\boldsymbol{u}}_m$ and $\boldsymbol{u}_n$ are deterministically obtained from $\tilde{\boldsymbol{x}}_m$ and $\boldsymbol{x}_n$. Therefore, in Eq.S3 the terms containing the kernel mappings $[f_1(.), ..., f_L(.)]$ are as follows:

$$\begin{aligned}
\mathcal{L}_f &= \mathbb{E}_{\sim q}\big[\sum_\ell \log p(v^{(\ell)}|\boldsymbol{u}, \boldsymbol{x}, \{\tilde{\boldsymbol{x}}_m, \tilde{\boldsymbol{u}}_m, \tilde{v}_m\}_{m=1}^M)\big]+ \\
&\sum_\ell \mathbb{E}_{\sim q}\big[\log p(\tilde{\boldsymbol{v}}_{1:M}^{(\ell)}|\boldsymbol{0}, \mathcal{K}_{prior}(\tilde{\boldsymbol{u}}_{1:M}^{(\ell)}, \tilde{\boldsymbol{u}}_{1:M}^{(\ell)}))\big]- \\
&\sum_\ell \mathbb{E}_{\sim q}\big[\log q_2(\tilde{\boldsymbol{v}}_{1:M}^{(\ell)})\big] \\
&= \mathbb{E}_{\sim q}\big[\sum_\ell \log p(v^{(\ell)}|\boldsymbol{u}, \boldsymbol{x}, \{\tilde{\boldsymbol{x}}_m, \tilde{\boldsymbol{u}}_m, \tilde{v}_m\}_{m=1}^M)\big]- \\
&\sum_{\ell=1} \mathbb{E}_{\sim q}\big[KL\Big(q_2(\tilde{\boldsymbol{v}}_{1:M}^{(\ell)})\ \ ||\ \ p(\tilde{\boldsymbol{v}}_{1:M}^{(\ell)}|\boldsymbol{0}, \mathcal{K}_{prior}(\tilde{\boldsymbol{u}}_{1:M}^{(\ell)}, \tilde{\boldsymbol{u}}_{1:M}^{(\ell)}))\Big)\big].
\end{aligned} \tag{S8}$$

We simplify the two terms on the right-hand-side of Eq.S8. The first term is the expected log-likelihood of a Gaussian distribution (i.e. the conditional log-likelihood of $\tilde{v}^\ell$ as in Eq.1 of the main article). Also the variational distribution $q(.)$ is Gaussian. Therefore, we can use Lemma.2 to simplify the first term:

$$\begin{aligned}
&\mathbb{E}_{\sim q}\big[\sum_{\ell=1}^L \log p(v^{(\ell)}|\boldsymbol{u}, \boldsymbol{x}, \{\tilde{\boldsymbol{x}}_m, \tilde{\boldsymbol{u}}_m, \tilde{v}_m\}_{m=1}^M)\big] = \\
&\sum_{\ell=1}^L \mathbb{E}_{\sim q}\big[\log p(v^{(\ell)}|\boldsymbol{u}, \boldsymbol{x}, \{\tilde{\boldsymbol{x}}_m, \tilde{\boldsymbol{u}}_m, \tilde{v}_m\}_{m=1}^M)\big] = \\
&\sum_{\ell=1}^L \Big[-\frac{\big(\mu_v(\boldsymbol{u}^{(\ell)}, \tilde{\mathbf{u}}_{1:M}^{(\ell)}, \tilde{v}_{1:M}^{(\ell)}) - g_\ell(\boldsymbol{x})\big)^2 + \sigma_g^2}{cov_v(\boldsymbol{u}^{(\ell)}, \tilde{\mathbf{u}}_{1:M}^{(\ell)}, \tilde{v}_{1:M}^{(\ell)})} \\
&\quad -\frac{1}{2}\log\big(cov_v(\boldsymbol{u}^{(\ell)}, \tilde{\mathbf{u}}_{1:M}^{(\ell)}, \tilde{v}_{1:M}^{(\ell)})\big) \\
&\quad -\frac{1}{2}\log(2\pi)\Big].
\end{aligned} \tag{S9}$$

Note that the two terms of Eq.S9 are the two terms which were presented and discussed in Eq.5 of the main article.

Now we simplify the KL-term on the right-hand-side of

Eq.S8. According to Lemma.1 we have that

$$KL\Big(q_2(\tilde{\boldsymbol{v}}_{1:M}^{(\ell)}) \ || \ p(\tilde{\boldsymbol{v}}_{1:M}^{(\ell)}|\mathbf{0}, \mathcal{K}_{prior}(\tilde{\boldsymbol{u}}_{1:M}^{(\ell)}, \tilde{\boldsymbol{u}}_{1:M}^{(\ell)}))\Big) = + 0.5\big(\log(\frac{\sigma_{gp}^2}{\sigma_{\varphi}^2})\big) - 0.5M + \frac{\sigma_{\varphi}^2}{\sigma_{gp}^2} + \frac{\boldsymbol{\varphi}_{1:M}^{(\ell)^T}\boldsymbol{\varphi}_{1:M}^{(\ell)}}{\sigma_{gp}^2}, \tag{S10}$$

where $\varphi$ are the variational parameters of $q_2(.)$ as in Eq.4 of the main article. Therefore, the KL-term of Eq.S8 is a constant with respect to the kernel mappings $[f_1(.), ..., f_L(.)]$ and can be discarded. All in all, the lower-bound for optimizing the kernel-mappings is equal to the right-hand-side of Eq.S9 which was introduced and discussed in Sec.2.3. of the main article.

## S1.2 Deriving the Lower-bound With Respect to the ANN Parameters

According to Eq.4 of the main article, in our formulation the ANN's parameters appear as some variational parameters. Therefore, the likelihood of all variables (Eq.S6) does not generally depend on the ANN's parameters. But according to the general ELBO formulation in Eq.S3 the ELBO $\mathcal{L}$ depends on ANN's parameters, because when computing the expectation the variables are drawn from the variational distribution $q(.)$. We estimated the ELBO of Eq.S3 by the average over few samples. More precisely, given a training instance $\boldsymbol{x}$, we firstly computed the kernel-space representations as:

$$\boldsymbol{u}^{(\ell)} = f_\ell(\boldsymbol{x}), \ \ 1 \leq \ell \leq L. \tag{S11}$$

Afterwards, we used the reparametrization trick for Eq.1 of the main article to draw a sample for $\boldsymbol{v}^{(\ell)}$ as follows:

$$\begin{aligned} z_{q2}^{(\ell)} &\sim \mathcal{N}(0,1), \\ v^{(\ell)} &\sim \mu_v(\boldsymbol{u}^{(\ell)}, \tilde{\mathbf{u}}_{1:M}^{(\ell)}, \tilde{v}_{1:M}^{(\ell)}) + z_{q2}^{(\ell)} cov_v(\boldsymbol{u}^{(\ell)}, \tilde{\mathbf{u}}_{1:M}^{(\ell)}, \tilde{v}_{1:M}^{(\ell)}), \end{aligned} \tag{S12}$$

where $\mu_v(.,.,.)$ and $cov_v(.,.,.)$ are defined in Eqs.2 and 3 of the main article. Moreover, we continue the forward pass of the original pipeline to get a sample $\mathcal{Y}$. Having drawn $\boldsymbol{x}$, $\boldsymbol{u}$, $v$, and $\mathcal{Y}$ from the variational distribution, we estimate the ELBO of Eq.S3 by these samples.

$$\begin{aligned} \mathcal{L} &= \\ &\mathbb{E}_{\sim q}\big[\log p(\text{all variables})\big] - \mathbb{E}_{\sim q}\big[\log q(\text{hidden variables})\big] \\ &\approx \log p(\text{all variables})\Big|_{\boldsymbol{x},\boldsymbol{u},v,\mathcal{Y}} - \sum_m^M \sum_\ell^L \mathbb{E}_{\sim q_2}\big[\log q_2(\tilde{v}_m^{(\ell)})\big] \end{aligned} \tag{S13}$$

In the above equation, the second term on the right-hand-side is the entropy of a normal distribution and it only depends on the variance of the $q_2$ distribution. As we let the variance of $q_2$ be fixed ($\sigma_g^2$ in Eq.4 of the main article), the second term is a constant. Therefore,

$$\mathcal{L} \approx \log p(\text{all variables})\Big|_{\boldsymbol{x},\boldsymbol{u},v,\mathcal{Y}}. \tag{S14}$$

Among the likelihood term on the right-hand-side of Eq.S6 the conditional distribution of all variables before $\boldsymbol{u}_n$ (e.g. $\boldsymbol{x}_n$ and $\mathcal{X}_n$) are independent of the ANN's parameters (i.e. the parameters of the function $g(.)$). On the other hand, for all variables that appear after $\boldsymbol{u}_n$, the conditional distribution depends on the ANN's parameters. Indeed, according to Eq.S14

$$\begin{aligned} \mathcal{L}_{ann} \approx &\big[\sum_{\ell=1}^{L} \log p(v^{(\ell)}|\boldsymbol{u}, \boldsymbol{x}, \{\tilde{\boldsymbol{x}}_m, \tilde{\boldsymbol{u}}_m, \tilde{v}_m\}_{m=1}^M)\big]\Big|_{\boldsymbol{x},\boldsymbol{v}} + \\ &\log p(\mathcal{Y}|Parent(\mathcal{Y}))\Big|_{\boldsymbol{x},\boldsymbol{v},\mathcal{Y}} + \\ &\big(\sum_{\text{other vars after } \boldsymbol{u}_n} \log p(t|Parent(t))\big)\Big|_{\boldsymbol{x},\boldsymbol{v},\mathcal{Y}}. \end{aligned} \tag{S15}$$

In the above equation, the first term on the right-hand-side is the log-likelihood of the normal distribution of Eq.1:

$$\begin{aligned} &\log p(v^{(\ell)}|\boldsymbol{u}, \boldsymbol{x}, \{\tilde{\boldsymbol{x}}_m, \tilde{\boldsymbol{u}}_m, \tilde{v}_m\}_{m=1}^M) = \\ &-\frac{1}{2}\big[\sum_{\ell=1}^{L} \frac{(\mu_v(\boldsymbol{u}^{(\ell)}, \tilde{\mathbf{u}}_{1:M}^{(\ell)}, \tilde{v}_{1:M}^{(\ell)}) - g_\ell(\boldsymbol{x}))^2}{cov_v(\boldsymbol{u}^{(\ell)}, \tilde{\mathbf{u}}_{1:M}^{(\ell)}, \tilde{v}_{1:M}^{(\ell)})}\big] \\ &+ \big(\text{some terms independent from } g(.)\big). \end{aligned} \tag{S16}$$

In Eq.S15 the term $p\big(\mathcal{Y}|Parent(\mathcal{Y})\big)$ is the likelihood of the output(s) of the whole pipeline as illustrated by Fig.1a of the main article, given the ANN's output and all other intermediate variables on which the final output $\mathcal{Y}$ depends. This likelihood turns out to be equivalent to commonly-used losses like the cross-entropy loss or the mean-squared loss. Here we elaborate upon how this happens. Let the task be a classification, and let $\hat{\mathcal{Y}} \in \mathbb{R}^L$ be the pipeline's output. The final model prediction $\mathcal{Y}$ is done as follows:

$$\mathcal{Y} \sim \mathcal{C}ategorical(\hat{\mathcal{Y}}_K, ..., \hat{\mathcal{Y}}_K) \tag{S17}$$

Therefore we have that

$$p\big(\mathcal{Y}|Parent(\mathcal{Y})\big) = (\hat{\mathcal{Y}}_1)^{I[\mathcal{Y}==1]} \times ... \times (\hat{\mathcal{Y}}_K)^{I[\mathcal{Y}==K]}, \tag{S18}$$

where $I[.]$ is the indicator function. So, we have that

$$\begin{aligned} &\log p\big(\mathcal{Y}|Parent(\mathcal{Y})\big) = \\ &I[\mathcal{Y} == 1]\log(\hat{\mathcal{Y}}_1) + ... + I[\mathcal{Y} == K]\log(\hat{\mathcal{Y}}_K). \end{aligned} \tag{S19}$$

Therefore, when the pipeline is for classification, $\log p(\mathcal{Y}|\mathbf{v}, \ \ etc.)$ will be equal to the cross-entropy loss. This conclusion was introduced and discussed in Eq.6 of the main article. We can draw similar conclusions when the pipeline is for other tasks like regression, or even a combination of tasks.

In the general pipeline of Fig.S1, if all stages after $\boldsymbol{v}$ are deterministic (of course except the final stage which is probabilistic like Eq.S17), the third term on the right-hand-side of Eq.S15 becomes 1. Therefore, the right-hand-side of Eq.S15 is equal to Eq.6 of the main article. As we discussed in Sec.2.3 of the main article, $\mathcal{L}_{ann}$ has two terms: the first terms encourages the GP-ANN analogy and the second term seeks to lower the task-loss.

**Algorithm S1** Method Forward_GP

**Input:** Input instance $\boldsymbol{x}$ and inducing instance $\tilde{\boldsymbol{x}}$, list of matrices $\mathbf{U}$, list of vectors $\mathbf{V}$.

**Output:** List of GP posterior means $\mu$, and covariances $cov$.

*Initialisation* : $\mu = list(L)$, $cov = list(L)$.

1: **for** $\ell = 1$ to $L$ **do**
2: $u = f_\ell(\boldsymbol{x})$ //map $\boldsymbol{x}$ to the kernel space of the $\ell$-th GP.
3: $\mathbf{U}_\ell \leftarrow \mathbf{U}[L]$ //get the inducing points of the $\ell$-th GP.
4: $\mathbf{V}_\ell \leftarrow \mathbf{V}[L]$ //observed values at the inducing points.
5: **if** training **then**
6: $\mathbf{U}_\ell[\tilde{\boldsymbol{x}}.index] \leftarrow f_\ell(\tilde{\boldsymbol{x}})$ //to pass gradient w.r.t. $f_\ell(.)$
7: **end if**
8: $\mu[\ell] \leftarrow \boldsymbol{u}^T \mathbf{U}_\ell^T \big(\mathbf{U}_\ell \mathbf{U}_\ell^T + \sigma_{gp}^2 \mathbf{I}\big)^{-1} \mathbf{V}_\ell$.
9: $cov[\ell] \leftarrow \boldsymbol{u}^T \boldsymbol{u} - \boldsymbol{u}^T \mathbf{U}_\ell^T \big(\mathbf{U}_\ell \mathbf{U}_\ell^T + \sigma_{gp}^2 \mathbf{I}\big)^{-1} \mathbf{U}_\ell \boldsymbol{u}$.
10: **end for**
11: **return** $\mu$ and $cov$

**Algorithm S2** Method Optim_KernMappings

**Input:** Input instance $\boldsymbol{x}$ and inducing instance $\tilde{\boldsymbol{x}}$, list of matrices $\mathbf{U}$, list of vectors $\mathbf{V}$.

**Output:** Kernel-space mappings $[f_1(.), ..., f_L(.)]$.

*Note the important modifiactions to Alg.S2 which are explained in Sec.S5.*

*Initialisation* : $loss \leftarrow 0$.

1: $\mu$, $cov \leftarrow$ forward_GP($\mathbf{x}$, $\tilde{\boldsymbol{x}}$, $\mathbf{U}$, $\mathbf{V}$) //feed $\mathbf{x}$ to GPs.
2: $\mu_{ann} \leftarrow g(\mathbf{x})$ //feed $\mathbf{x}$ to ANN.
3: **for** $\ell = 1$ to $L$ **do**
4: $loss \leftarrow loss + \frac{(\mu[\ell] - \mu_{ann}[\ell])^2 + \sigma_g^2}{cov[\ell]} + \log(cov[\ell])$. //Eq.5.
5: **end for**
6: $\boldsymbol{\delta} \leftarrow \frac{\partial \ loss}{\partial \ params\big([f_1(.), ..., f_L(.)]\big)}$.//the gradient of loss.
7: $params\big([f_1, ..., f_L]\big) \leftarrow params\big([f_1, ..., f_L]\big) - lr \times \boldsymbol{\delta}$ //update the parameters.
8: $lr \leftarrow$ updated learning rate
9: **return** $[f_1(.), ..., f_L(.)]$

### S1.3 Deriving the Lower-bound With Respect to $q_2(.)$ Parameters

In Eq.4 of the main article we considered the variational parameters $\{\varphi_m^{(\ell)}\}_{m=1}^M$ for the hidden variables $\{\tilde{v}_m^{(\ell)}\}_{m=1}^M$. The ELBO of Eq.S3 can be optimized with respect to $\{\varphi_m^{(\ell)}\}_{m=1}^M$ as well. But we noticed that optimizing $\{\varphi_m^{(\ell)}\}_{m=1}^M$ is computationally unstable. Therefore, we set $\{\varphi_m^{(\ell)}\}_{m=1}^M$ according to the following rule:

$$\begin{aligned} &\varphi_m^{(\ell)} = g_\ell(\tilde{\boldsymbol{x}}_m), \\ &\quad 1 \leq m \leq M, \;\; 1 \leq \ell \leq L. \end{aligned} \tag{S20}$$

We set $\{\varphi_m^{(\ell)}\}_{m=1}^M$ as above because $\tilde{v}_m^{(\ell)}$ is simply the $\ell$-th GP posterior mean at the inducing point $\tilde{\boldsymbol{x}}_m$. To make the GP's posterior mean equal to the ANN's output, $\tilde{v}_\ell^{(m)}$ should be equal to the ANN's (i.e. $g(.)$'s) output at the $m$-th inducing point.

## S2 Algorithm Details

During training, to compute GP's posterior we firstly need to have the $M$ inducing points $\{(\tilde{\boldsymbol{u}}_m^{(\ell)}, \tilde{v}_m^{(\ell)})\}_{m=1}^M$.

**Algorithm S3** Method Init_GPparams

**Input:** Dataset of inducing points $[\tilde{\boldsymbol{x}}_1, ..., \tilde{\boldsymbol{x}}_M]$.

**Output:** List of matrices $\mathbf{U}$, list of vectors $\mathbf{V}$.

*Initialisation* : $\mathbf{U} = list(L)$, $\mathbf{V} = list(L)$.

1: **for** $\ell = 1$ to $L$ **do**
2: $\mathbf{V}[\ell] \leftarrow [g(\tilde{\boldsymbol{x}}_1)[\ell], ..., g(\tilde{\boldsymbol{x}}_M)[\ell])]$.
3: **end for**
4: **for** $\ell = 1$ to $L$ **do**
5: $\mathbf{U}[\ell] \leftarrow [f_\ell(\tilde{\boldsymbol{x}}_1), ..., f_\ell(\tilde{\boldsymbol{x}}_M)]$.
6: **end for**
7: **return** $\mathbf{U}$ and $\mathbf{V}$

**Algorithm S4** Method Explain_ANN

**Input:** Training dataset $ds_train$, and the inducing dataset $ds_inducing$.

**Output:** Kernel-space mappings $[f_1(.), ..., f_L(.)]$, and the other GP parameters $\mathbf{U}$ and $\mathbf{V}$.

*Initialisation* : $\mathbf{U}$, $\mathbf{V}$ $\leftarrow$ Init_GPparams(ds_inducing).

1: **for** $iter = 1$ to $max_iter$ **do**
2: $\boldsymbol{x} \leftarrow randselect(ds_train)$.
3: $\tilde{\boldsymbol{x}} \leftarrow randselect(ds_inducing)$
4: $[f_1(.), ..., f_L(.)] \leftarrow$ Optim_KernMapings($\boldsymbol{x}, \tilde{\boldsymbol{x}}, \mathbf{U}, \mathbf{V}$).
5: $\tilde{\boldsymbol{x}} \leftarrow randselect(ds_inducing)$.
6: **for** $\ell = 1$ to $L$ **do**
7: //update kernel-space representations.
8: $U[\ell][\tilde{\boldsymbol{x}}.index] \leftarrow f_\ell(\tilde{\boldsymbol{x}})$
9: **end for**
10: **end for**
11: **return** $[f_1(.), ..., f_L(.)]$, $\mathbf{U}$, $\mathbf{V}$

It is computationally prohibitive to repeatedly update $\{\tilde{\boldsymbol{u}}_m^{(\ell)}\}_{m=1}^M$ by mapping all $M$ instances to the kernel space as $\tilde{\boldsymbol{u}}_m^{(\ell)} = f_\ell(\tilde{\boldsymbol{x}}_{\boldsymbol{m}})$. On the other hand, as the kernel-space mappings $\{f_\ell(.)\}_{\ell=1}^L$ keep changing during training, we need to somehow track how the inducing points $\{\tilde{\boldsymbol{u}}_m^{(\ell)}\}_{m=1}^M$ change during training. To this end, we consider a matrix whose $m$-th row contains the value of $f_\ell(\tilde{\boldsymbol{x}}_m)$ at some point during training, where $\tilde{\boldsymbol{x}}_m$ is the $m$-th inducing instance. During training, we keep updating the rows of this matrix by feeding mini-batches of instances to $f_\ell(.)$. Note that we have as many GPs as the number of ANN's output heads. Therefore, for each GP we consider a separate matrix containing the representations of the inducing instances in the $\ell$-th kernel space. In Algs.S1, S2, S3, and S4 the variable $\mathbf{U}$ is a list containing all of the the aforementioned matrices. To explain a given ANN, we let the ANN to be fixed and we only train the GPs' parameters. This procedure is explained in Alg.S4. In each iteration, the kernel-mappings are updated according to the objective function of Eq.5 (line 3 of Alg.S2). Afterwards, to make the matrices in $\mathbf{U}$ track the changes in $[f_1(.), ..., f_L(.)]$, we map an inducing instance (or a mini-batch of inducing instances) to the kernel spaces, and we update the corresponding matrices and rows in $\mathbf{U}$ according to the newly obtained kernel-space representations. Updating $\mathbf{U}$ is done in line 8 of Alg.S4. The method in Alg.S1 computes the GPs' posterior means and covariances at any instance like $\boldsymbol{x}$, given the observed inducing points as specified by $\mathbf{U}$ and $\mathbf{V}$. Note that this

**Algorithm S5** Method Efficiently_Compute_AATinvb

**Input:** Matrix $\mathbf{A}$ of size $M \times D$, vector $\boldsymbol{b}$ of size $M \times 1$, and positive scalar $\sigma$.

**Output:** The vector $\mathbf{output} = (\mathbf{A}\mathbf{A}^T + \sigma^2\mathbf{I})^{-1}\boldsymbol{b}$.

1: $\tilde{\mathbf{E}}, \tilde{\boldsymbol{\lambda}} \leftarrow eigendecomp(\mathbf{A}^T\mathbf{A} + \sigma^2\mathbf{I})$.
2: $[\tilde{\boldsymbol{e}}_1, ..., \tilde{\boldsymbol{e}}_D] \leftarrow \tilde{\mathbf{E}}$
3: $[\tilde{\lambda}_1, ..., \tilde{\lambda}_D] \leftarrow \tilde{\boldsymbol{\lambda}}$
4: $[\boldsymbol{e}_1, ..., \boldsymbol{e}_D] \leftarrow [\mathbf{A}\tilde{\boldsymbol{e}}_1, ..., \mathbf{A}\tilde{\boldsymbol{e}}_D]$
5: $[\lambda_1, ..., \lambda_D] \leftarrow [\tilde{\lambda}_1, ..., \tilde{\lambda}_D]$
6: $\mathbf{E} \leftarrow [\boldsymbol{e}_1, ..., \boldsymbol{e}_D]$
7: $\boldsymbol{\Lambda} \leftarrow diagonal(\frac{1}{\lambda_1+\sigma^2}, ..., \frac{1}{\lambda_D+\sigma^2})$
8: $\mathbf{output} \leftarrow \mathbf{E}\boldsymbol{\Lambda}\mathbf{E}^T\boldsymbol{b} + \frac{1}{\sigma^2}(\boldsymbol{b} - \mathbf{E}\mathbf{E}^T\boldsymbol{b})$ //according to //Eq.S21 in supplementary material
9: **return output**

method returns two outputs, because a GP's posterior at $\mathbf{x}$ is a normal distribution described by its mean and variance. In Alg.S1 lines 8 and 9 correspond to the equations of GP posterior (i.e. Eqs. 1 and 2 of the main article). The method in Alg.S1 is used both during training and testing. During training, this method is called whenever ANN's output and GP's posterior are encouraged to be close. During training, according to line 6 of Alg.S1 only the matrix row(s) corresponding to the fed inducing instance(s) are the result of mapping the inducing instance(s) via the kernel-mapping, and all other rows are kept fixed. Line 6 of Alg.S1 allows for computing the gradient of loss with respect to kernel-mappings $[f_1(.), ..., f_L(.)]$. During testing we call Alg.S1 to get the GP's posterior at a test instance like $\boldsymbol{x_{test}}$. Alg.S3 initializes the GP parameters $\mathbf{U}$ and $\mathbf{V}$. For the $\ell$-th GP, the vector $\mathbf{V}[\ell]$ is initialized to the $\ell$-th output head of the ANN at all inducing images. In Alg.S3, the vector $\mathbf{V}[\ell]$ is initialized in line 2. Moreover, for the $\ell$-th GP the matrix $\mathbf{U}[\ell]$ is initialized by mapping all inducing instances to the $\ell$-th kernel-space via the mapping $f_\ell(.)$. In Alg.S3 the matrix $\mathbf{U}[\ell]$ is initialized in line 5. The method in Alg.S3 is called only once before training the GP. For instance, when explaining an ANN in Alg.S4, the initialisation is done once at the beginning of the procedure.

## S2.1 Efficiently Computing Gaussian Process Posterior

Let $\boldsymbol{A}$ be an arbitrary $M \times D$ matrix where $M >> D$. Moreover, let $\boldsymbol{b}$ be a $M$-dimensional vector and let $\sigma$ be a scalar. The computational techniques [10] allow us to efficiently compute:

$$(\mathbf{A}\mathbf{A}^T + \sigma^2\boldsymbol{I}_{M\times M})^{-1}\ \boldsymbol{b}.$$

The idea is that $\mathbf{A}\mathbf{A}^T$ and therefore its inverse are of rank $D$. Therefore, $(\mathbf{A}\mathbf{A}^T)^{-1}$ has $D$ non-zero eigenvalues like $\{\lambda_1, ..., \lambda_D\}$ and the rest of its eigenvalues are zero. Let the corresponding eigenvectors be $\{\boldsymbol{e_1}, ..., \boldsymbol{e_D}\}$. To compute $(\mathbf{A}\mathbf{A}^T)^{-1}\boldsymbol{b}$ we can simply project $\boldsymbol{b}$ to the $D$-dimensional space of the eigenvectors. By doing so, we avoid the $\mathcal{O}(M^3)$ computational complexity. Let $\{\lambda_1, ..., \lambda_D\}$ be the non-zero eigenvalues of $\mathbf{A}\mathbf{A}^T$ and let $\{\boldsymbol{e}_1, ..., \boldsymbol{e}_D\}$ be the corresponding eigenvectors. From linear algebra, it follows that for $\mathbf{A}\mathbf{A}^T + \sigma^2\mathbf{I}_{M\times M}$ the eigenvalues and the eigenvectors are $\{\lambda_1 + \sigma^2, ..., \lambda_D + \sigma^2, \sigma^2, ..., \sigma^2\}$ and $\{\boldsymbol{e_1}, ..., \boldsymbol{e_D}\}$, respectively. Note that $M - D$ eigenvectors are added all of which are equal to $\sigma^2$. Similarly, from linear algebra it follows that for the inverse of $\mathbf{A}\mathbf{A}^T + \sigma^2\mathbf{I}_{M\times M}$ the eigenvalues and eigenvectors are $\{\frac{1}{\lambda_1+\sigma^2}, ..., \frac{1}{\lambda_D+\sigma^2}, \frac{1}{\sigma^2}, ..., \frac{1}{\sigma^2}\}$ and $\{\boldsymbol{e_1}, ..., \boldsymbol{e_D}, \boldsymbol{e_{D+1}}, ..., \boldsymbol{e_M}\}$ respectively. Note that although there are $M$ eigenvectors, only the first $D$ eigenvectors appear in our computations. More precisely, let $\mathbf{E} \in \mathbb{R}^{M\times D}$ be a matrix whose columns are $\{\boldsymbol{e_1}, ..., \boldsymbol{e_D}\}$. Let $\boldsymbol{\Lambda}$ be a diagonal matrix whose diagonal is formed by $\{\frac{1}{\lambda_1+\sigma^2}, ..., \frac{1}{\lambda_D+\sigma^2}\}$. In the space of the $D$ eigenvectors the linear transformation on any vector like $\boldsymbol{b}$ is equal to $\mathbf{E}\boldsymbol{\Lambda}\mathbf{E}^T\boldsymbol{b}$, meaning that multiplication by $\mathbf{E}^T$ transforms $\boldsymbol{b}$ to the space of the $D$ eigenvectors, multiplication by $\boldsymbol{\Lambda}$ performs the transformation in that space, and multiplication by $\mathbf{E}$ transforms the result back to the original space. The $(M - D)$ eigenvalues that correspond to the rest of the eigenvectors are all the same and are equal to $\frac{1}{\sigma^2}$. Therefore, there is no need to project $\boldsymbol{b}$ to the space of the $(M - D)$ eigenvectors because the linear transformation in that space is simply a scaling by $\frac{1}{\sigma^2}$. All in all, we have that

$$\left(\mathbf{A}\mathbf{A}^T + \sigma^2\mathbf{I}_{M\times M}\right)^{-1}\boldsymbol{b} \;=\; \mathbf{E}\boldsymbol{\Lambda}\mathbf{E}^T\boldsymbol{b} + \frac{1}{\sigma^2}(\boldsymbol{b} - \mathbf{E}\mathbf{E}^T\boldsymbol{b}). \quad \text{(S21)}$$

Complexity of computing the right-hand-side of Eq.S21 is way lower than the $\mathcal{O}(M^3)$ requirement of the standard matrix inversion. We borrowed more computational ideas from the work on fast spectral clustering [10]. To compute the first $D$ eigenvlaues and eigenvectors of $\mathbf{A}\mathbf{A}^T$, we worked with the $D$-by-$D$ matrix $\mathbf{A}^T\mathbf{A}$ rather than the $M$-by-$M$ matrix $\mathbf{A}\mathbf{A}^T$ (recall that $D << M$), because given the eigenvalues and eigenvectors of $\mathbf{A}^T\mathbf{A}$, those of $\mathbf{A}\mathbf{A}^T$ are easily computable [10]. The procedure is explained in Alg.S5. In Alg.S5, lines 1-3 compute the eigenvalues/vectors of the matrix $\mathbf{A}^T\mathbf{A}$. Afterwards, lines 4 and 5 compute the first $D$ eigenvalues/vectors of $\mathbf{A}\mathbf{A}^T$ using those of $\mathbf{A}^T\mathbf{A}$. Finally, line 8 computes $(\mathbf{A}\mathbf{A}^T + \sigma^2\mathbf{I})^{-1}\boldsymbol{b}$ according to the right-hand-side of Eq.S21. To make the computations faster, we made use of the following equation $\mathbf{A}\mathbf{A}^T = \sum_m \mathbf{A}[m,:]\mathbf{A}[m,:]^T$, where $\mathbf{A}[m,:]$ is the $m$-th row of the matrix $\mathbf{A}$. Thanks to this equation, we compute $\mathbf{A}\mathbf{A}^T$ only once at the beginning of the training. Afterwards, as each mini-batch alters only some rows of $\mathbf{A}$, we update the previously computed $\mathbf{A}\mathbf{A}^T$ by considering only the effect of the modified rows.

## S2.2 Computing Pixel Contributions to the Similarity

We first explain the idea of CAM [34], afterwards we modify it for the architectures of our kernel modules. Let the kernel mapping $f(.)$ be a convolutional neural network that produces a volumetric map of size $C \times H \times W$ followed by a spatial average pooling that produces the $C$-dimensional vector in the kernel-space. In this case, $\mathcal{K}(\boldsymbol{x}_1, \boldsymbol{x}_2)$ is as follows:

$$\begin{aligned}\mathcal{K}(\boldsymbol{x}_1, \boldsymbol{x}_2) &= f(\boldsymbol{x}_1)^T\ f(\boldsymbol{x}_2) \\ &= \Big(\sum_{i=1}^{H}\sum_{j=1}^{W}\boldsymbol{z}_{ij}^{(1)}\Big)^T\Big(\sum_{k=1}^{H}\sum_{\ell=1}^{W}\boldsymbol{z}_{k\ell}^{(2)}\Big) \\ &= \sum_{i=1}^{H}\sum_{j=1}^{W}\sum_{k=1}^{H}\sum_{\ell=1}^{W}\big(\boldsymbol{z}_{ij}^{(1)^T}\boldsymbol{z}_{k\ell}^{(2)}\big),\end{aligned} \quad \text{(S22)}$$

where $\boldsymbol{z}^{(1)}$ and $\boldsymbol{z}^{(2)}$ are the volumetric maps of size $C \times H \times W$ and the indices $(i, j)$ and $(k, \ell)$ index the spatial locations over the volumetric maps. The last term in Eq.S22 shows that the total similarity $\mathcal{K}(\boldsymbol{x}_1, \boldsymbol{x}_2)$ is the sum of the contributions from each pair of positions $(i, j)$ on $\boldsymbol{x}_1$ and $(k, \ell)$ on $\boldsymbol{x}_2$. To compute the contribution of a specific location like $(i, j)$ on $\boldsymbol{x}_1$, we sum up the contributions of $(i, j)$ on $\boldsymbol{x}_1$ and all possible locations $\{(k, \ell)\}_{k=1}^{H}{}_{\ell=1}^{W}$ on $\boldsymbol{x}_2$.

The kernel-mappings that we used have a slightly different architecture than a volumetric map followed by spatial average pooling. Our kernel mappings produce a volumetric map of size $C \times H \times W$ followed by a spatial average pooling that produces a $C$-dimensional vector. Afterwards, the resulting vector is divided by its $\ell_2$-norm to produce a vector of norm 1. Consequently, this vector of norm 1 is fed to a leaky ReLU layer that produces the final kernel-space representation $f(\boldsymbol{x})$. For this architecture the pixel contributions can be computed according to an equation similar to Eq.S22 as follows. Our kernel mappings produce the volumetric map $\boldsymbol{z}$ of size $C \times H \times W$ followed by a spatial average pooling that produces the $C$-dimensional vector $\boldsymbol{a}$:

$$\boldsymbol{a} = \sum_{i=1}^{H} \sum_{j=1}^{W} \boldsymbol{z}_{ij}. \tag{S23}$$

Afterwards, the resulting vector is divided by its $\ell_2$-norm to produce the vector $\boldsymbol{b}$ of norm 1:

$$\boldsymbol{b} = [\frac{a_1}{||\boldsymbol{a}||_2}, \ \dots, \frac{a_C}{||\boldsymbol{a}||_2}]. \tag{S24}$$

Consequently, this vector of norm 1 is fed to a leaky ReLU layer that produces the final kernel-space representation $f(\boldsymbol{x})$:

$$f(\boldsymbol{x}) = leakyReLU(\boldsymbol{b}). \tag{S25}$$

We begin with simplifying Eq.S25. The leaky ReLU activation function multiplies the input by a constant and this constant depends on the sign of the input. Therefore, applying the leaky ReLU activation is equivalent to multiplication by a diagonal matrix $\boldsymbol{\Lambda}$. Therefore,

$$f(\boldsymbol{x}) = \boldsymbol{\Lambda}\boldsymbol{b}. \tag{S26}$$

Let $\boldsymbol{x}_1$ and $\boldsymbol{x}_2$ be two images, and $\boldsymbol{z}^{(1)}$ and $\boldsymbol{z}^{(2)}$ be the corresponding volumetric maps. We have that

$$\begin{aligned} \boldsymbol{a}^{(1)} &= \sum_{i=1}^{H} \sum_{j=1}^{W} \boldsymbol{z}_{ij}^{(1)}, \\ \boldsymbol{a}^{(2)} &= \sum_{k=1}^{H} \sum_{\ell=1}^{W} \boldsymbol{z}_{k\ell}^{(2)}. \end{aligned} \tag{S27}$$

And

$$\begin{aligned} \boldsymbol{b}^{(1)} &= [\frac{a_1^{(1)}}{||\boldsymbol{a}^{(1)}||_2}, \ \dots, \frac{a_C^{(1)}}{||\boldsymbol{a}^{(1)}||_2}], \\ \boldsymbol{b}^{(2)} &= [\frac{a_1^{(2)}}{||\boldsymbol{a}^{(2)}||_2}, \ \dots, \frac{a_C^{(2)}}{||\boldsymbol{a}^{(2)}||_2}]. \end{aligned} \tag{S28}$$

And

$$\begin{aligned} f(\boldsymbol{x}^{(1)}) &= \boldsymbol{\Lambda}^{(1)}\boldsymbol{b}^{(1)}, \\ f(\boldsymbol{x}^{(2)}) &= \boldsymbol{\Lambda}^{(2)}\boldsymbol{b}^{(2)}. \end{aligned} \tag{S29}$$

Now we simplify the similarity $\mathcal{K}(\boldsymbol{x}_1, \boldsymbol{x}_2)$:

$$\begin{aligned} \mathcal{K}(\boldsymbol{x}_1, \boldsymbol{x}_2) &= \big(\boldsymbol{\Lambda}^{(1)}\boldsymbol{b}^{(1)}\big)^T \big(\boldsymbol{\Lambda}^{(2)}\boldsymbol{b}^{(2)}\big) \\ &= \big(\boldsymbol{\Lambda}^{(1)^T}\boldsymbol{\Lambda}^{(2)}\big)\big(\boldsymbol{b}^{(1)^T}\boldsymbol{b}^{(2)}\big) \\ &= \frac{\big(\boldsymbol{\Lambda}^{(1)^T}\boldsymbol{\Lambda}^{(2)}\big)}{||\boldsymbol{a}^{(1)}||_2 \ ||\boldsymbol{a}^{(2)}||_2} \ (\sum_{i=1}^{H}\sum_{j=1}^{W} \boldsymbol{z}_{ij}^{(1)})^T (\sum_{k=1}^{H}\sum_{\ell=1}^{W} \boldsymbol{z}_{k\ell}^{(2)}) \\ &= \frac{\big(\boldsymbol{\Lambda}^{(1)^T}\boldsymbol{\Lambda}^{(2)}\big)}{||\boldsymbol{a}^{(1)}||_2 \ ||\boldsymbol{a}^{(2)}||_2} \ \sum_{i=1}^{H}\sum_{j=1}^{W}\sum_{k=1}^{H}\sum_{\ell=1}^{W} \big(\boldsymbol{z}_{ij}^{(1)^T} \boldsymbol{z}_{k\ell}^{(2)}\big). \end{aligned} \tag{S30}$$

Indeed, as the used architecture for kernel-mappings is slightly different than producing a volumetric map followed by spatial average pooling, instead of Eq.S22, we used Eq.S30 that we derived above.

## S3 Examining Faithfulness of GPs to ANNs

In Sec.4.1. of the main article, we examined the faithfulness of the found GPs to their corresponding ANNs. In this section we provide more information and insights about the analogy between the GPs found by our proposed GPEX and their corresponding ANNs. Figs.S2, S4, and S6 illustrate the scatter plots of ANN-GP outputs on Cifar10 [15], MNIST [6], and Kather [12], respectively. These scatter plots are obtained on the testing set which has been invisible to the proposed GPEX. Note that in Figs.S2, S4, and S6 each ANN's output head and its corresponding GP have a seprate scatter plot.

In the main article, we discussed that our proposed GPEX is applicable to any subcomponent of a pipeline. To verify this, in Sec.4.1. of the main article we applied the proposed GPEX to attention subcomponents of classifier pipelines. Here we provide more information about the faithfulness of the found GPs to the attention subcomponents. Figs.S3, S5, and S7 illustrate the scatter plots for attention submodules and their corresponding GPs.

For Cifar10 [15] in Fig.S3, each attention mask is 3 x 3 and we have 9 scatter plots. According to Fig.S3, in attention masks some output heads like head 1, head 2, and head 3 do not turn on for any instnace (the values change around -2, and sigmoid of -2 is a small number). Therefore, in Fig.3 of the main article we have excluded the attention heads which are always off. Similarly, for MNIST [6] and Kather [12] we see some attention heads are always off in Figs.S5 and S7, and we have excluded those heads in Fig.3 of the main article.

So far we reported corelation coeffients (Fig.3 of the main article) and scatter plots (Figs.S2, S3, S4, S5, S6, S7) to examine the faithfulness of GPs to their corresponding ANNs. To get more insights, we selected mini-batches of testing instances and fed each mini-batch to both ANN and corresponding GPs. The output from ANN (and simmilarly GPs) is a matrix of shape $batchsize \times D_v$, where $D_v$ is the number of output heads from the ANN. Ideally, we should get two identical $batchsize \times D_v$ matrices for each mini-batch, because the GPs are supposed to be faithfull to ANNs. Figs. S59, S60, S61, and S62 illustrate the heatmaps for four randomly fed mini-batches from Cifar10 [15], MNIST [6], Kather [12], and DogsWolves [30], respectively. According to Figs. S59, S60, S61, and S62 the outputs from GPs almost match those from their corresponding ANNs. In Figs. S59, S60, S61, and S62 the red rectangles show the test instances for which the GP's decision (i.e. the class with the highest score) does not match the ANN's decision. According to Figs. S59, S60, S61, and S62 the disagreement between GPs prediction and ANN prediction mostly happens when either some output activations are very close to one another or all activations are close to zero. This is consistent with the scatter plots of Figs. S2, S4, and S6 in which the scatters are slightly dispersed for intermediate values. Tab. S1 reports the test accuracy of the ANNs and their corresponding GPs. We see that GPs' accuracies are slightly lower than those of the corresponding ANNs. Figs. S59, S60, S61, and S62 provide insights about how this small disagreement can be potentially solved in future research by, e.g., preventing the ANN from having near-zero activations or having output heads which are very close to one another. We repeated the experiment with 5 different random splits and reported the results in Fig. S67. According to Fig. S67 the correlation coefficients are high for different training/testing splits.

We repeated the experiment of Figs. S59, S60, S61, and S62 for the attention submodules and corresponding GPs. Reults are provided in Figs. S63, S64, and S65. According to Figs. S63, S64, and S65 our proposed GPEX has found GPs which are faithful to the attention submodules.

## S4 Explaining ANNs' Decisions

In Sec.4.2 of the main article we applied our proposed method (i.e. Alg.S4) to some ANN classifiers. Afterwards, we explained the decisions made by the ANNs via the GPs and the kernel-spaces that our proposed GPEX has found. Here we are going to provide more explanations for ANNs' decisions on more testing instances.

We explain the decision made for a test instance like $\boldsymbol{x}_{test}$ as follows. We consider the GP and the kernel-space that correspond to the ANN's head with maximum value (i.e. the ANN's head that relates to the predicted label). Consequently, among the instances in the inducing dataset, we find the 10 closest instances to $\boldsymbol{x}_{test}$, like $\{\boldsymbol{x}_{i1}, \boldsymbol{x}_{i2}, ..., \boldsymbol{x}_{i10}\}$. Intuitively the ANN has labeled $\boldsymbol{x}_{test}$ in that way because it has found $\boldsymbol{x}_{test}$ to be similar to $\{\boldsymbol{x}_{i1}, \boldsymbol{x}_{i2}, ..., \boldsymbol{x}_{i10}\}$. Besides finding the nearest neighbours, we provide explanation as to why $\boldsymbol{x}_{test}$ and an instance like $\boldsymbol{x}_{ij}, 1 \leq j \leq 10$ are considered similar by the model. The procedure is explained in Sec.S2.2.

For MNIST digit classification, some test instances and nearest neighbours in training set are shown in Figs.S8, S9, S10, and S11. In these figures each row corresponds to a test instance. The first column depicts the test instance itself and columns 2 to 11 depict the 10 nearest neighbours. According to rows 2 and 3 of Fig.S8, the classifier has labeled the two images as digit 1 because it has found 1 digits with similar inclinations in the training set. We see the model has also taken the inclination into account for the test instances of rows 8 and 9 of Fig.S8 and rows 1, 2, and 3 of Fig.S11. In Fig.S8, according to rows 4, 5, and 6 the test instances are classified as digit 2 because 2 digits with similar styles are found in the training set. We see the model has also taken the style into account for the test instances of rows 7, 8, 9, 10, 11 of Fig.S8 and rows 1, 2, 3, 4, 5, 6, 7, and 8 of Fig.S9. For instance, the test instance in row 1 of Fig.S9 is a 4 digit with a short tail and the two nearest neighbours are alike. Or for the test instances in rows 5, 6, 7, and 8 of Fig.S10 the test instances have incomplete circles in the same way as their nearest neighbours.

Figs.S12, S13, S14, S15, S16, S17, S18, and S19 illustrate sample explanations for similarities. For instance row 1 of Fig.S12 illustrates a test instance as well as the 10 nearest neighbours. The second row of Fig.S12 highlights to what degree each region of each nearest neighbour contributes to its similarity to the test instance. The third row of Fig.S12 illustrates to what degree each region of the test instance contributes to its similarity to each of the nearest neighbours. For example, according to rows 1, 2, and 3 of Fig.S17 the cross pattern of the 8 digits have had a significant contribution to their similarities. For MNIST [6], more similarity explanations are provided in Figs.S12, S13, S14, S15, S16, S17, S18, and S19.

Figs.S36, S37, S38, S39, S40, S41, S42, S43, S44, S45, S46, S47, S48, and S49 illustrate some sample explanations for Cifar10 [15]. Like before, each row corresponds to a test instance, the first column depicts the test instance itself and columns 2 to 11 depict the 10 nearest neighbours. In rows 8, 9, 10, and 11 of Fig.S44 and rows 1 and 2 of Fig.S45, the test instances are captured from horses' heads from closeby, and the nearest neighbours are alike. However, in rows 3, 4, 5, 6, 7, 8, and 9 of Fig.S45 the test images are taken from faraway and the found similar training images are also taken from faraway. Intuitively, as the classifier is not aware of 3D geometry, it finds training images which are captured from the same distance. We constantly observe this pattern in more explanations: row 6, 7, 8, 9, 10, and 11 in Fig.S39, all rows of Fig.S40, rows 1, 2, 6, 7, 8, 9, 10 and 11 of Fig.S41, rows 1, 7, 8, 9, 10, and 11 of Fig.S42, rows 1, 2, 3, 4, 5, 6, 7, and 8 of Fig.S43, all rows of Fig.S45 and rows 1-10 of Fig.S46.

Animal faces tend to be recognized by similar faces. We see this pattern in rows 2, 3, 4, 5 and 6 of Fig.S40, rows 6, 7, 8, and 9 of Fig.S41, rows 7 and 8 of Fig.S43, rows 8, 9, 10, and 11 of Fig.S44 and rows 1, 2, 10, and 11 of Fig.S45. To classify airplanes, the model has taken into account the inclination. For instance, in Fig.S36 the model has taken into account whether the airplane is taking off (rows 1, 8, 9, 10, and 11 of Fig.S36), flying straight (rows 2 and 4 of Fig.S36) or is inclined downwards (rows 3, 5, 6 and 7 of Fig.S36). Furthermore, the bat-like airplanes are recognized by the model because similar bat-like airplanes are found in the training set, as we see in rows 1, 2, 3, 4, 5, 6 and 7 of Fig.S37. Cessnas are often classified by finding cessnas in the training set, as we see in rows 8, 9 and 10 of Fig.S37 and row 1 of Fig.S38.

Since the classifier has no knowledge about 3D geometry, it tends to find training instances which are captured from the same angle as the test instance, as we see in rows 6, 7, 8, 9, 10 and 11 of Fig.S39, rows 7, 8, 9, 10 and 11 of Fig.S42, rows 9, 10 and 11 of Fig.S43, rows 1, 2, 3, 4, 5, 6 and 7 of Fig.S44, row 11 of Fig.S46, all rows of Fig.S47, and rows 1, 2, 3, 4, 5, 6, 7, and 8 of Fig.S48. In rows 3, 4, and 5 of Fig.S41 it seems the model takes into account the ostrich-like shape of the animal. In rows 2, 3, 4, and 6 of Fig.S42 the horns seem to have an effect. In rows 6, 7, 8, and 9 of Fig.S45, we see the model have made use of the riders to classify the test instances as horse. According to rows 1, 2, 3, 4, 5, 6, 7, 8, 9, and 10 of Fig.S46, the model distinguishes between medium sized ships and huge cargo ships. To classify firefighter trucks, model tends to find similar firefighter trucks in the training set, as we see in rows 10 and 11 of Fig.S47, and rows 1, 2, 3, and 4 of Fig.S48. For some testing instances, the model finds training instances which are almost identical to the test instance, as we see in rows 2 and 5 of Fig.S40, row 7 of Fig.S42, row 8 of Fig.S43, and row 8 of Fig.S48.

In rows 2, 4, 5, 6, 7, 8, 9, 10, and 11 of Fig.S38 it seems the classifier has taken into account the blue background. We used the proposed GPEX to explain as to why some testing instances get missclassified. Rows 9, 10, and 11 of Fig.S48 and all rows of Fig.S49 illustrate some instances which are misclassified. For instance in row 10 of Fig.S48 the test image shows an airplane, but the model has classified it as a cat, because it is similar to the cat faces shown in columns 2 to 11 (can you find the cat face in the airplane image?). In row 11 of Fig.S48, the car is classified as truck partially because it very similar to the truck at column 2. In row 2 of Fig.S49, the deer is classified as horse partially because it is very similar to the training image shown in column 2. In row 3 of Fig.S49, we hypothesize the dog is classified as cat because the model has taken into account the cyan and red colors in the background. In this case, adding dog images with cyan and red background may make the model classify this test instance correctly. In rows 5 and 6 of Fig.S49, the model correctly understands the test images are similar to some faces from other animals, but it fails to find similar frog faces in the training set. In this case, adding more images from frog faces may solve this issue. In row 7 of Fig.S49 the horse is classified as airplane, because the model thinks the horse image is similar to some airplane training images which are taking off. Interestingly, the jumping frog in column 4 has been considered similar to the horse image. It seems having inclined edges (due to taking off, jupming) has contributed to the similarities, and therefore the model has incorrectly classified the horse as airplane.

For the DogsWolves dataset [30] the explanations are provided in Figs.S25-S35. According to rows 10, 11, and 12 of Fig.S29, the red ball in the dog's mouth (as highlighted in row 12 of Fig.S29) has the most contribution to the similarities. According to row 2 of Fig.S29, patterns like human hand in column 4 or woody or pink background in columns 8, 10, and 11 are highlighted in nearest neighbours while in the test insntace (row 3 of Fig.S29) the red ball at the bottom right is highlighted. Our explanations consistently show that the model detects dogs by any pattern that rarely appear in a wolf image. For instance in rows 4-6 of Fig.S29, according to row 4 humans in columns 3, 9, and 11, and dog collars or costumes in columns 4, 5, 6, and 10, and the brick wall in the test instance (row 6 of Fig.S29) are used by the model. According to rows 9, 12, and 15 of Fig.S29, the flowers, the red ball in the dogs mouth, and the children are used by the model, respectively. According to rows 3, 6, 9, 12, and 15 of Fig.S30, the red rope, the dog's color, red patterns, brown background and brown background are used by the model, respectively. According to rows 3, 6, 9, 12, and 15 of Fig.S31, brown background, human, brown background, the red wallet, and the pink ball are used by the model, respectively. According to rows 3, 6, 9, 12, and 15 of Fig.S32, the child, pink pillow, brown color, orange background, and red blood are used by the model, respectively. Note that in Fig.S32 the last two instances (rows 10-15) are misclassified. In Fig.S33 all test instances get misclassified. According to rows 3, 6, 9, 12, and 15 of Fig.S33, colorful background, the red object attached to the wolf, background, white background, and dark-green background are used by the model, respectively. Figs.S34 and S35 illustrate more explanations. For instance, according to row 6 of Fig.S34 and row 12 of Fig.S35, the test instances are misclassified due to their dark background. Moreover, according to rows 3, 6, and 15 of Fig.S35, the test instances are misclassified due to their background. All in all, our explanations reveal that for the DogsWolves dataset [30] the model makes use of potentially incorrect clues to label instances. This is not surprising because the dataset has only 2000 images.

For Kather dataset [12], some explanations are shown in Figs.S20, S21, S22, S23, and S24. Like before, in Figs.S20 and S21 each row corresponds to a test instance, the first column depicts the test instance itself and columns 2 to 11 depict the 10 nearest neighbours. In row 1 of Fig.S20, the test image is classified as fat tissue. According to rows 1, 2, and 3 of Fig.S22, the similarity is due to the wire mesh formed by cellular membranes described by our expert pathologist. Row 13 of Fig.S22 shows cancer-associated stroma which is classified correctly. All 10 nearest neighbours are also cancer-associated stroma. Distinguishing between cancer-associated stroma and normal smooth muscle is a challenging task even for expert pathologists, and they often look similar. According to rows 13, 14, and 15 of Fig.S22, the model cares about both the stroma and nuclei. In row 7 of Fig.S22, the test image is correctly classified as lymphocytes. For a pathologist they represent scattered well defined round structures. According to rows 7, 8, and 9 of Fig.S22, the model considers all regions which matches the way pathologists recognize lymphocytes. In rows 1, 2 and 3 of Fig.S23 and rows 1, 2, and 3 of Fig.S24, for the two test instances the model takes into account nuclei which is not the same way that a pathologists would classify the images. We hypothesize that for the model it is easier to extract features from nuclei than to consider the context information. Because even small changes in nuclei is easily measurable by the model while it is not easily noticeable by human eyes. The test image in row 7 of Fig.S24 gets missclassified. According to rows 7, 8, and 9 of Fig.S24 the artificial white holes are considered as glandular lumens by the model and that explains why the test instance gets misclassified. The test image in row 10 of Fig.S23 gets misclassified. According to rows 10, 11, and 12 of Fig.S23, the test image is smooth muscle. But it contains artifactual white spaces (retractions) like the found similar training instances. This make the model think the test image is similar to debris images that contain artifactual white spaces. For Kather dataset [12], more sample explanations are provide in Figs.S22, S23, and S24.

# S5 Practical Details and Parameter Settings

In this section we discuss some practical details which we have not yet discussed in this paper. Moreover, we provide the exact parameter settings that we used throughout our experiments. As explained in Sec.3 of the main article, there are $L$ kernel mappings that we denoted by $[f_1(.), f_2(.), ..., f_L(.)]$. One can implement this kernel mappings by, e.g., considering $L$ independent CNNs. However, doing so dramatically increases the computation cost. Therefore, we modeled the $L$ mappings by a common ResNet-50 [9] backbone. After the common backbone, we placed $L$ branches. Each branch has two convolutional layers followed by global spatial average pooling that produce a vector. Each branch ends with an L2 normalizer layer (that sets the L2-norm of the vector to 1) followed by a leaky-ReLU layer. During our experiments, we noticed that the L2-normalization layer and the final leaky-ReLU layer are essential. Without the L2 normalization layer, the vectors in the kernel-space can have arbitrarily-small or arbitrarily-big elements, and this makes the training unstable. We included the last leaky-ReLU layer, because according to GP posterior mean formula, vectors in the kernel-space go through a linear transformation. Therefore, without the last leaky-ReLU layer, the pipeline would have two consequtive linear layers. Throughout our experiments, we set the output of each branch (i.e. vectors in the kernel-space of each GP) to be 20-dimensional.

As illustrated by Fig.S66 (to be discussed in Sec.S7), we need to make the inducing dataset as large as possible. Therefore, throughout our experiments we selected the whole training dataset as the inducing dataset. Unlike training instances, we didn't apply data-augmentation on inducing instances. By doing so, the training dataset and the inducing dataset will have very similar instances. This causes a difficulty that we are going to discuss in this part. The kernel-mappings $[f_1(.), ..., f_L(.)]$ are trained according to Alg.S4. After selecting an instance like $\boldsymbol{x}$ from the training dataset, $\boldsymbol{x}$ is actually the augmented version of an inducing instance like $\tilde{\boldsymbol{x}}_m$. Indeed, we have that $\boldsymbol{x} = DataAug(\tilde{\boldsymbol{x}}_m)$. Because $\boldsymbol{x}$ and $\tilde{\boldsymbol{x}}_m$ are very similar, they will be very close to one another in the kernel-spaces regardless of what parameters $[f_1(.), ..., f_L(.)]$ have. Therefore, regardless of the kernel-mappings, the GP-mean will match the ANN value at $\boldsymbol{x}$, and there will be no training signal for the kernel-mappings $[f_1(.), ..., f_L(.)]$. Note that in this case the GPs match the ANNs only on training instances, and the analogy does not generalize to testing instances. To avoid this issue, we optimized the GP-ANN analogy (i.e. the objective in Eq.5 of the main article) on instances like $\lambda\boldsymbol{x}_i + (1-\lambda)\boldsymbol{x}_j$, where $\boldsymbol{x}_i$ and $\boldsymbol{x}_j$ are two instances randomly selected from the training set and $\lambda$ is a scalar uniformly selected from $[-1, 2]$.

When applying our proposed GPEX we used Adam optimizer [14]. Although the AMSGrad version of this optimizer is often recommended, for our proposed GPEX we noticed the Adam optimizer [14] without AMSGrad works the best. For explaining classifier ANNs, we used a learning-rate of 0.0001 while for explaining the attention submodules we used a learning rate of 0.00001. On a RTX3090 GPU, the experiments took around 3 days for the 4 image datasets and around 2 hours for the biological dataset. We ran Alg. S4 for 200 epochs. Afterwards, we ran line 8 of Alg. S4 for the inducing dataset. Afterwards, we continued Alg.S4 for 200 more epochs and repeating line 8 of Alg.S4 10 times instead of once.

# S6 Qualitative Comparision of GPEX and Representer Point Selection

We qualitatively compared the explanations of our proposed GPEX to those of representer point selection [33]. The results are provided in Figs.S50, S51, S52, S53, S54, S55, S56, and S57. In each triple, the first row shows the test instance and the 10 nearest neighbours found by our proposed GPEX. The second row shows the 10 nearest neighbours selected by representer point selection [33]. The third row shows the 10 nearest neighbours according to the kernel-space of representer point selection [33]. Representer point selection [33] assigns an importance weight to each training instance. Therefore, some training instances tend to appear as nearest neighbours regardless of what the testing instance is. We see this behaviour in rows 2, 5, 8, 11, and 14 of Figs.S50-S57. However, for our proposed GPEX the nearest neighbours can freely change for different test instances. We see this behaviour in rows 1, 4, 7, 10, and 13 of Figs.S50-S57. If we ignore the importance weights in representer point selection [33], the aforementioned issue in that method happens less frequently, as we see in rows 3, 6, 9, 12, and 15 of Figs.S50-S57. However, the issue is that without the importance weights, the explainer model in representer point selection [33] will not be faithful to the ANN itself.

# S7 Parameter Analysis

To analyze the effect of the number of inducing points (i.e. the variable $M$ in Sec. S2) we applied the proposed GPEX to the classifier CNN that we trained on Cifar10 dataset [15] in Sec. 4.1 of the main manuscript. This time, instead of considering all training instances as the inducing dataset, we randomly selected some training instances. In Fig. S66, the horizontal axis shows the size of the inducing dataset. For each size, we repeated the experiment 5 times (i.e. split 1-5 in Fig. S66). According to Fig. S66, to obtain GPs which are faithful to ANNs one needs to have a lot of inducing points. This highlights the importance of the scalability techniques that we used (the computational techniques are elaborated upon in Sec. S2.1 of supplementary material). Another intriguing point in Fig. S66 is that if we are to select a few training images as inducing points, the correlation coefficients highly depend on which instances are selected. More precisely, Fig. S66 suggests that one may be able to reach high correlation coefficients by selecting a few inducing points from the training set in a subtle way.

So far we analyzed the effect of the size of inducing dataset. Here we analyze two other important factors: the width of the second last layer and number of epochs for which the ANN has been trained. On Cifar10 [15] we trained ANNs with different number of neurons in the second last layer and we analyzed the ANN at different checkpoints during training (10, 50, 100, 150, and 200 epochs). The result is shown in Fig.S58. According to Fig.S58, increasing the width of the second last layer increases the correlation coefficients. However, as illustrated by Fig.S58, the proposed GPEX can achieve almost perfect match even when the second last layer of the ANN is not wide. Moreover, according to Fig.S58, our proposed GPEX can reach high correlation coefficients even when the ANN's parameters are not a local minimum of the classification loss. This empirical results show that most theoretical results like requiring all layers of the ANN to be wide [5], or requiring the ANN to be optimized on a loss [20] may not be necessary.

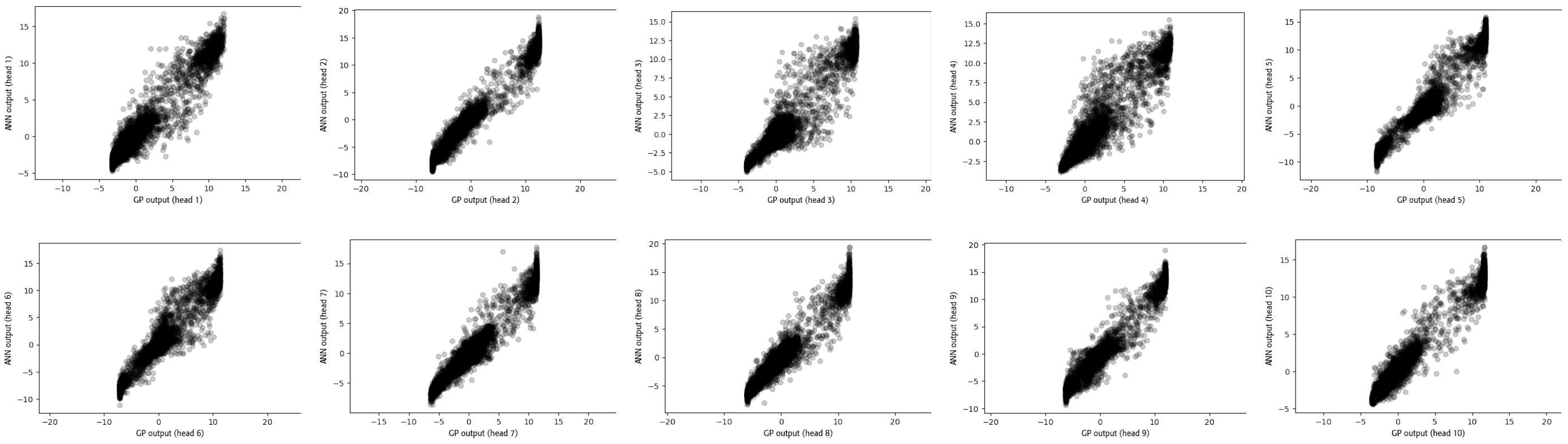

Fig. S2: Scatters for Cifar10 (classifier).

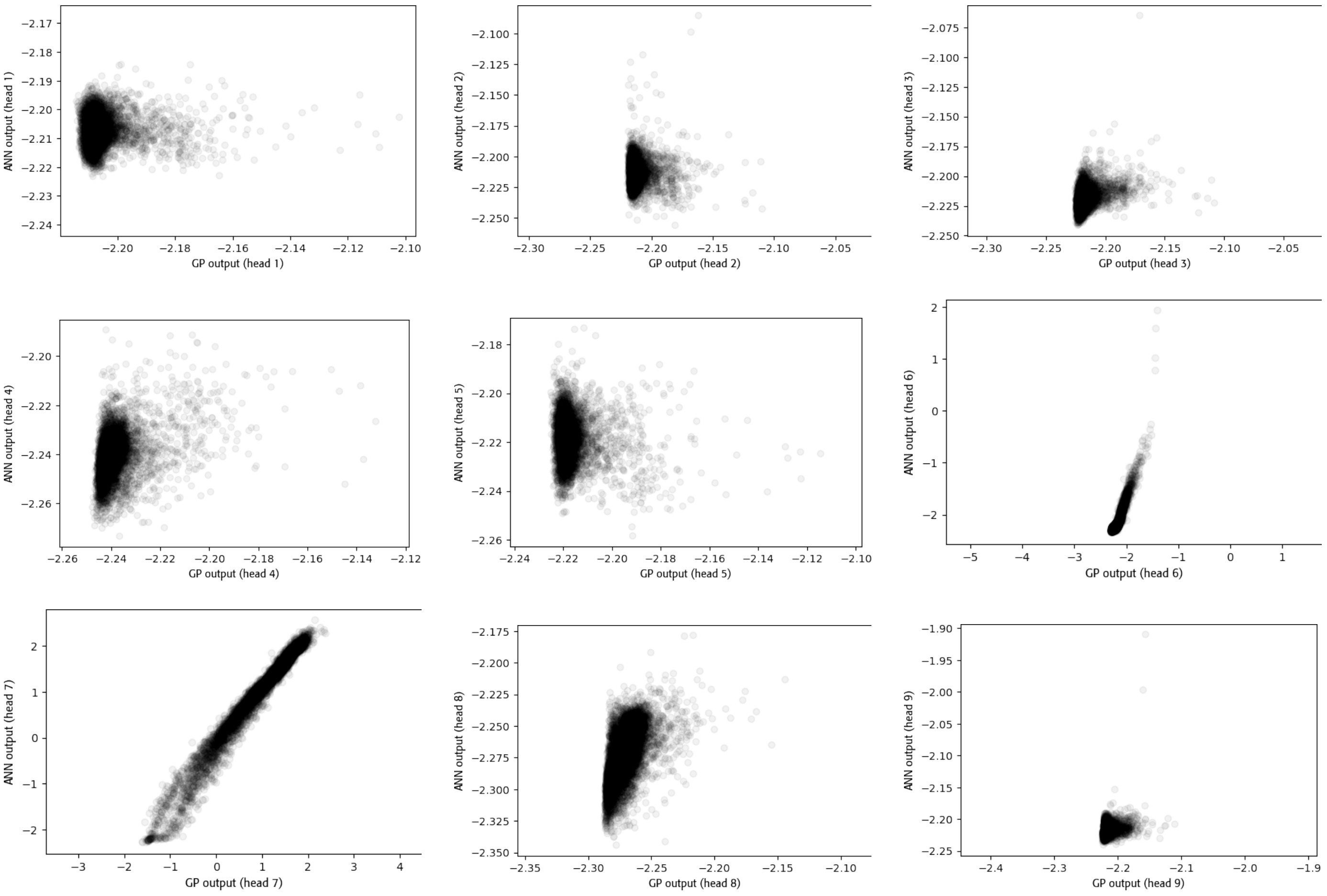

Fig. S3: Scatters for Cifar10 (attention).

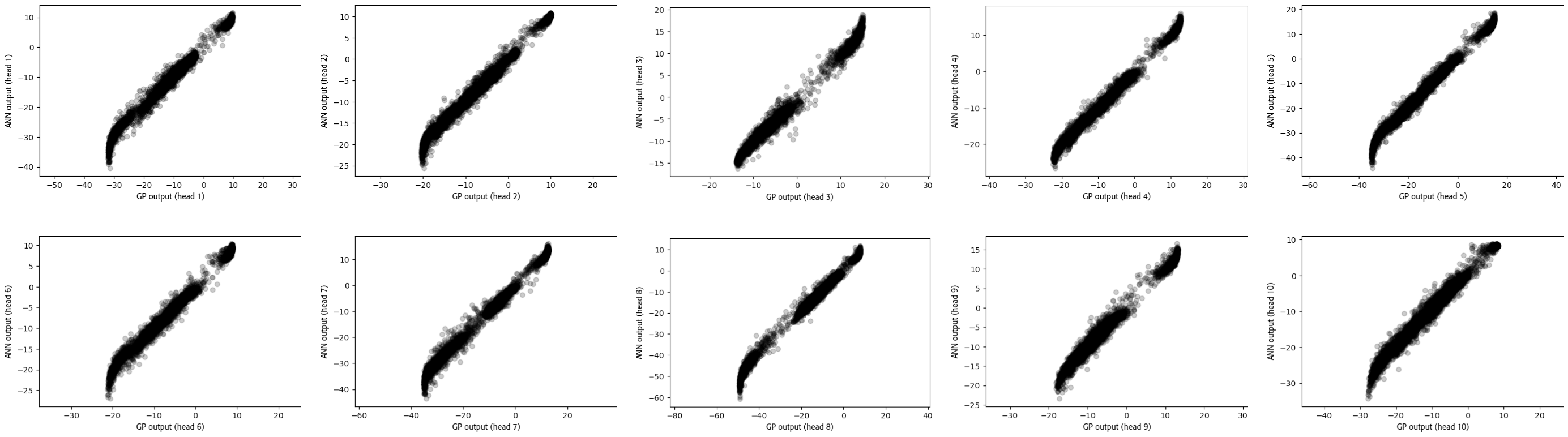

Fig. S4: Scatters for MNIST (classifier).

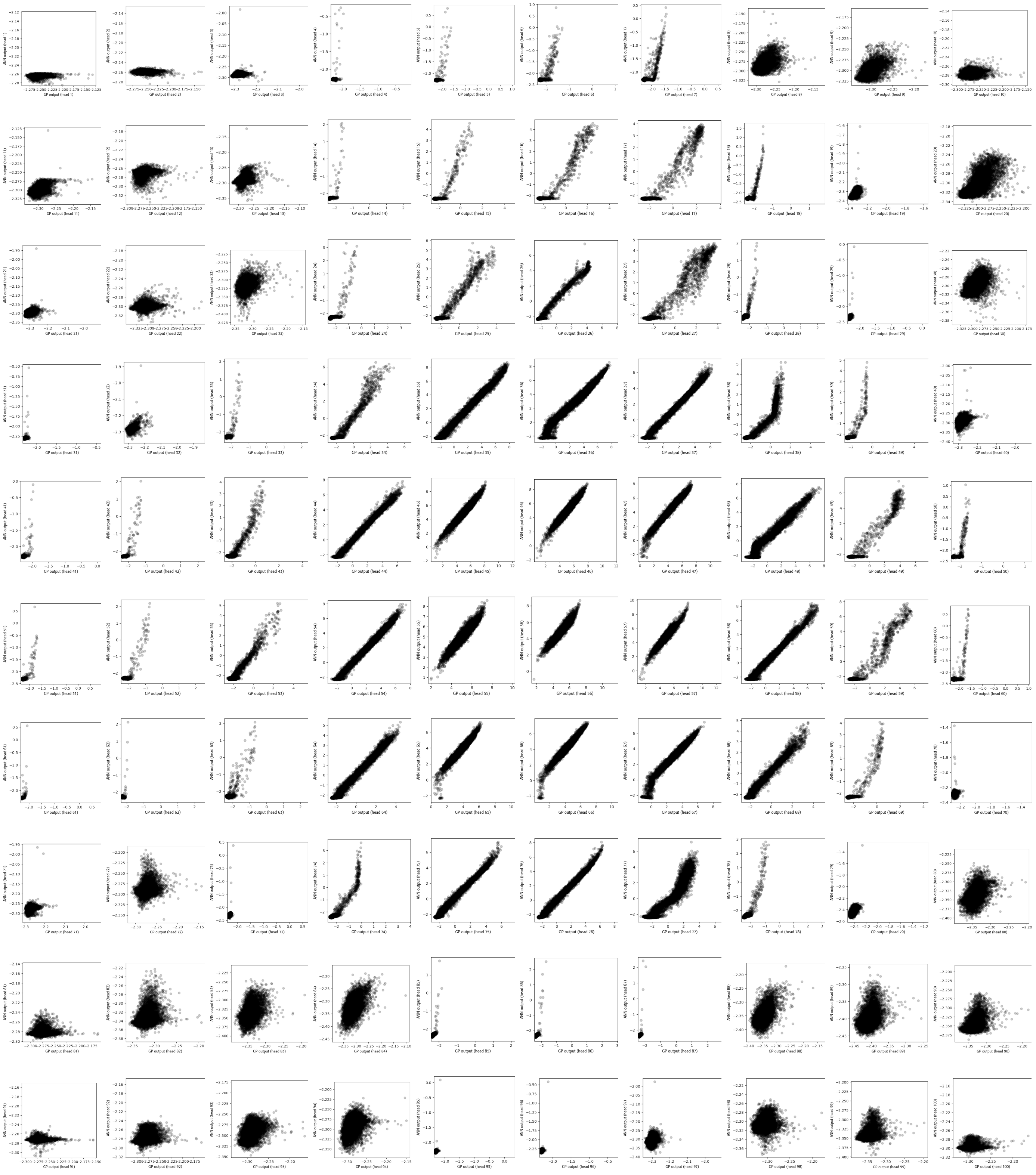

Fig. S5: Scatters for MNIST (attention).

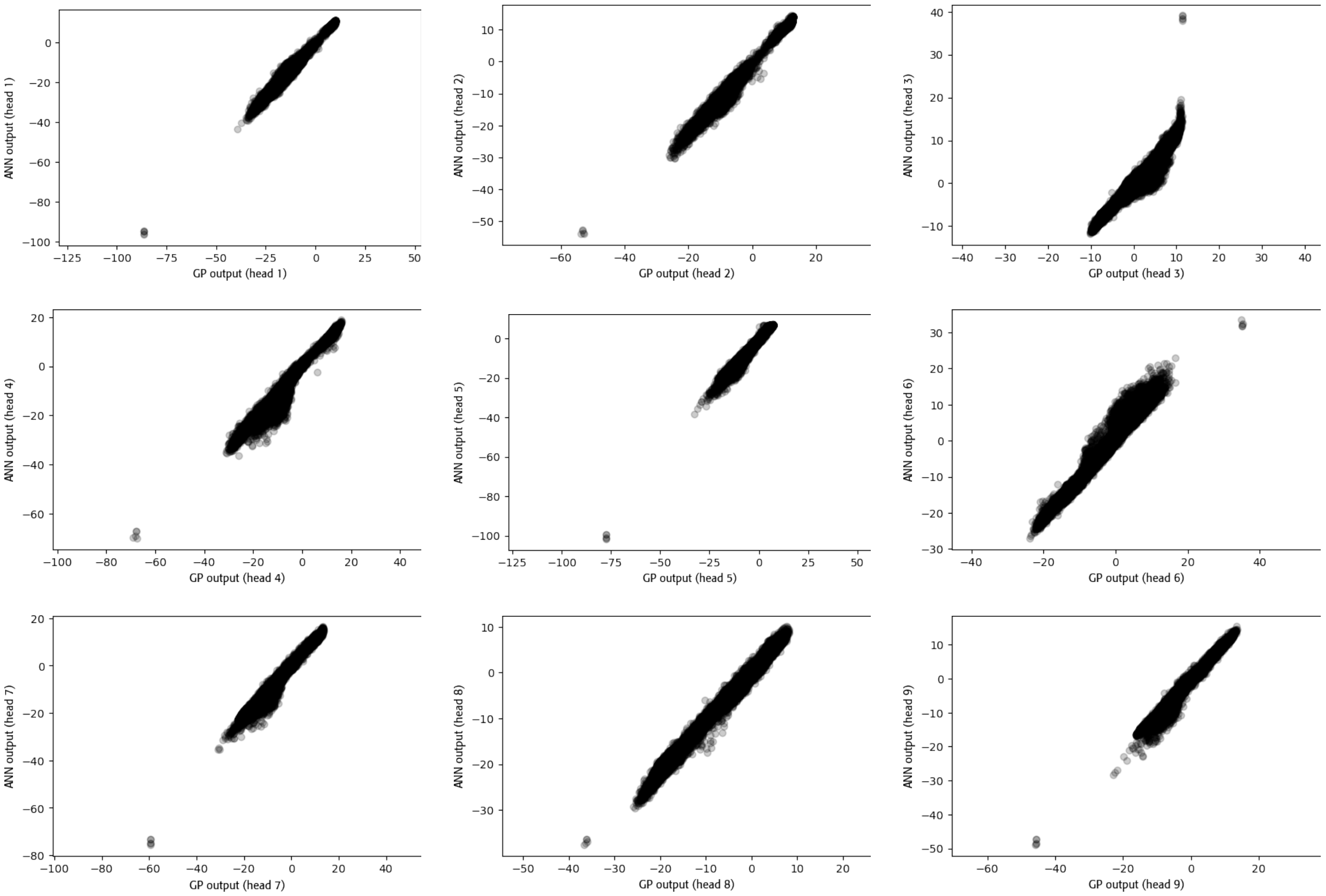

Fig. S6: Scatters for Kather dataset (classifier).

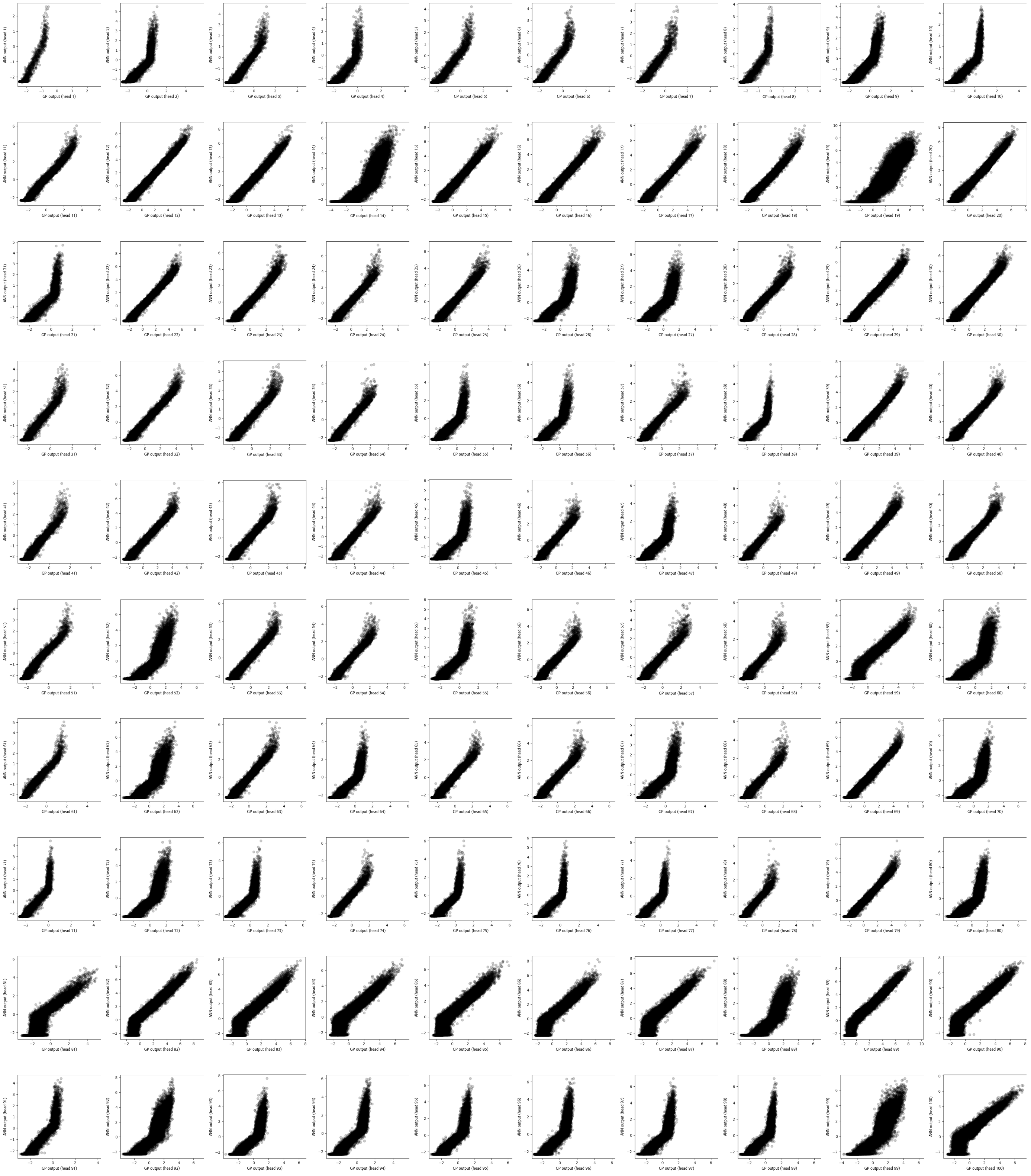

Fig. S7: Scatters for Kather dataset (attention).

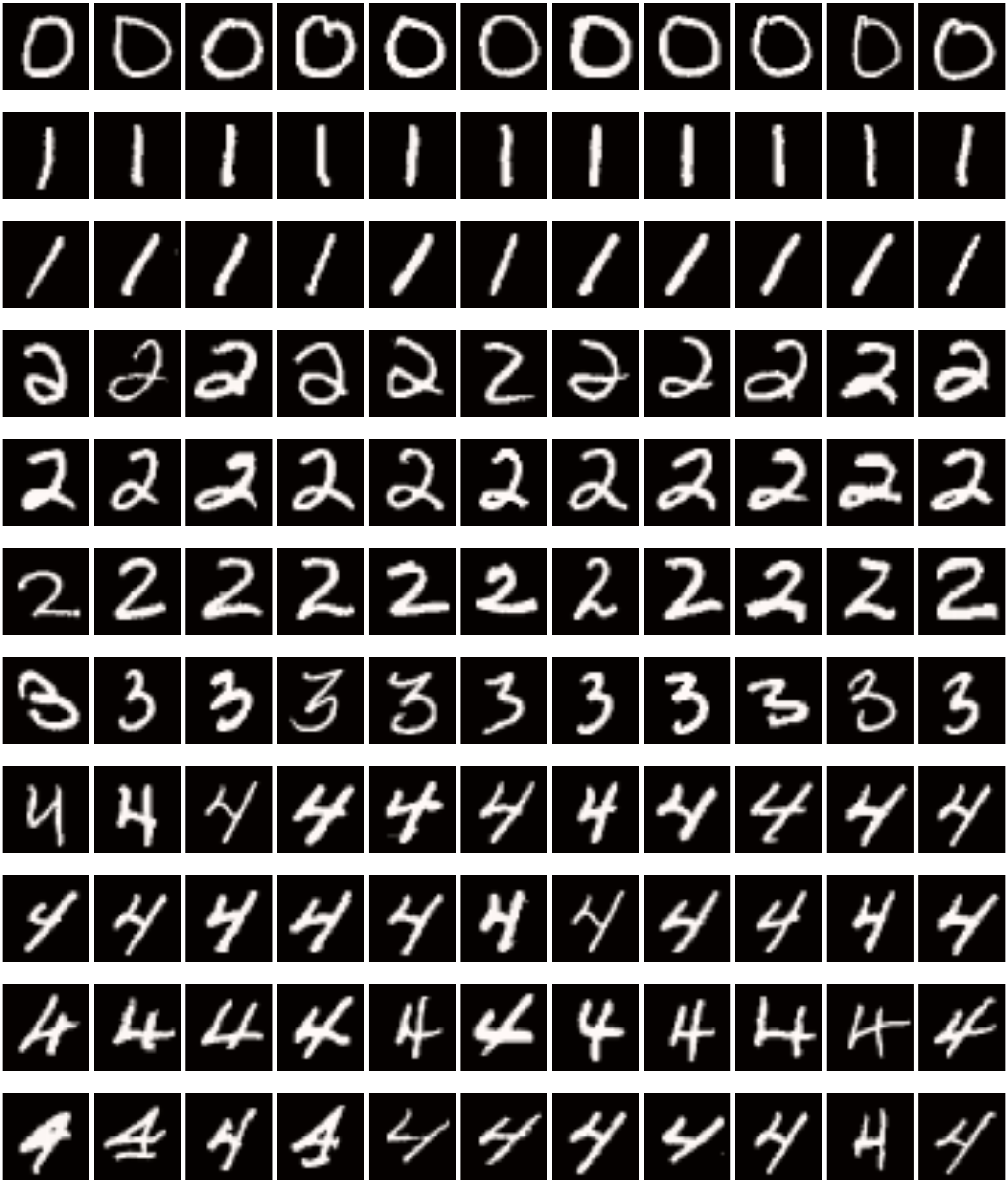

Fig. S8: Explanations for MNIST (set 1).

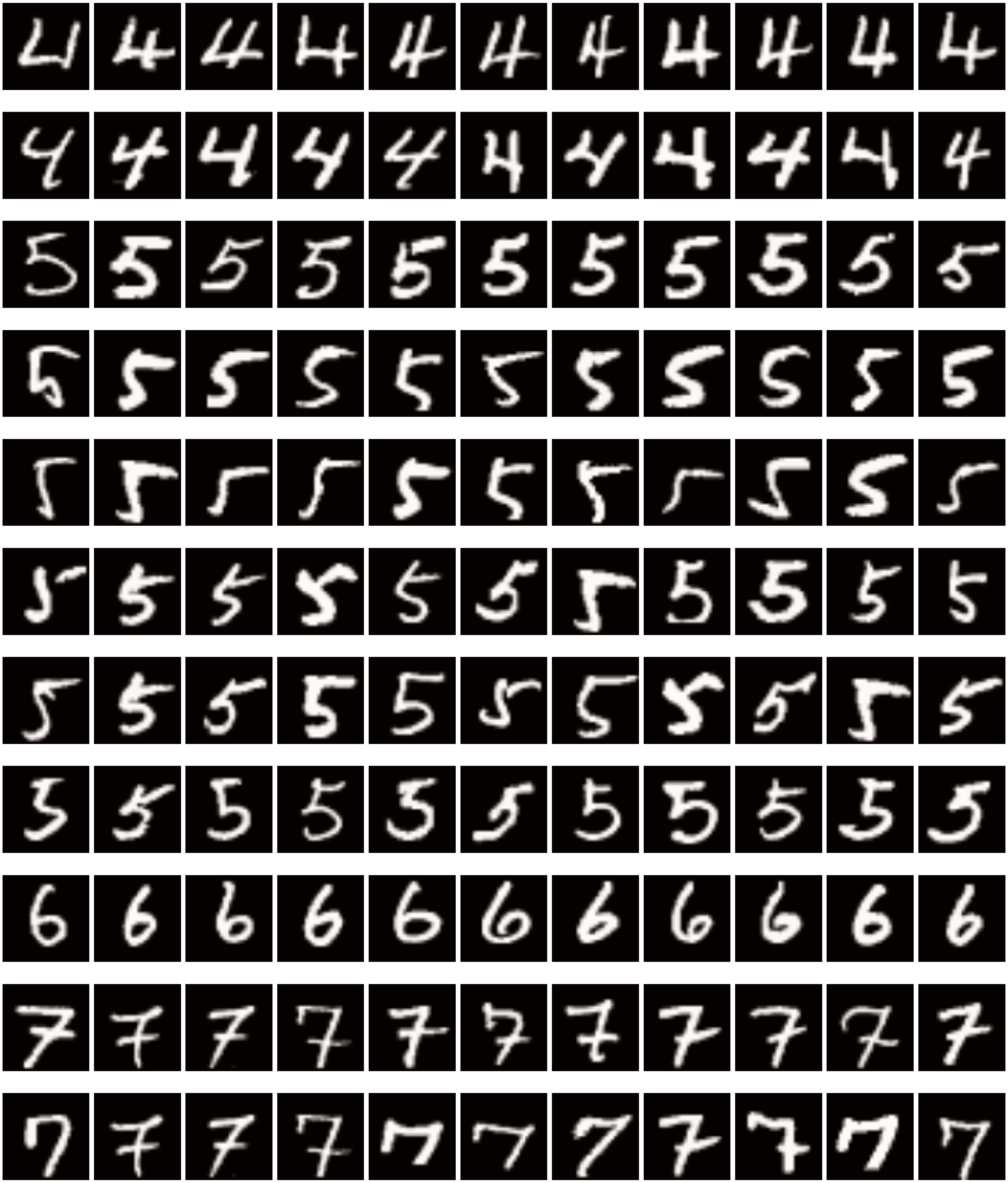

Fig. S9: Explanations for MNIST (set 2).

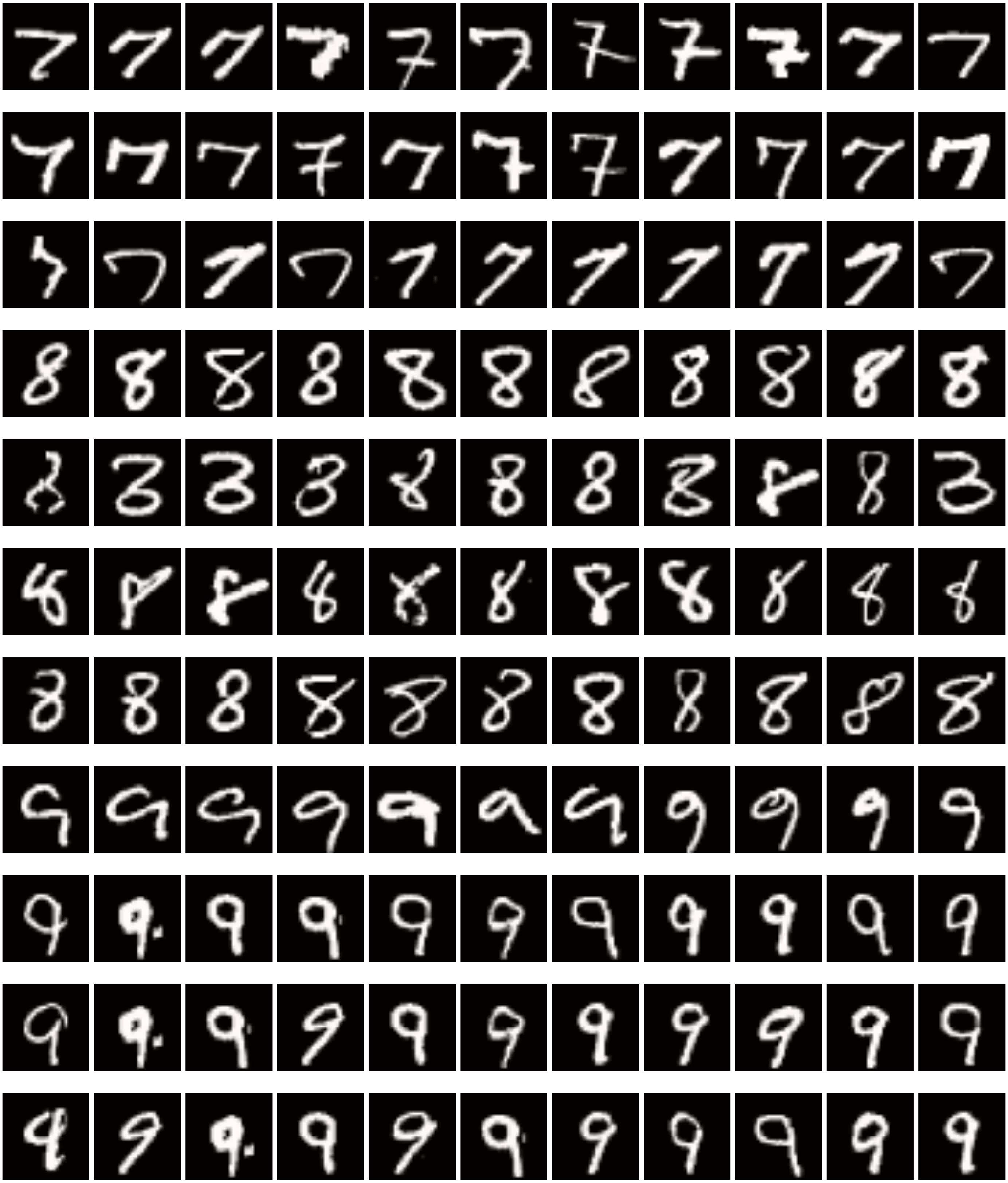

Fig. S10: Explanations for MNIST (set 3).

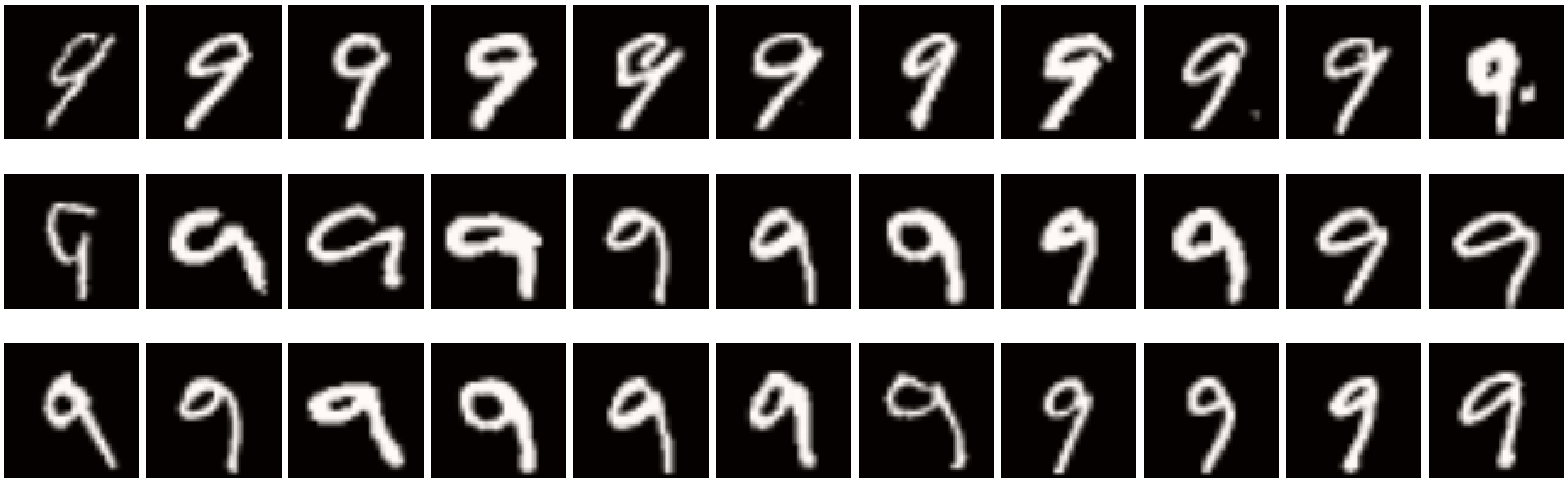

Fig. S11: Explanations for MNIST (set 4).

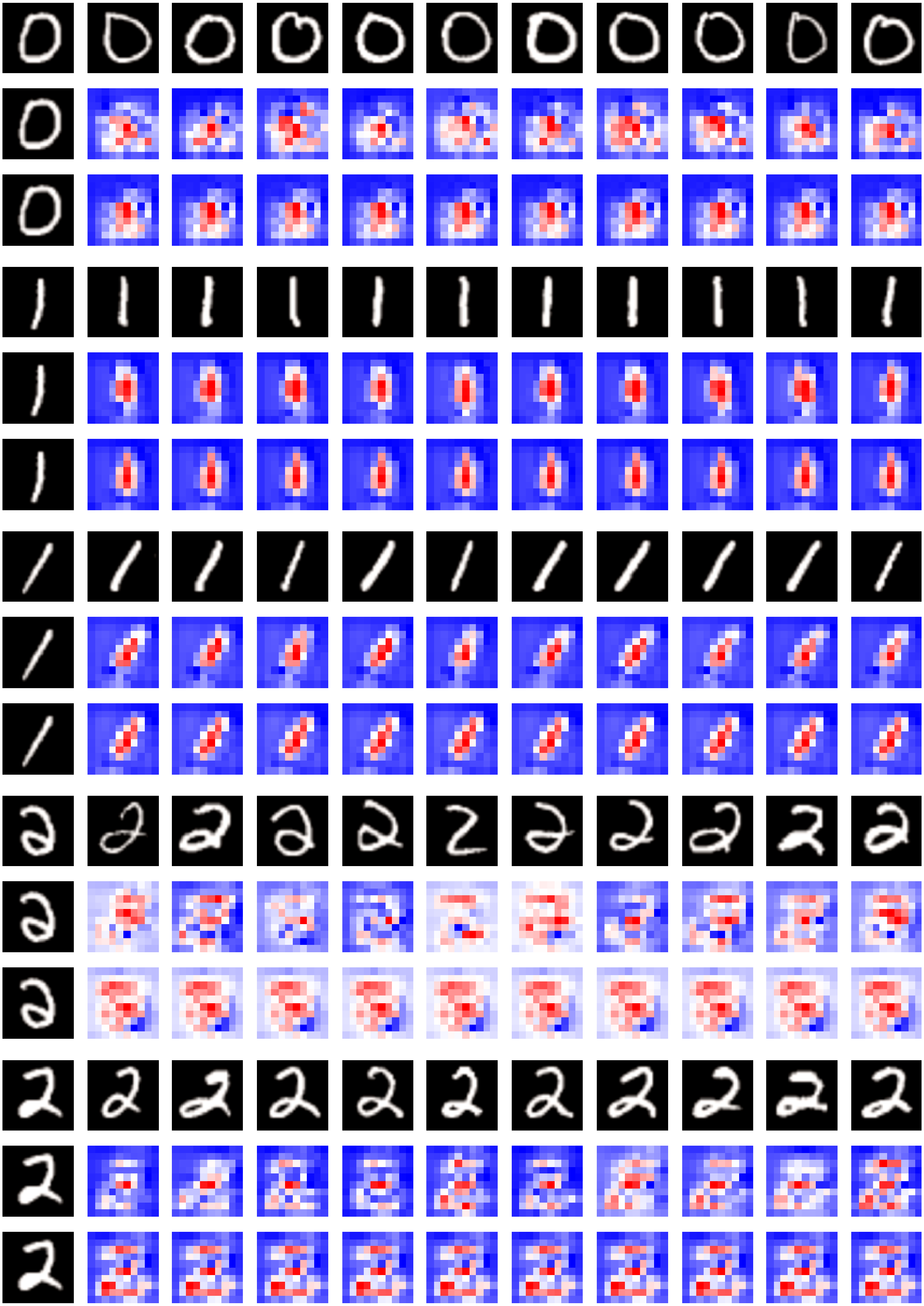

Fig. S12: Explanations for MNIST (set 5).

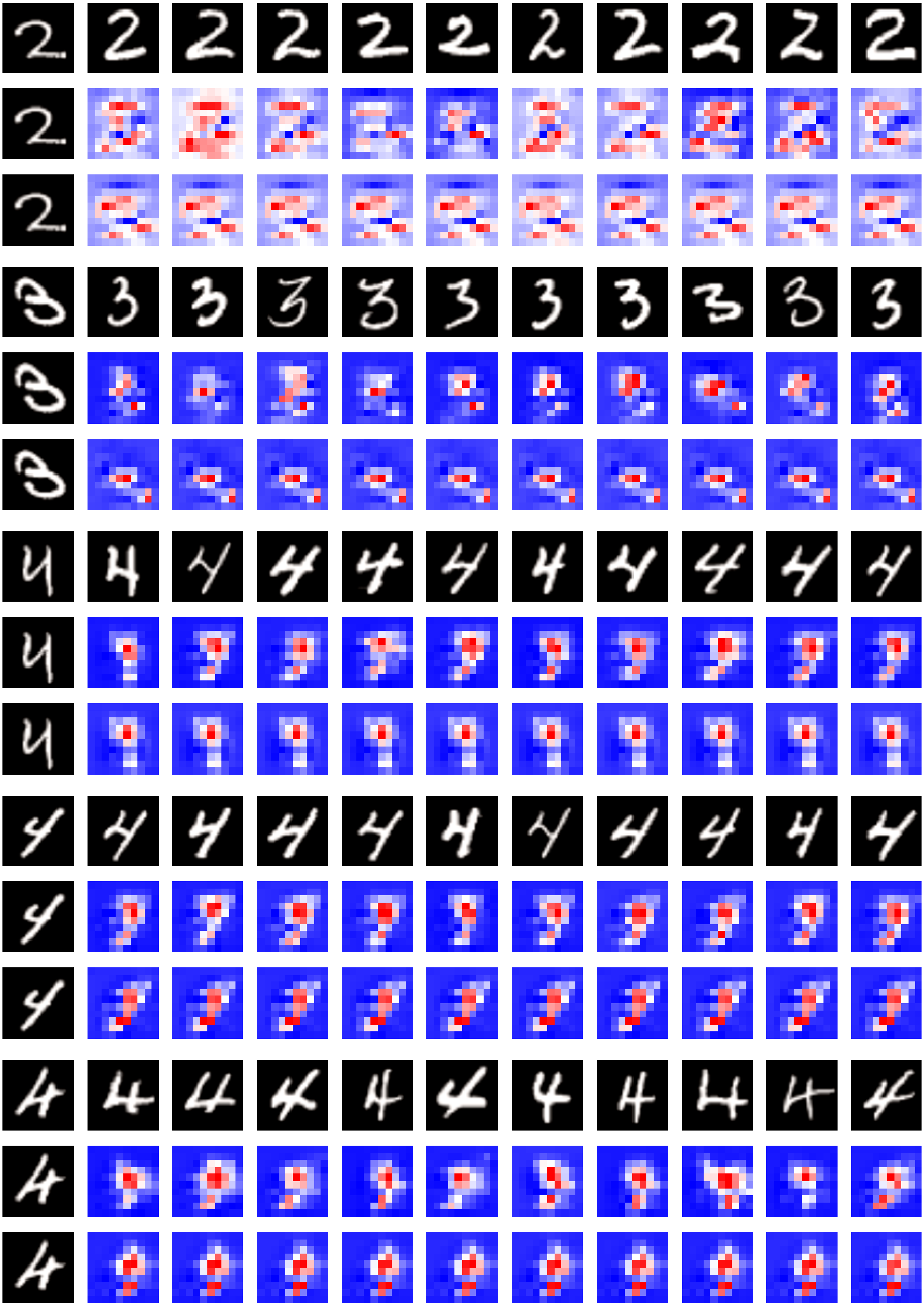

Fig. S13: Explanations for MNIST (set 6).

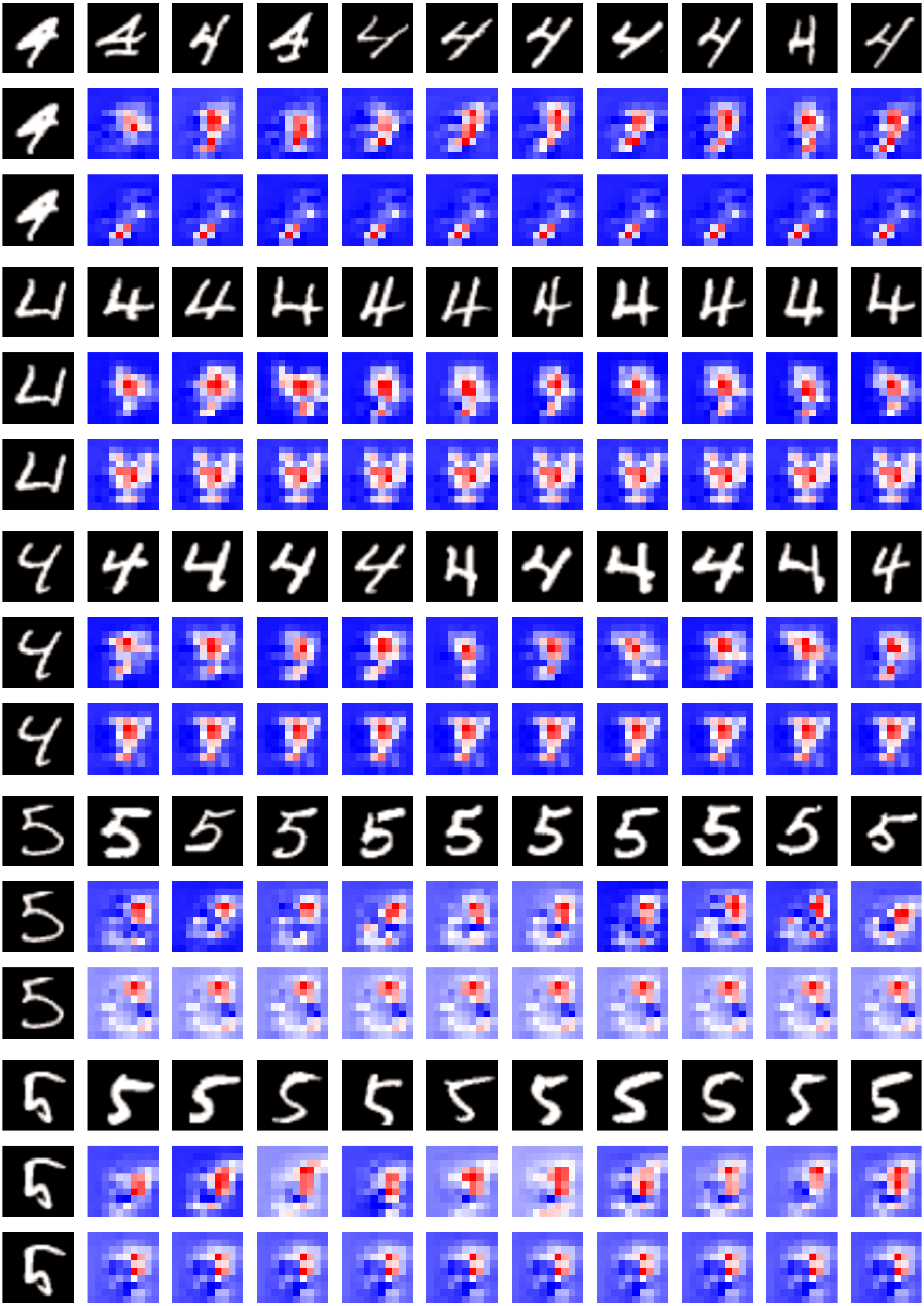

Fig. S14: Explanations for MNIST (set 7).

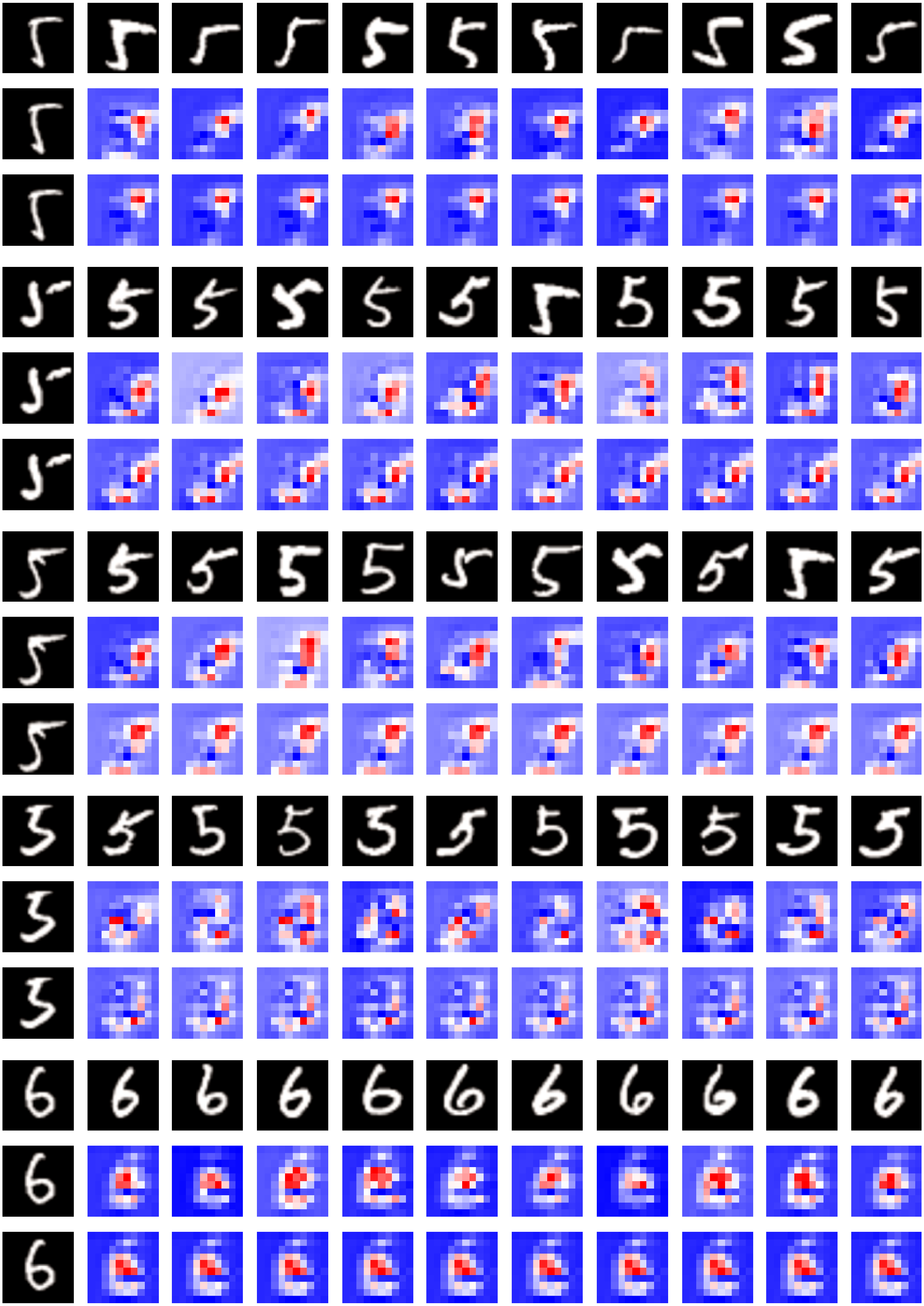

Fig. S15: Explanations for MNIST (set 8).

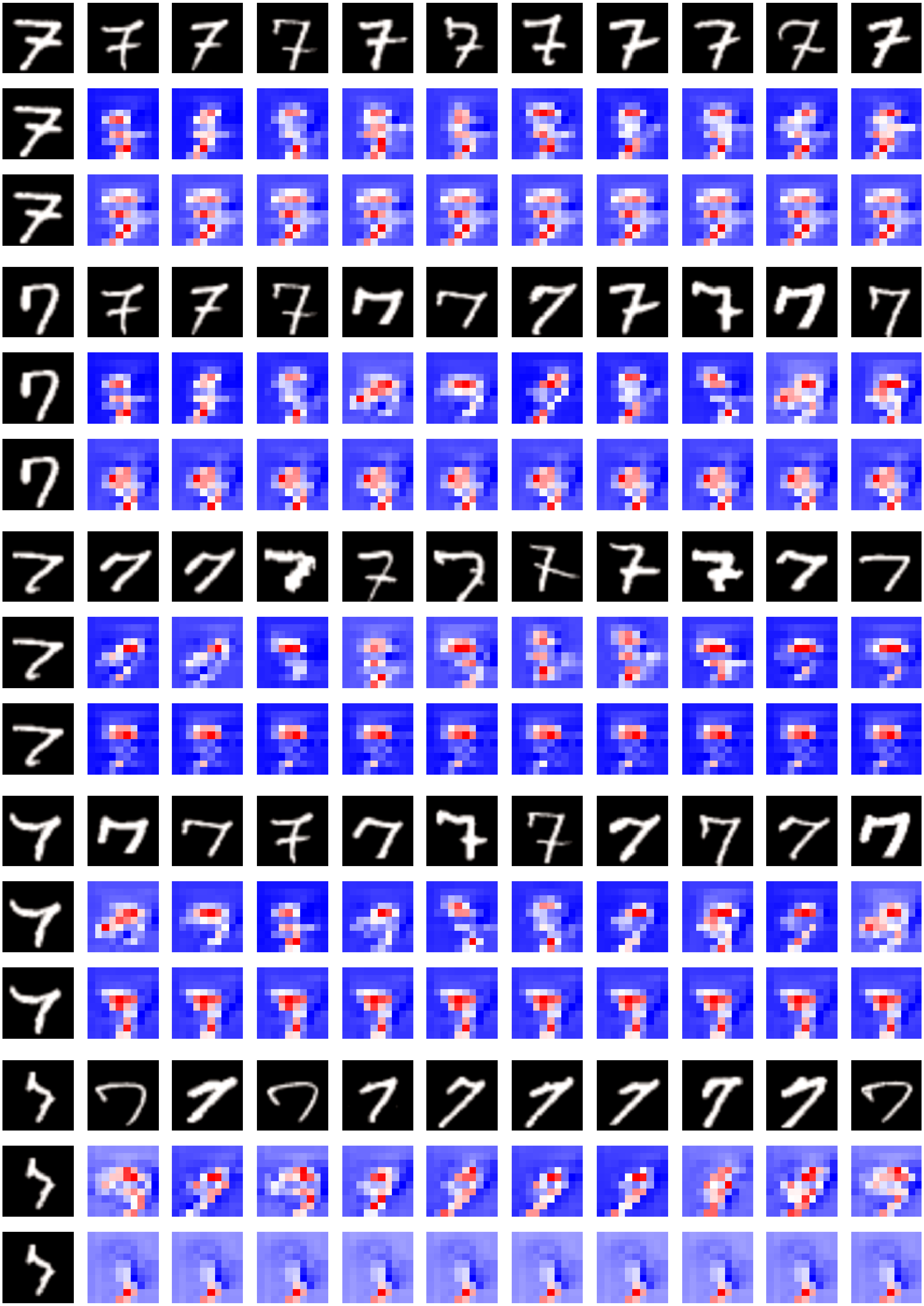

Fig. S16: Explanations for MNIST (set 9).

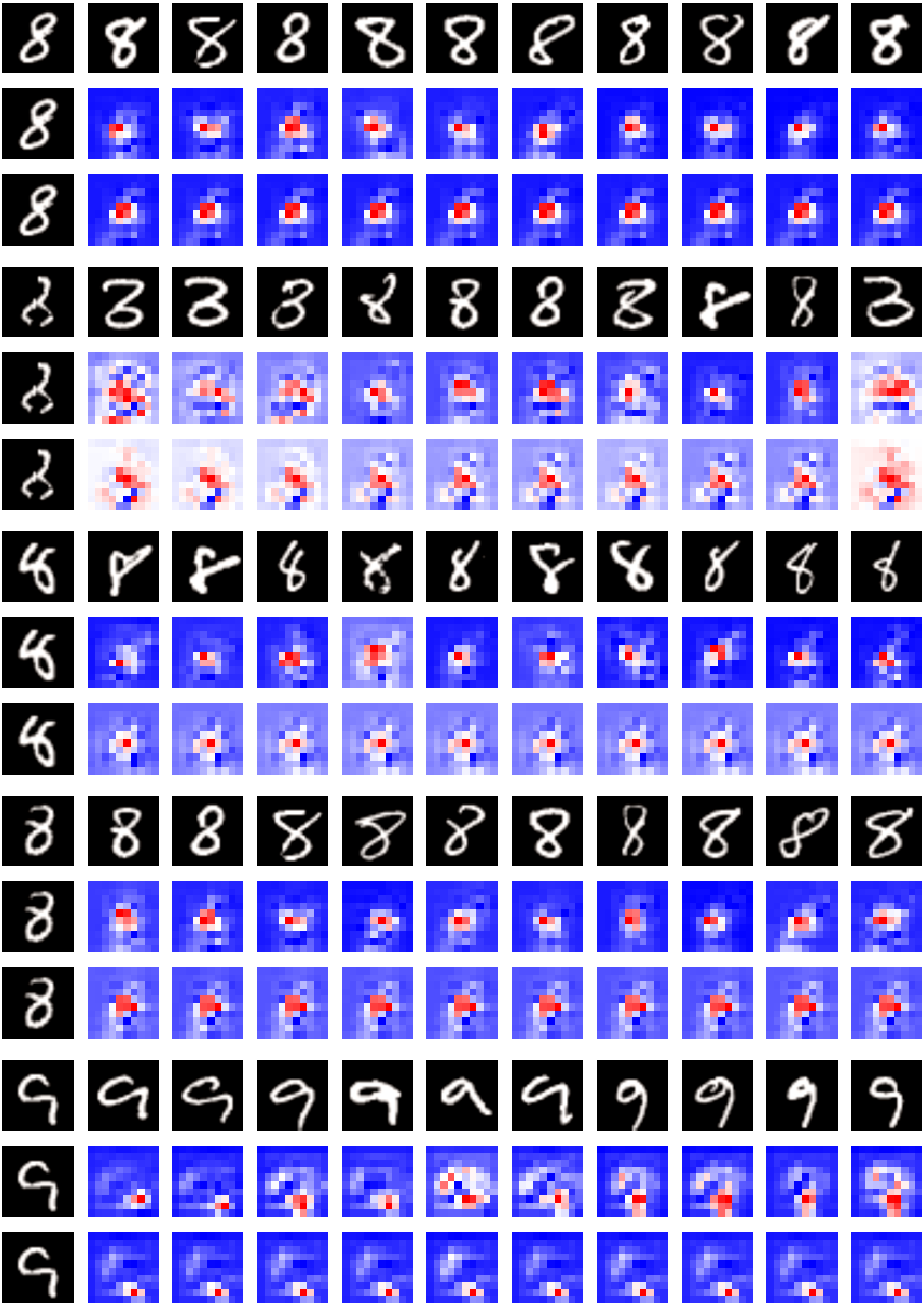

Fig. S17: Explanations for MNIST (set 10).

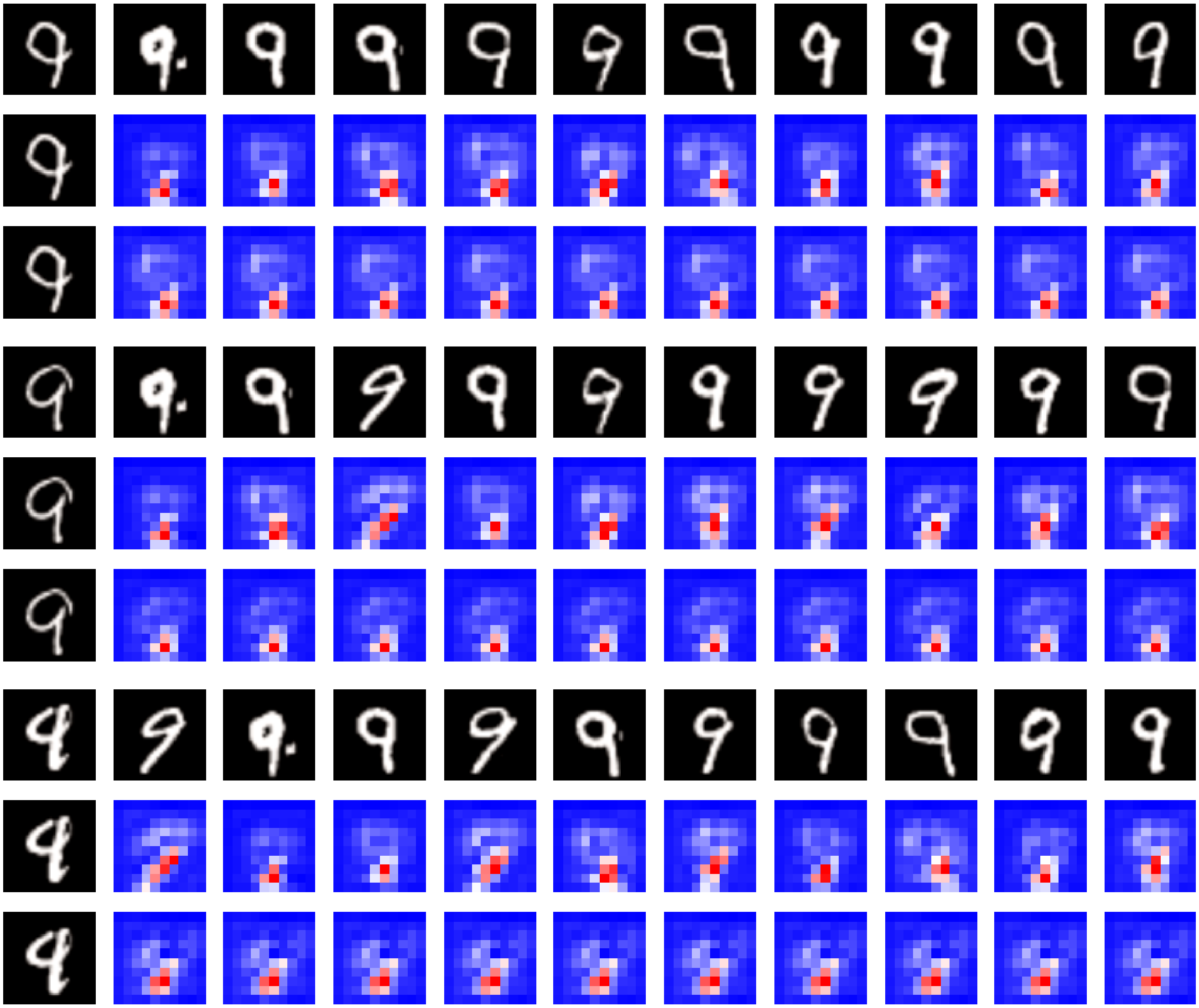

Fig. S18: Explanations for MNIST (set 11).

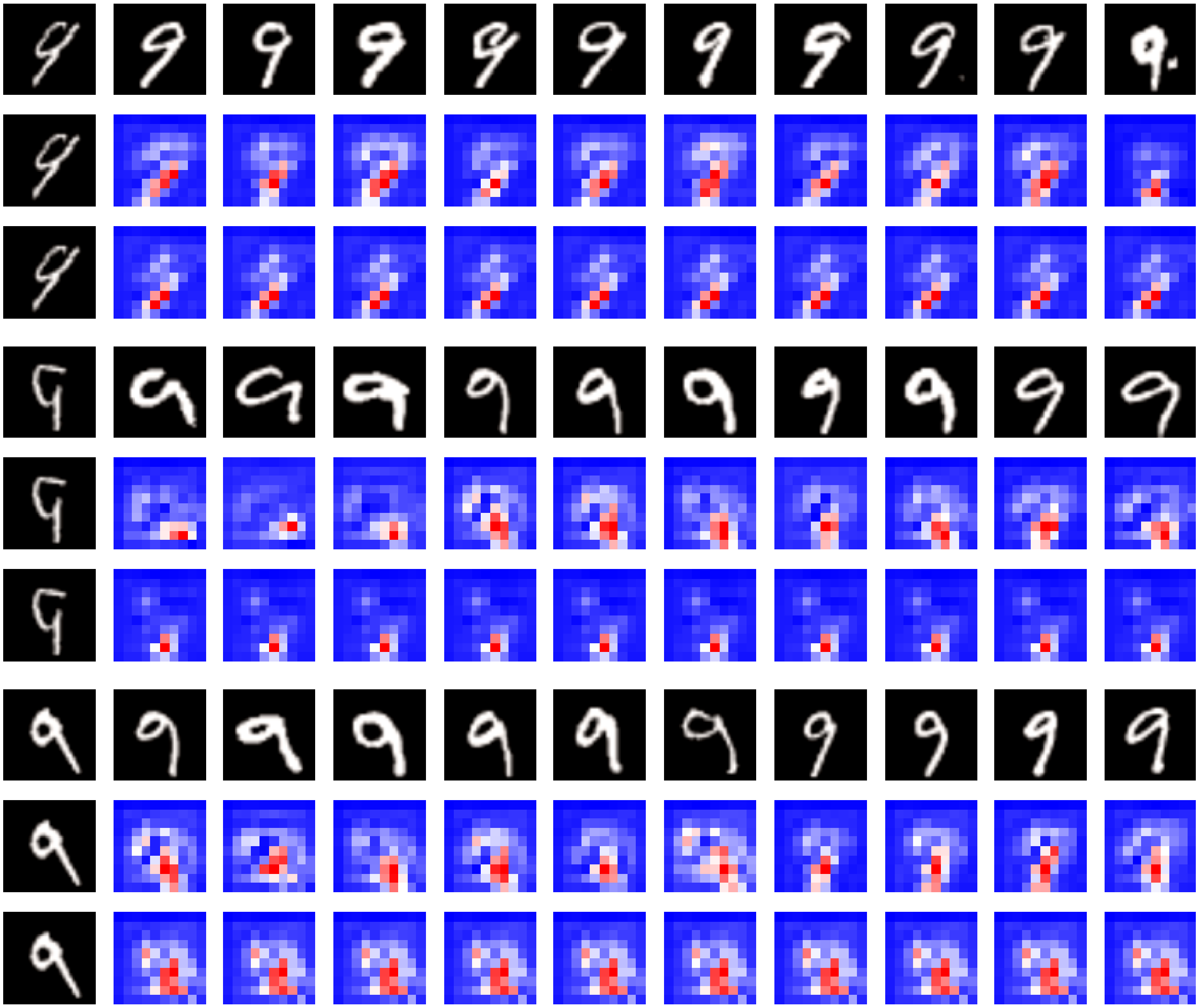

Fig. S19: Explanations for MNIST (set 12).

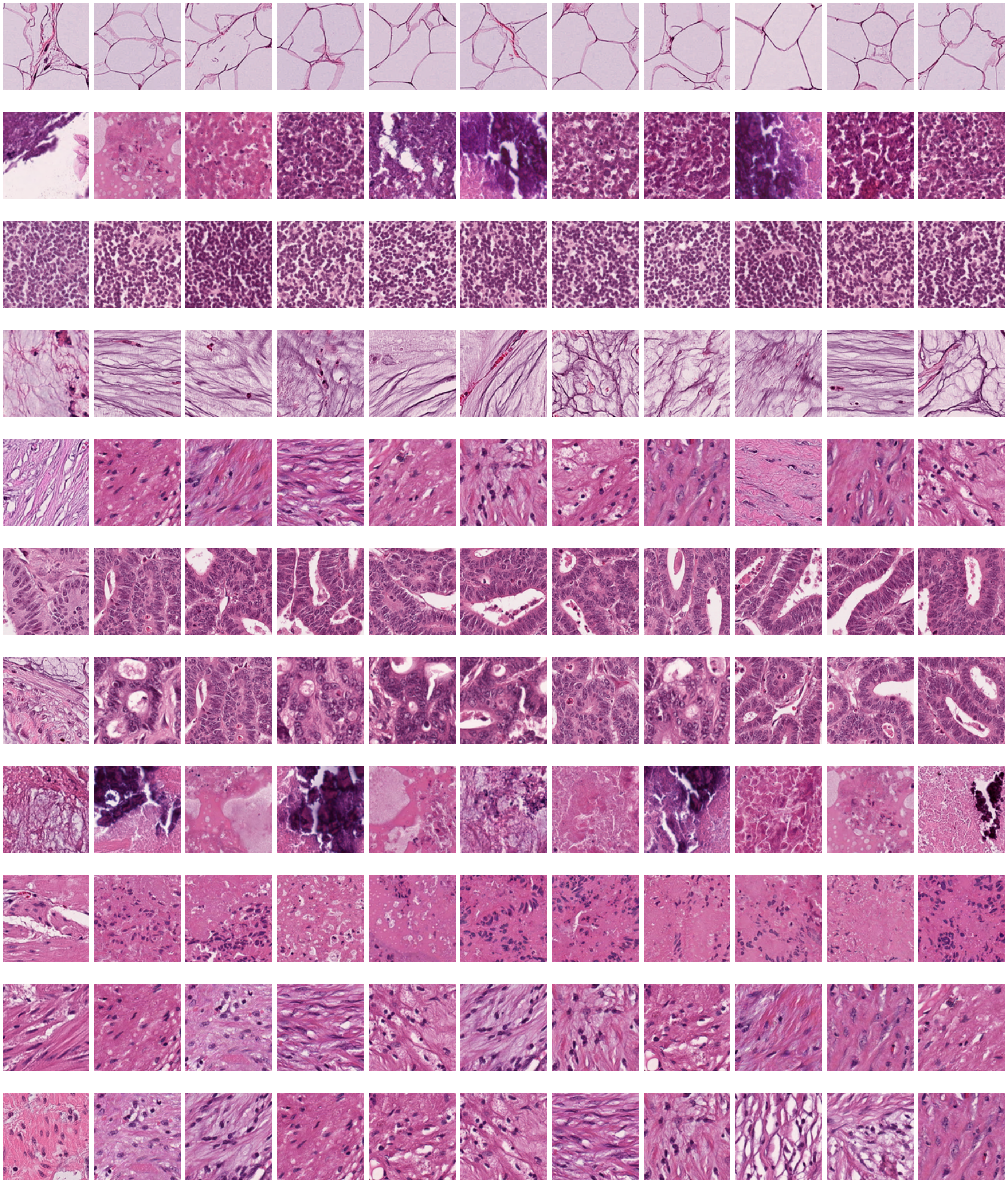

Fig. S20: Explanations for Kather dataset (set 1).

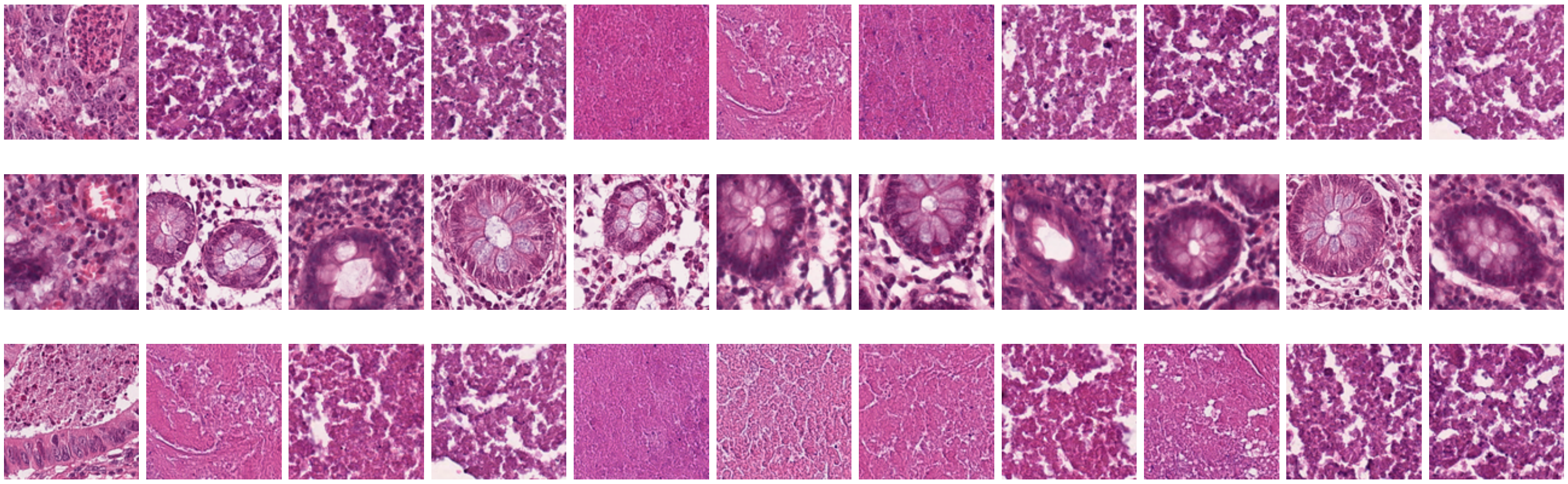

Fig. S21: Explanations for Kather dataset (set 2).

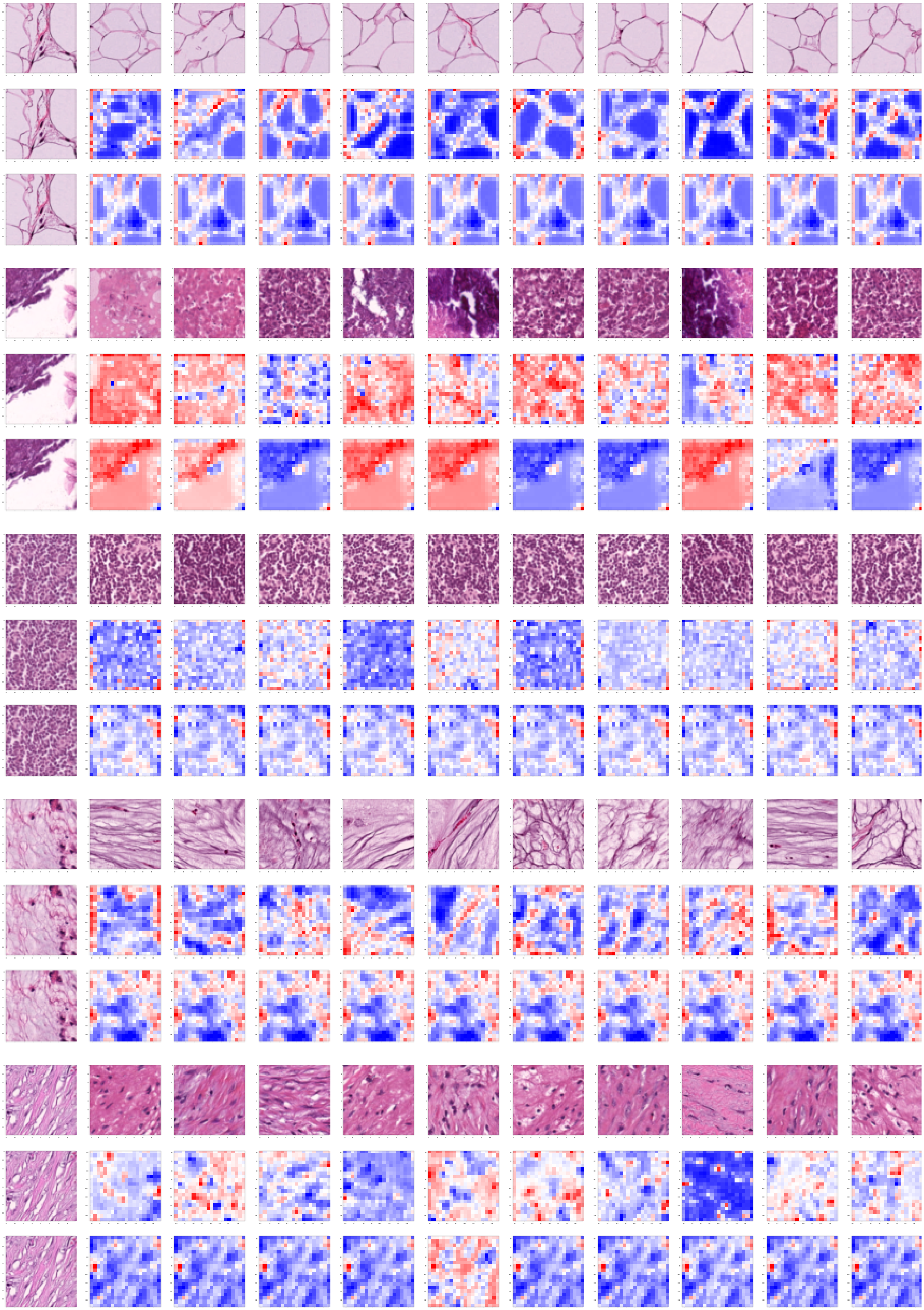

Fig. S22: Explanations for Kather dataset (set 3).

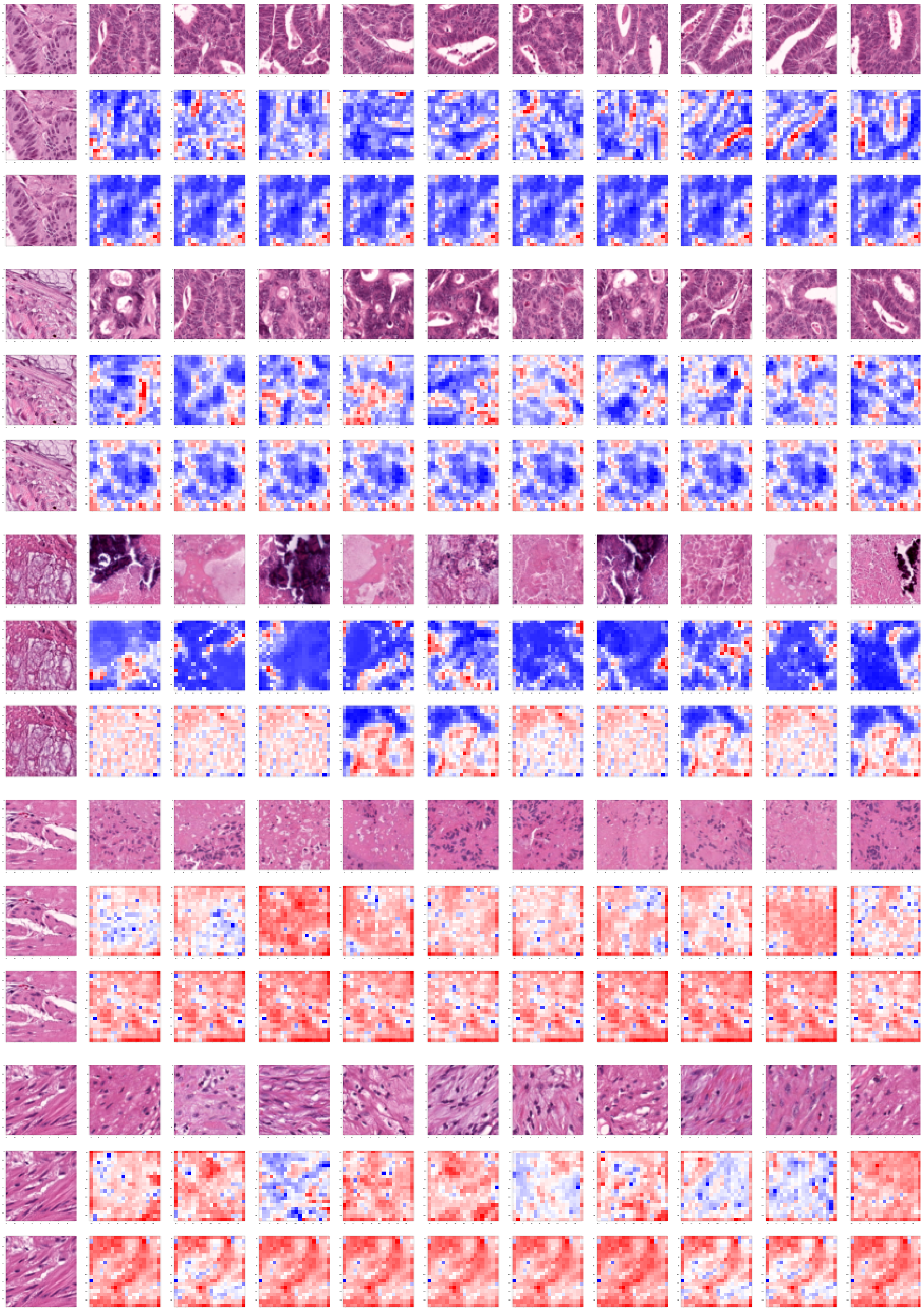

Fig. S23: Explanations for Kather dataset (set 4).

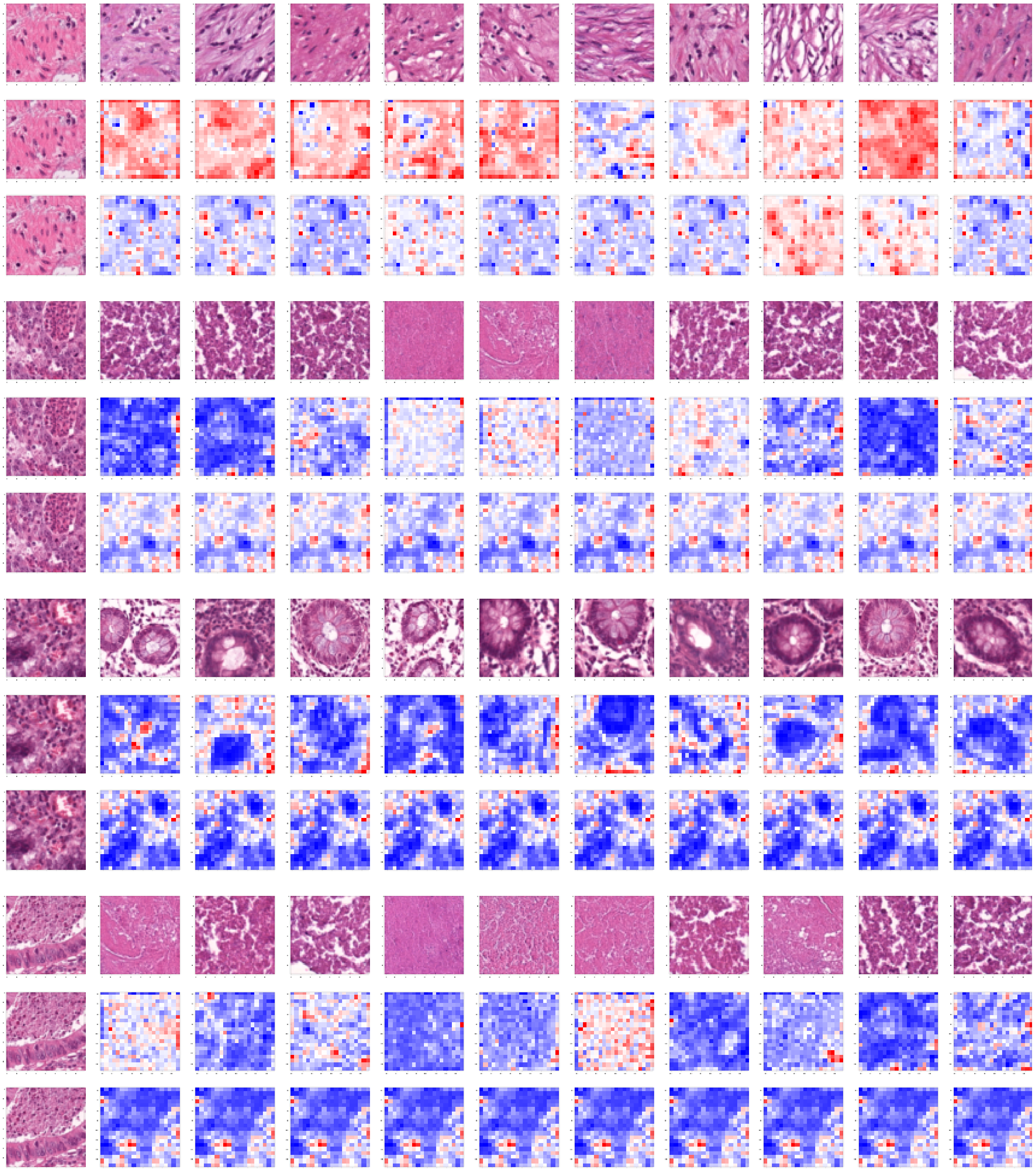

Fig. S24: Explanations for Kather dataset (set 5).

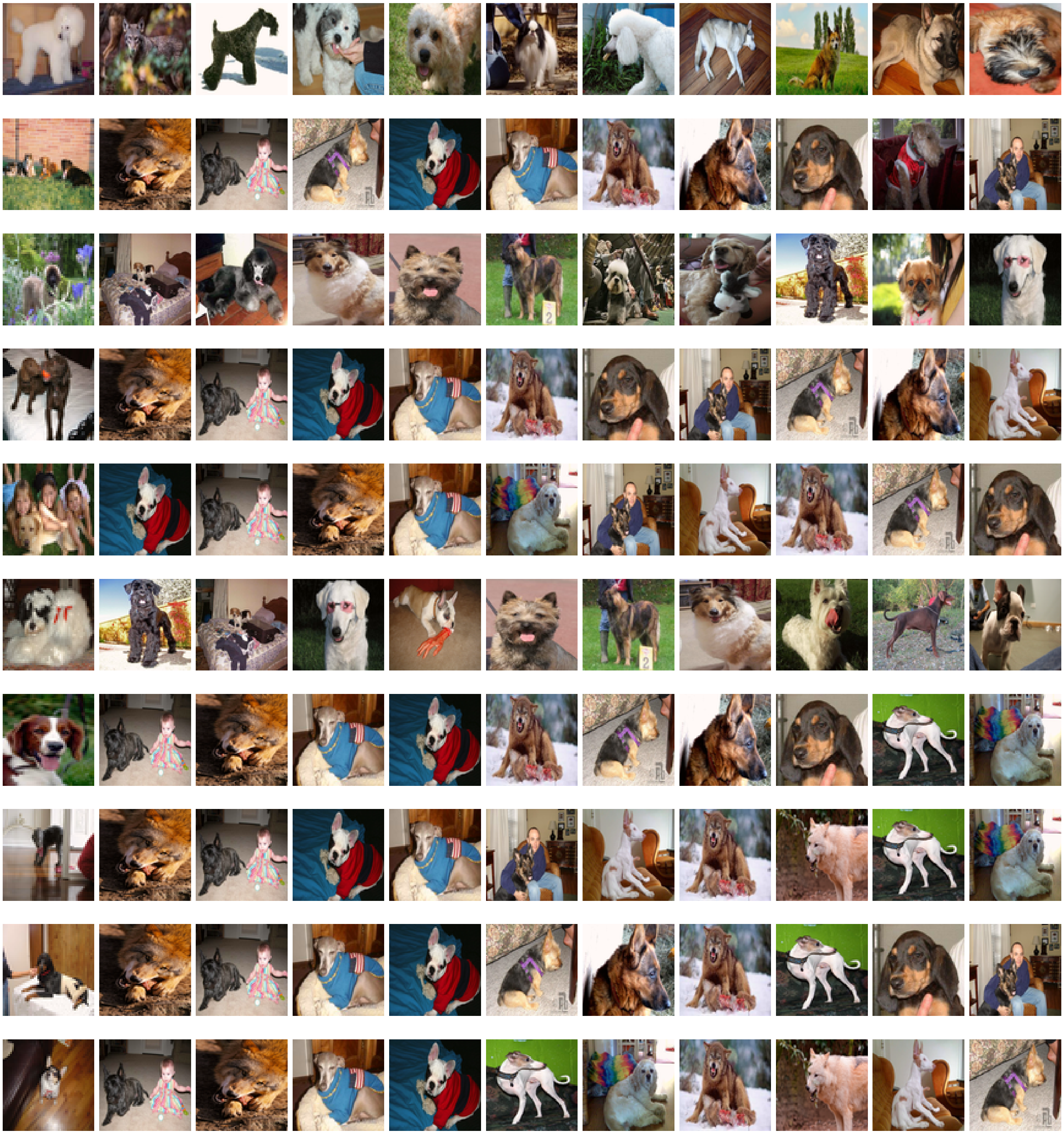

Fig. S25: Explanations for DogsWolves (set 1).

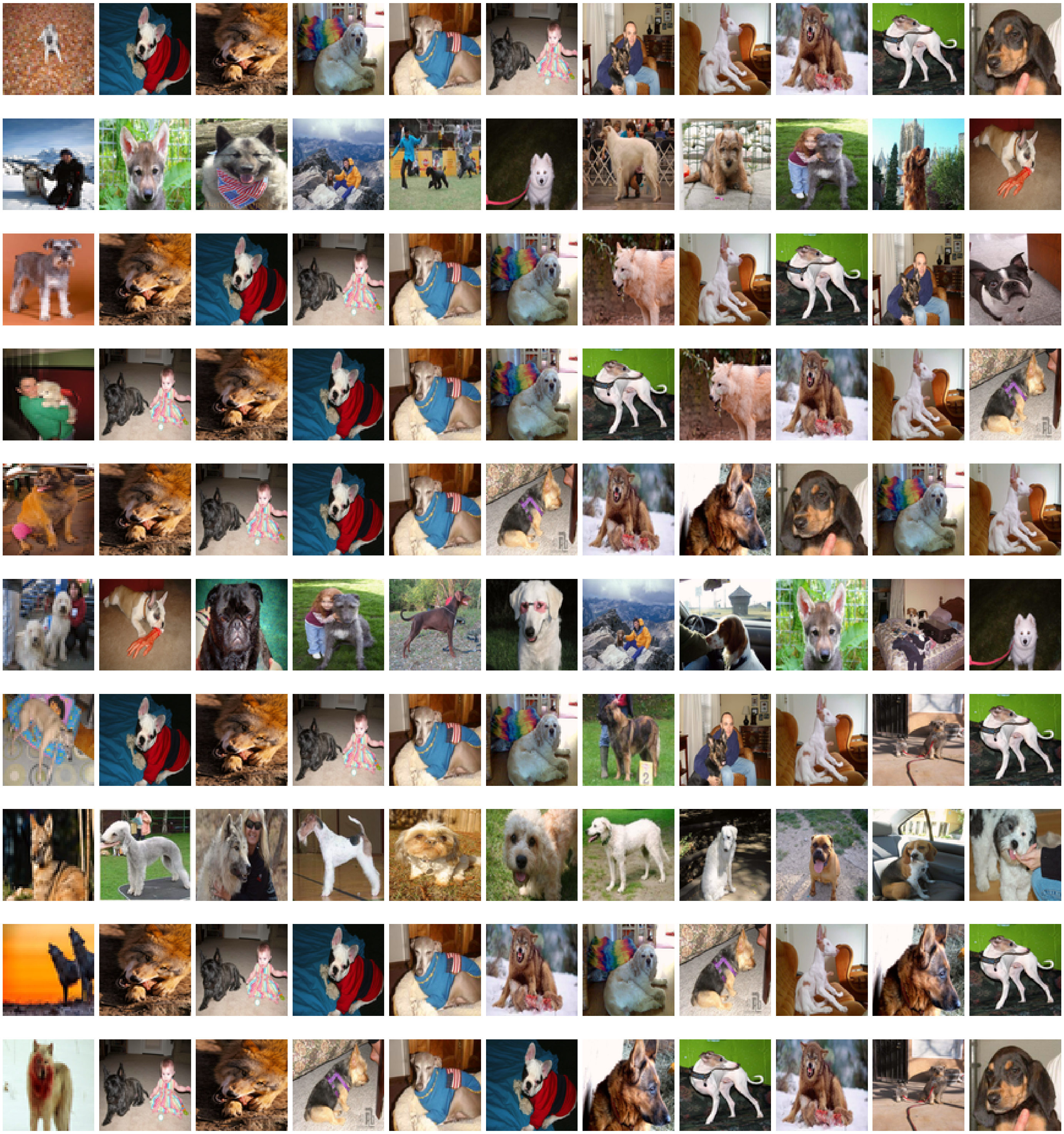

Fig. S26: Explanations for DogsWolves (set 2).

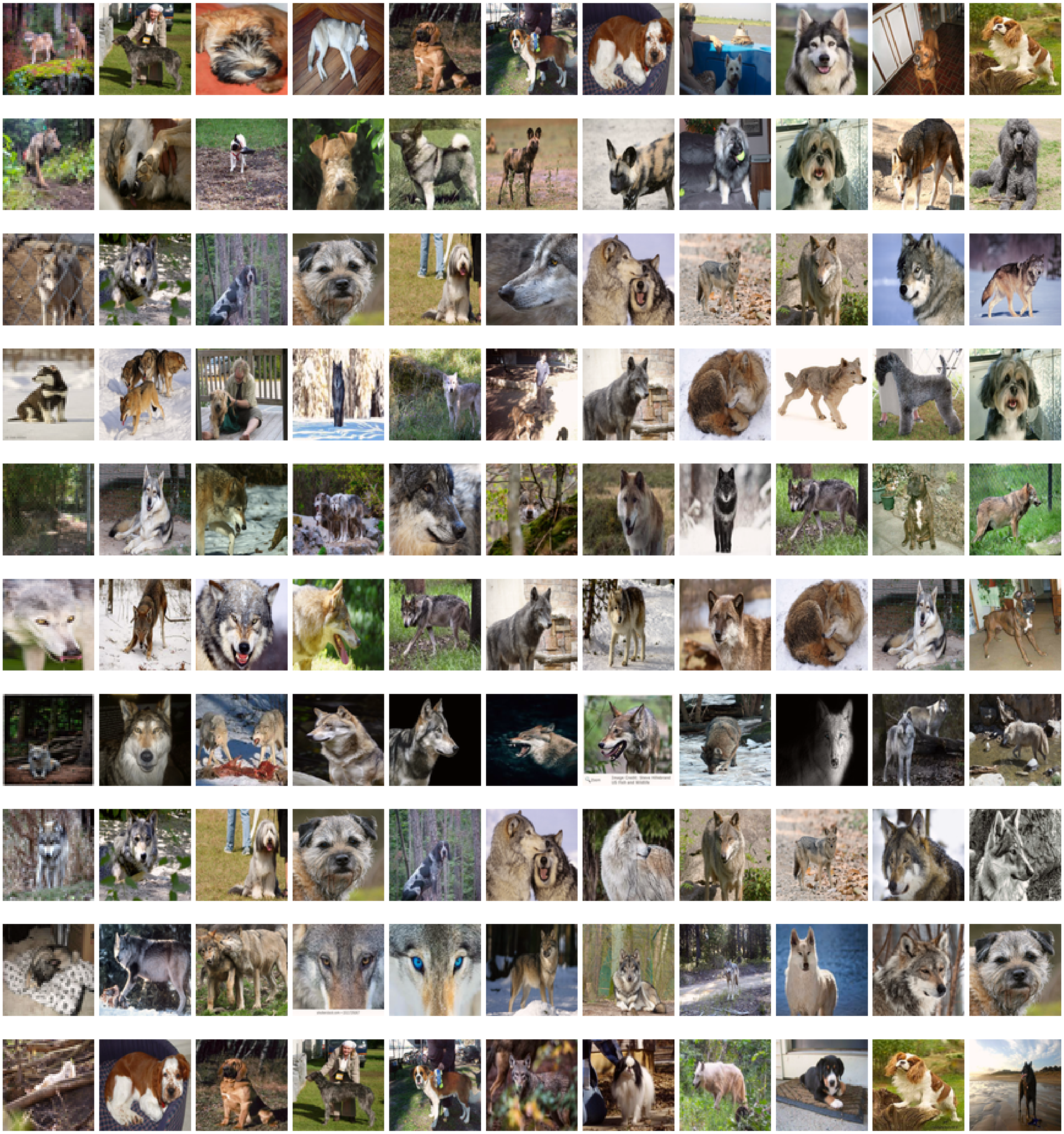

Fig. S27: Explanations for DogsWolves (set 3).

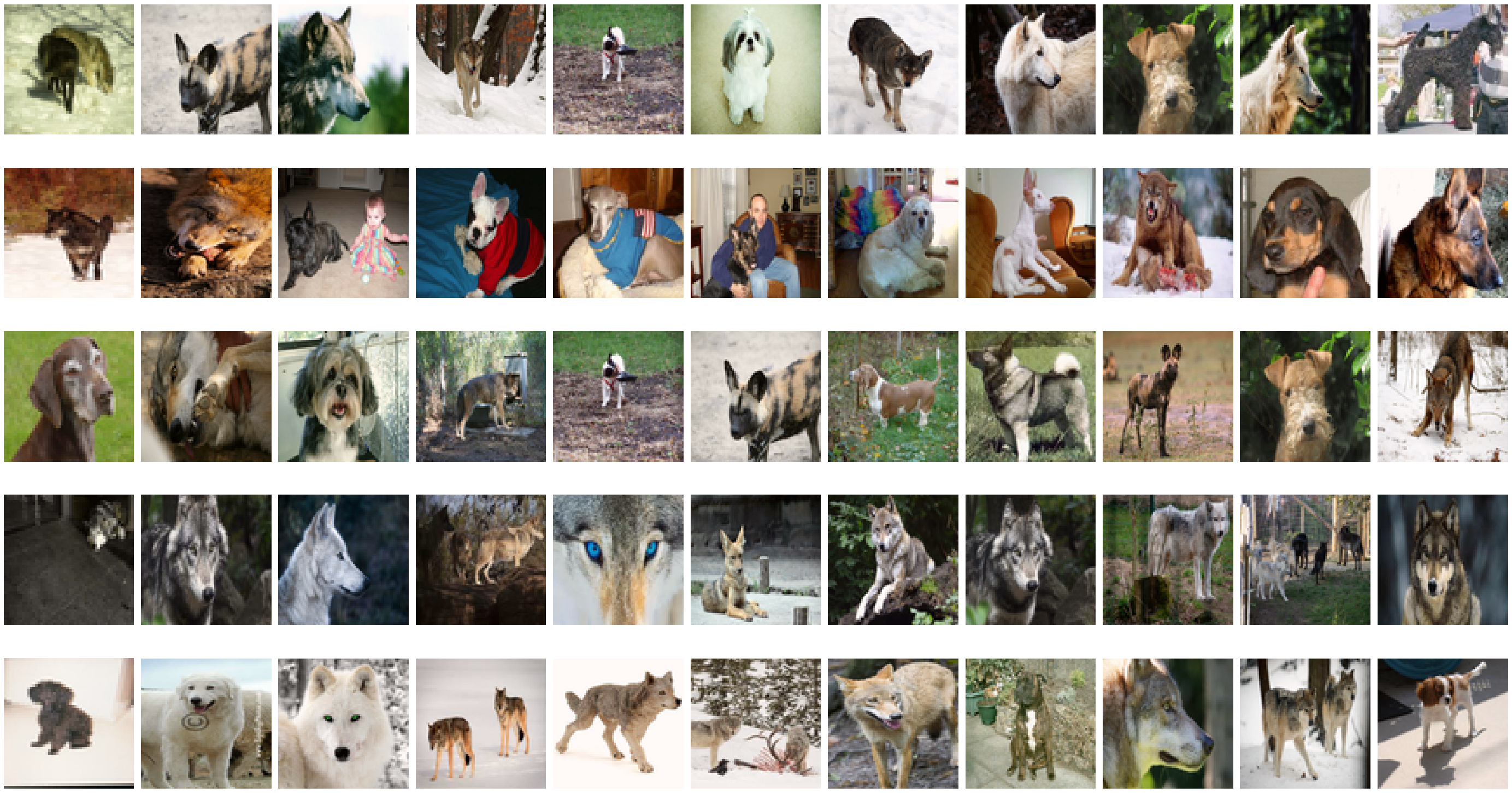

Fig. S28: Explanations for DogsWolves (set 4).

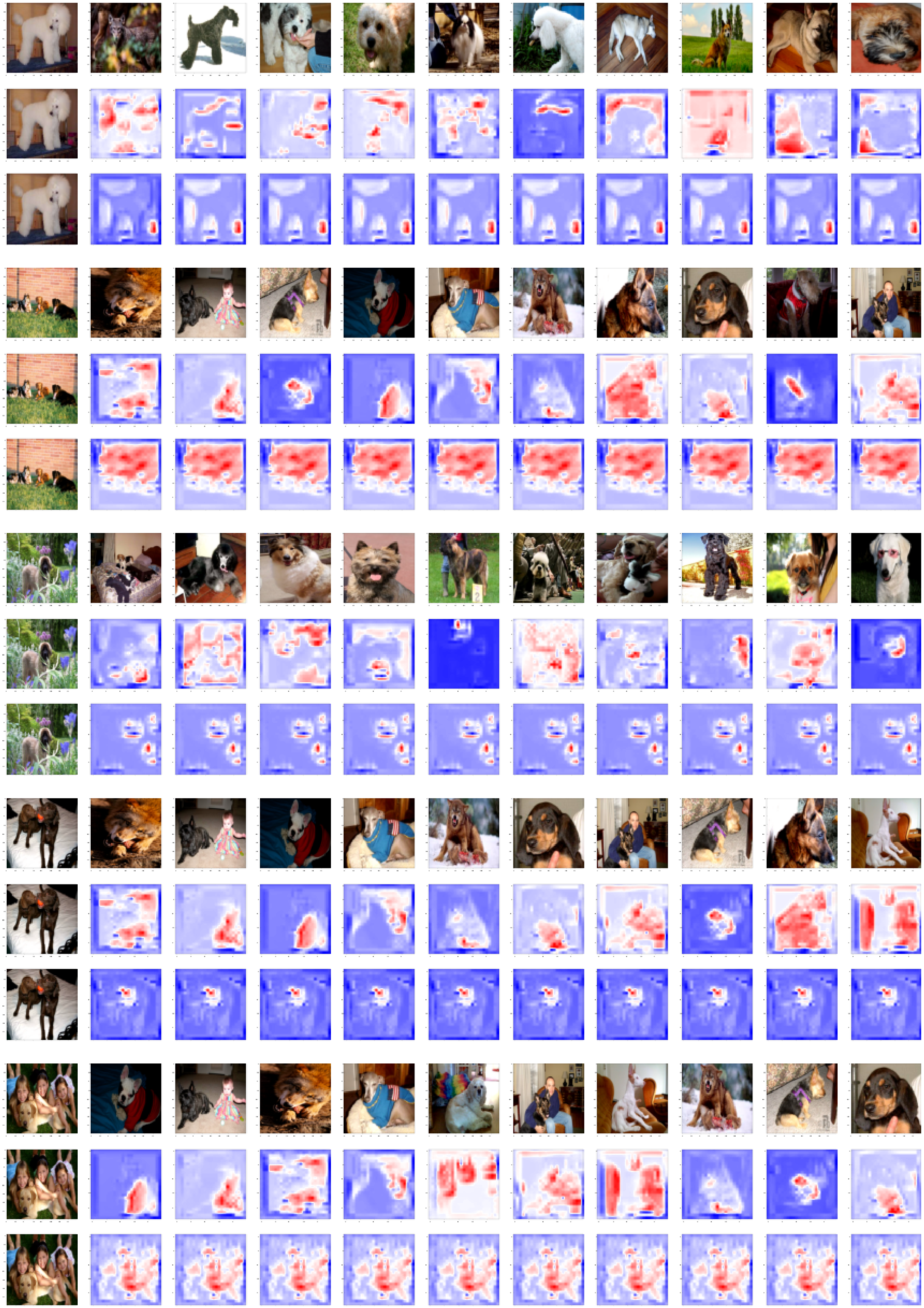

Fig. S29: Explanations for DogsWolves (set 5).

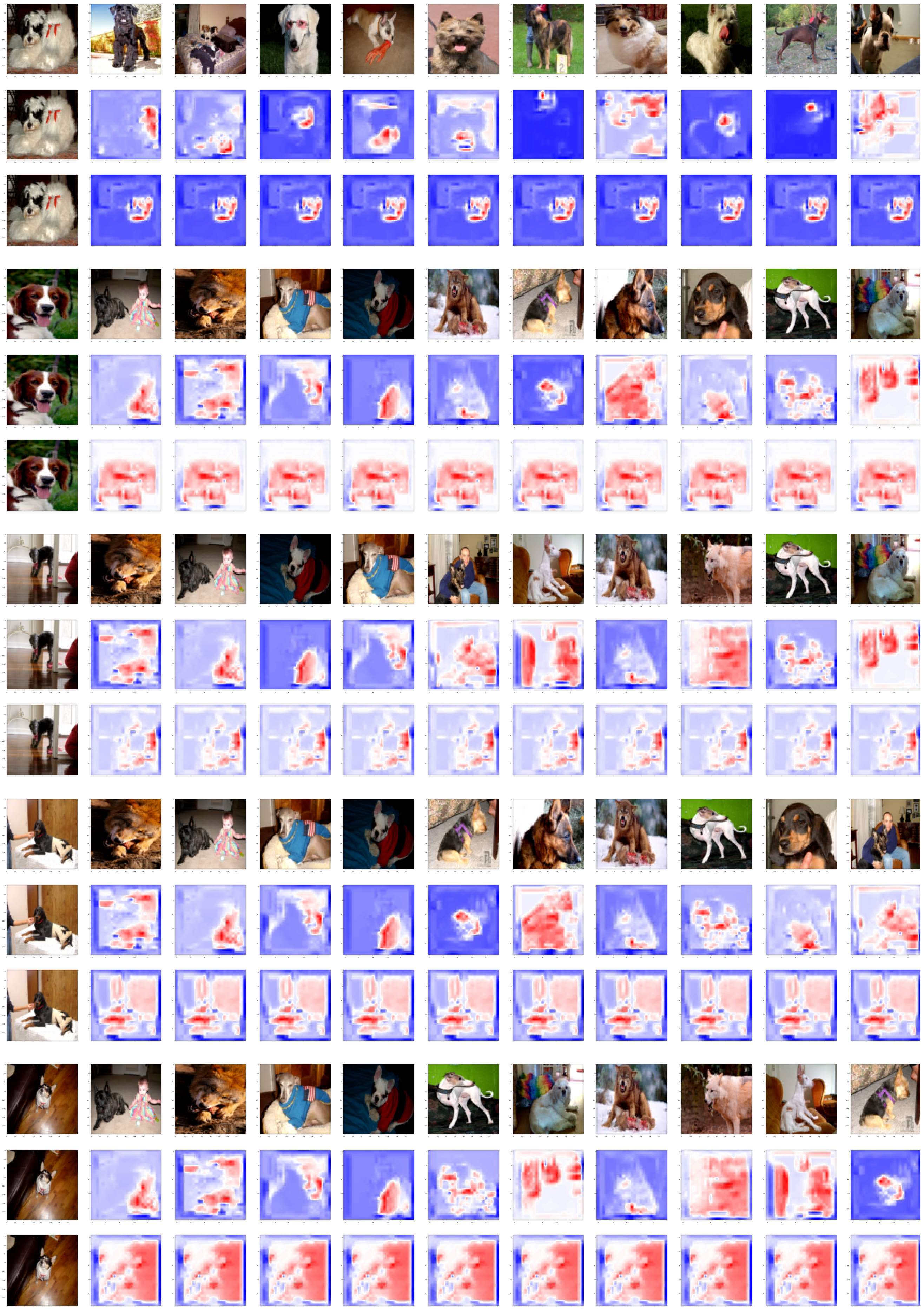

Fig. S30: Explanations for DogsWolves (set 6).

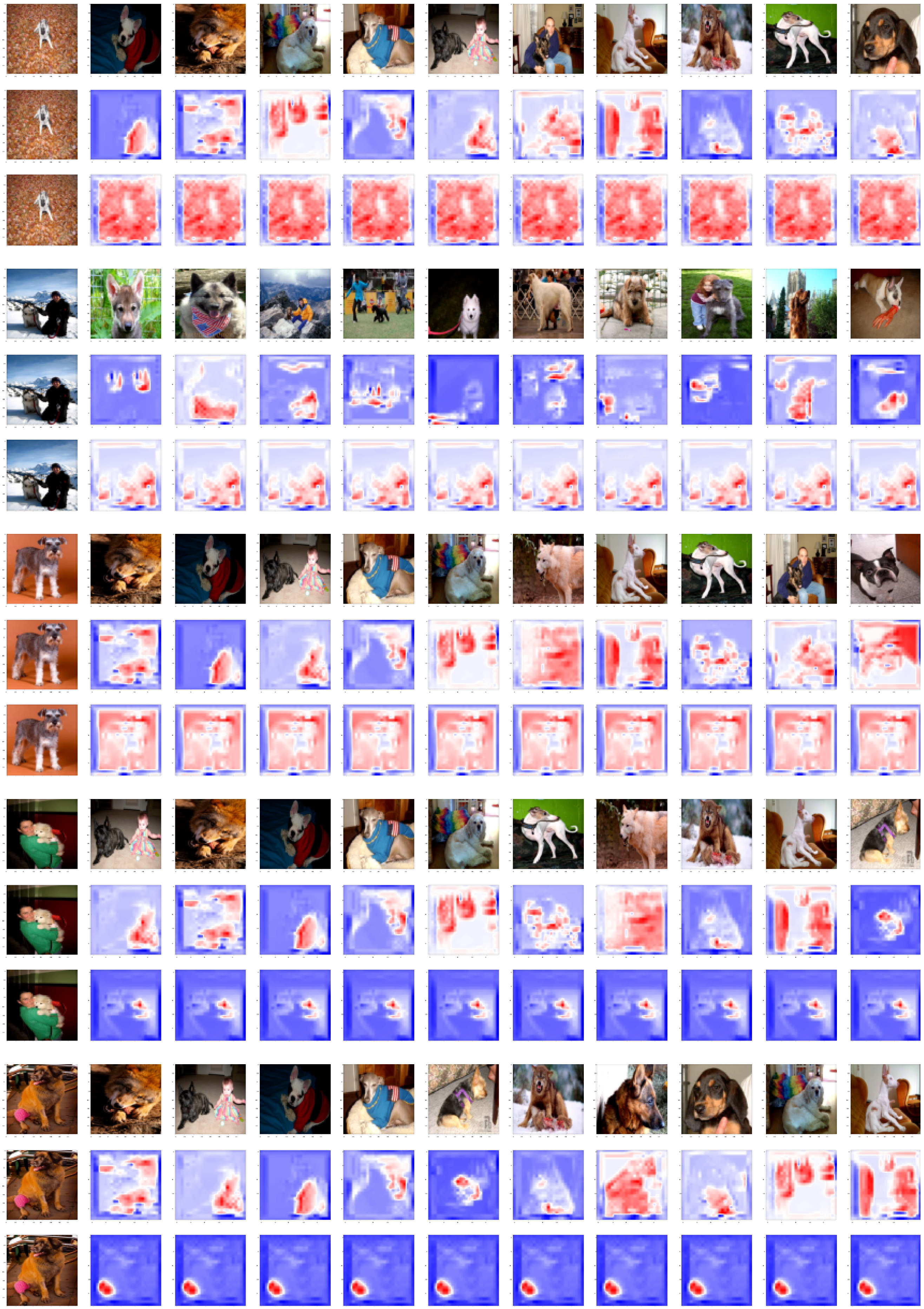

Fig. S31: Explanations for DogsWolves (set 7).

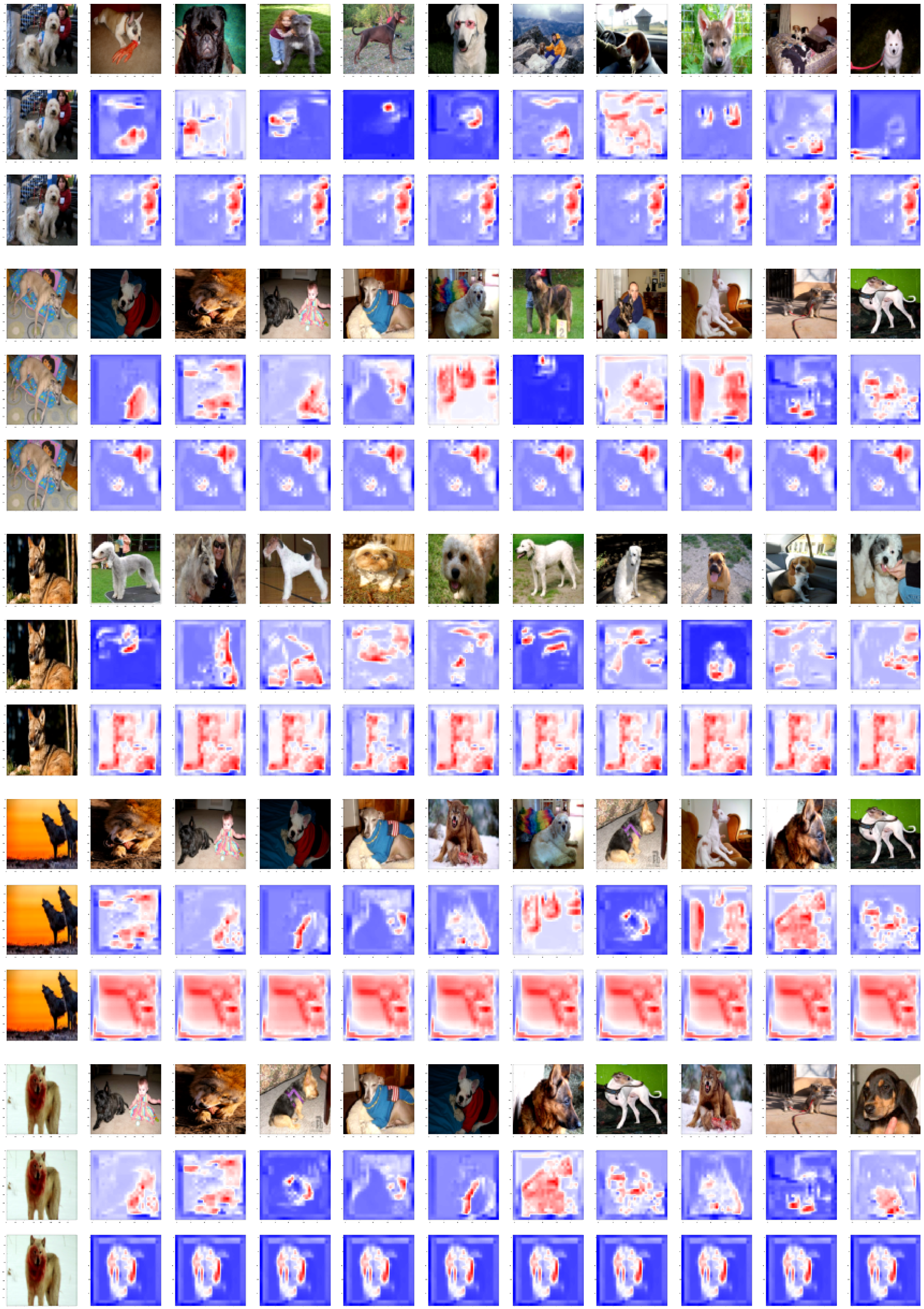

Fig. S32: Explanations for DogsWolves (set 8).

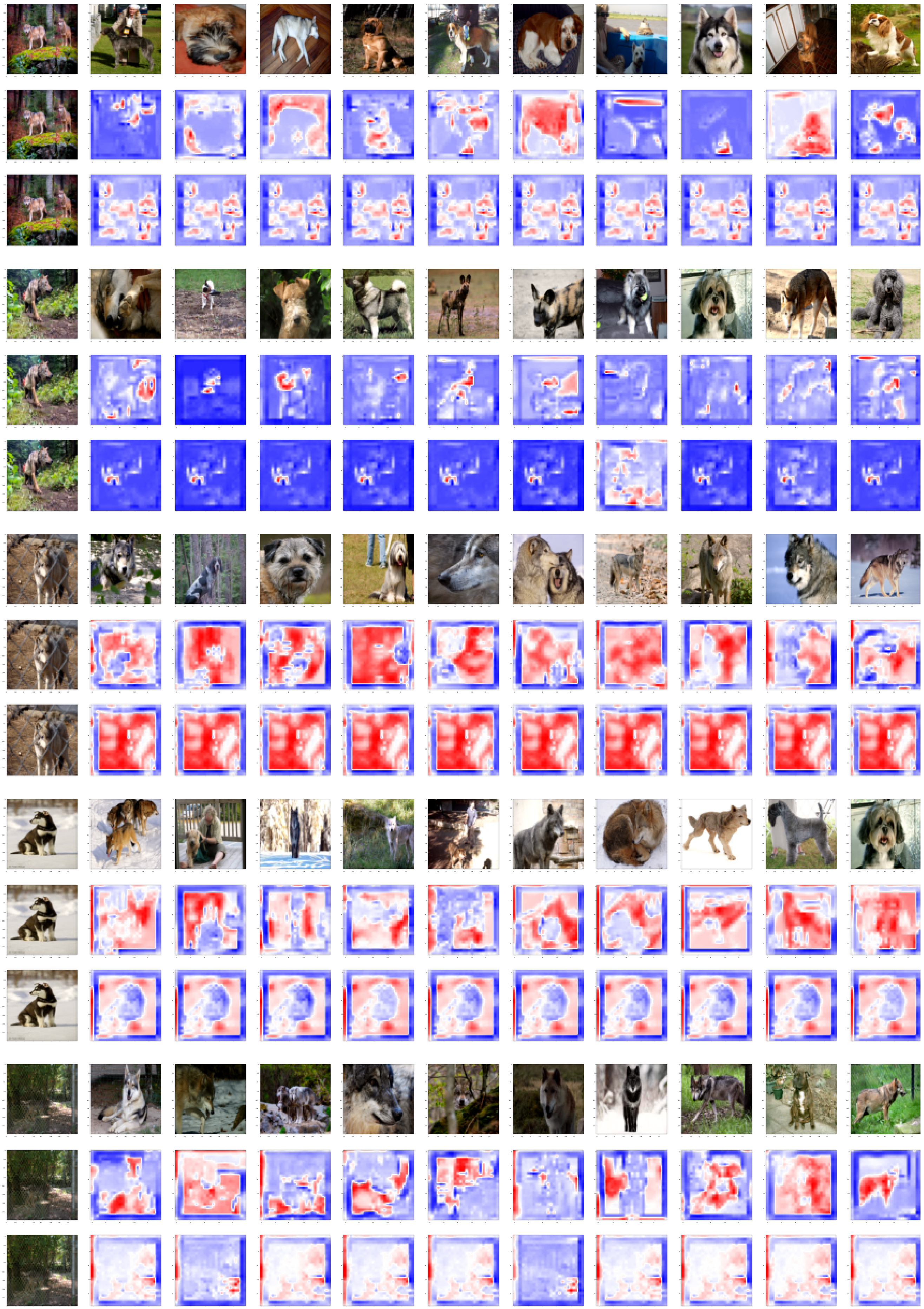

Fig. S33: Explanations for DogsWolves (set 9).

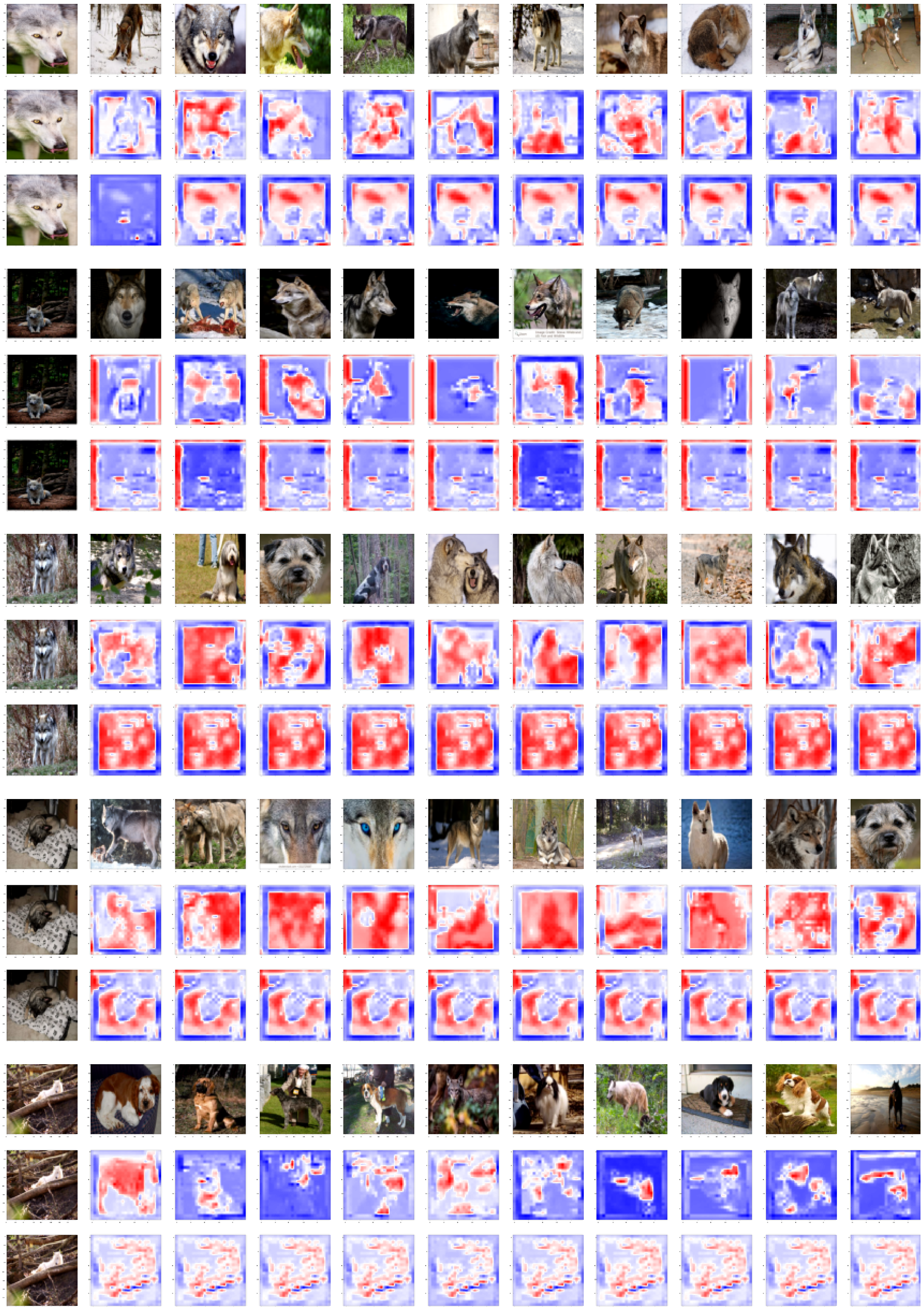

Fig. S34: Explanations for DogsWolves (set 10).

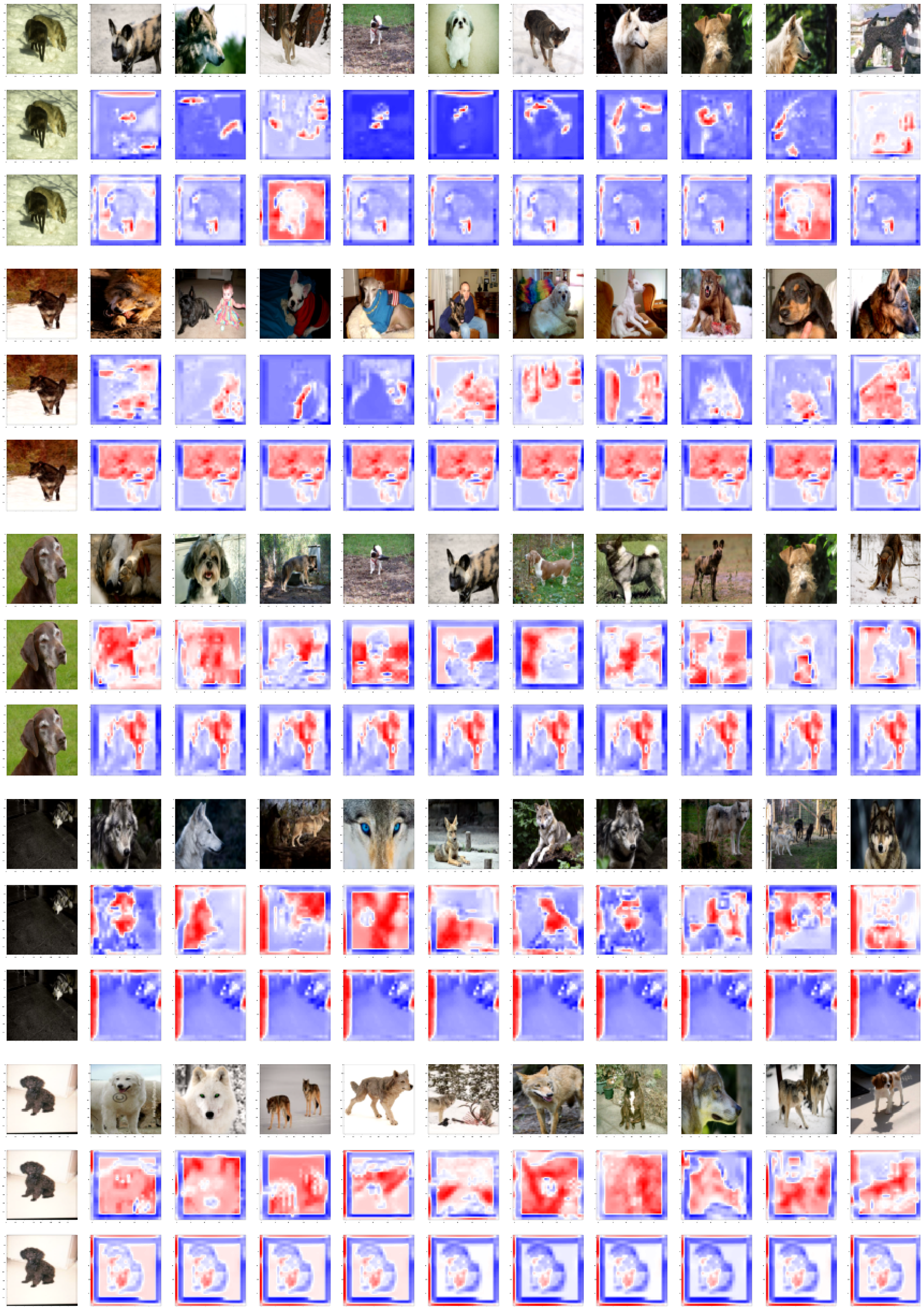

Fig. S35: Explanations for DogsWolves (set 11).

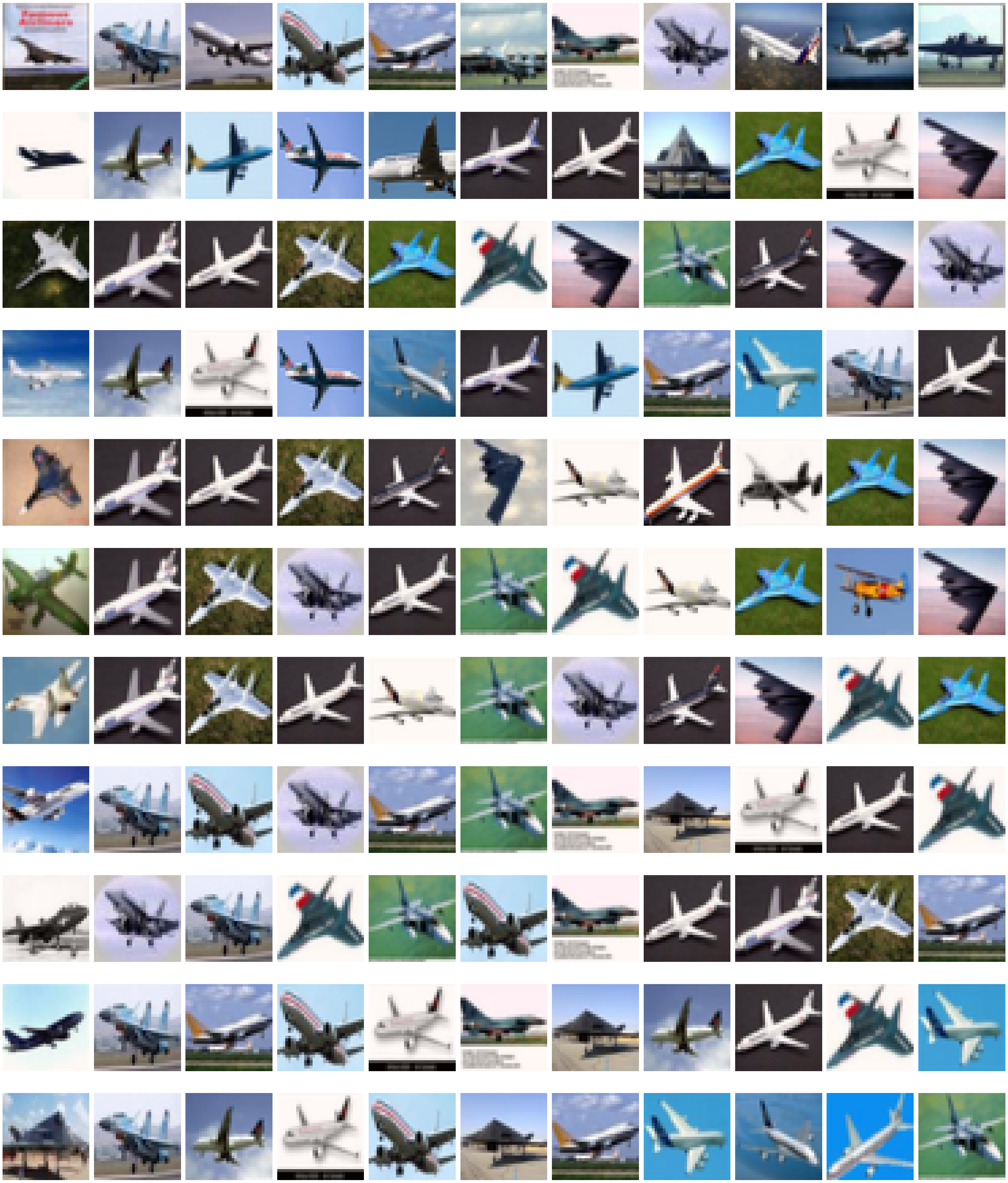

Fig. S36: Explanations for Cifar10 (set 1).

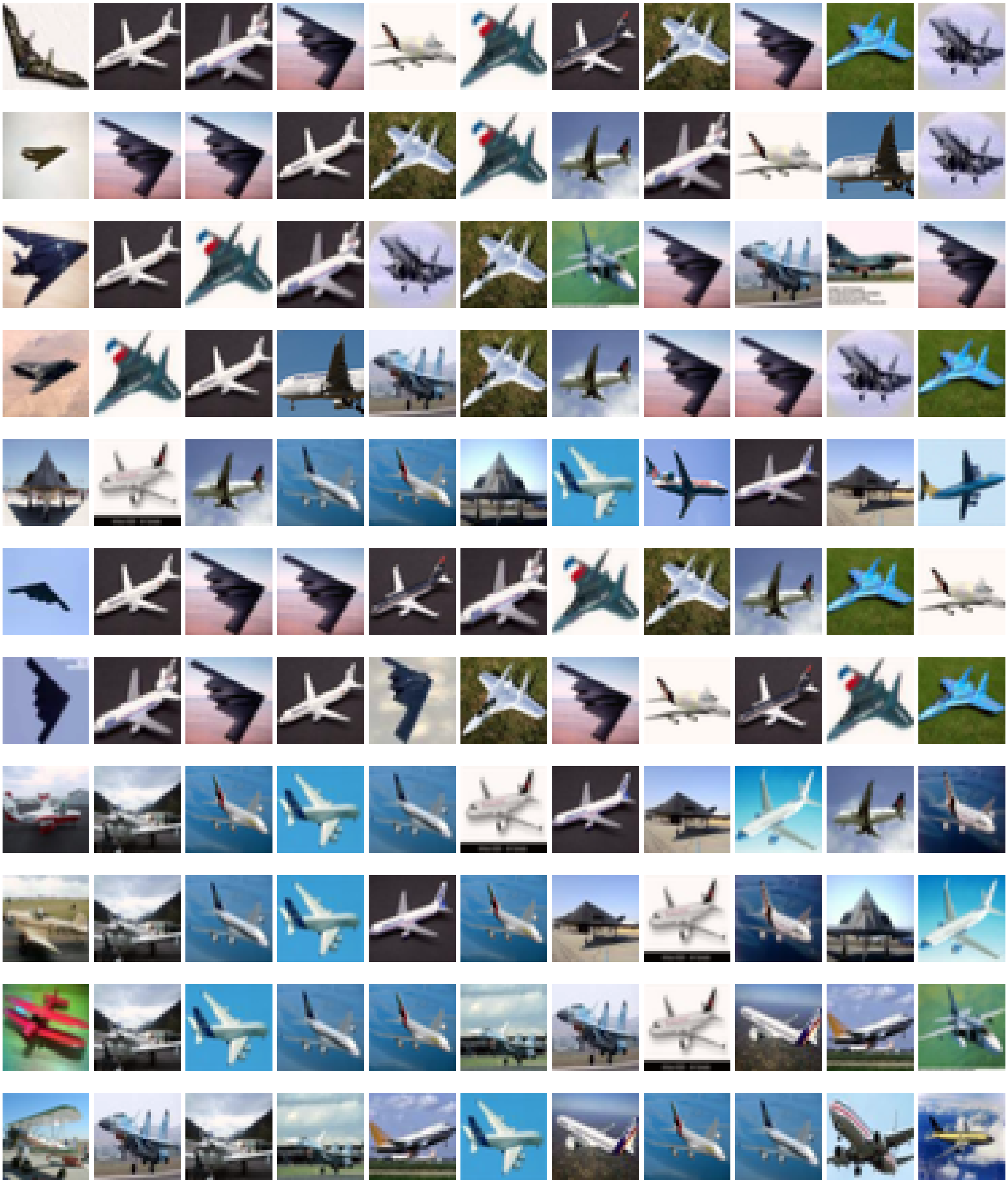

Fig. S37: Explanations for Cifar10 (set 2).

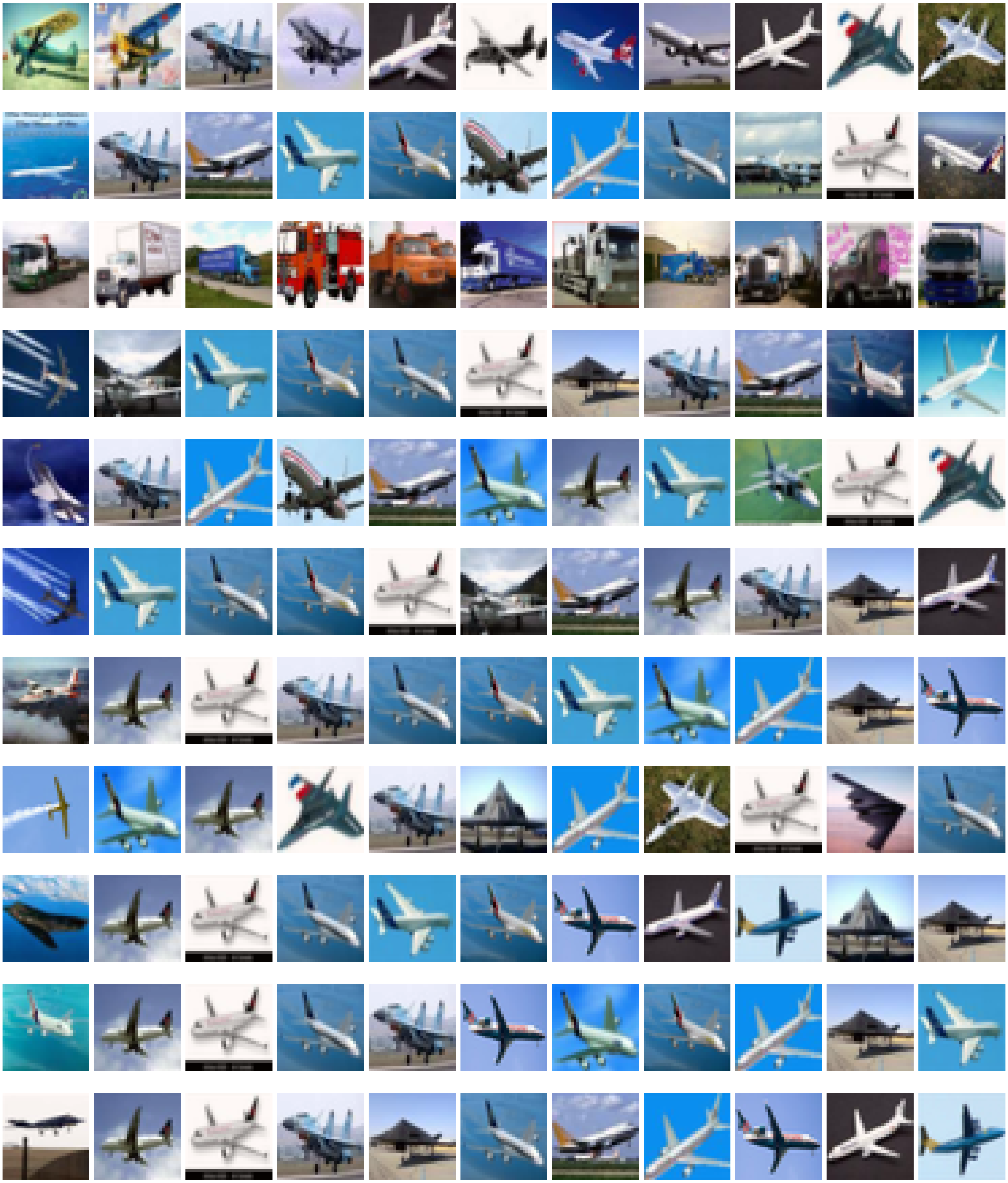

Fig. S38: Explanations for Cifar10 (set 3).

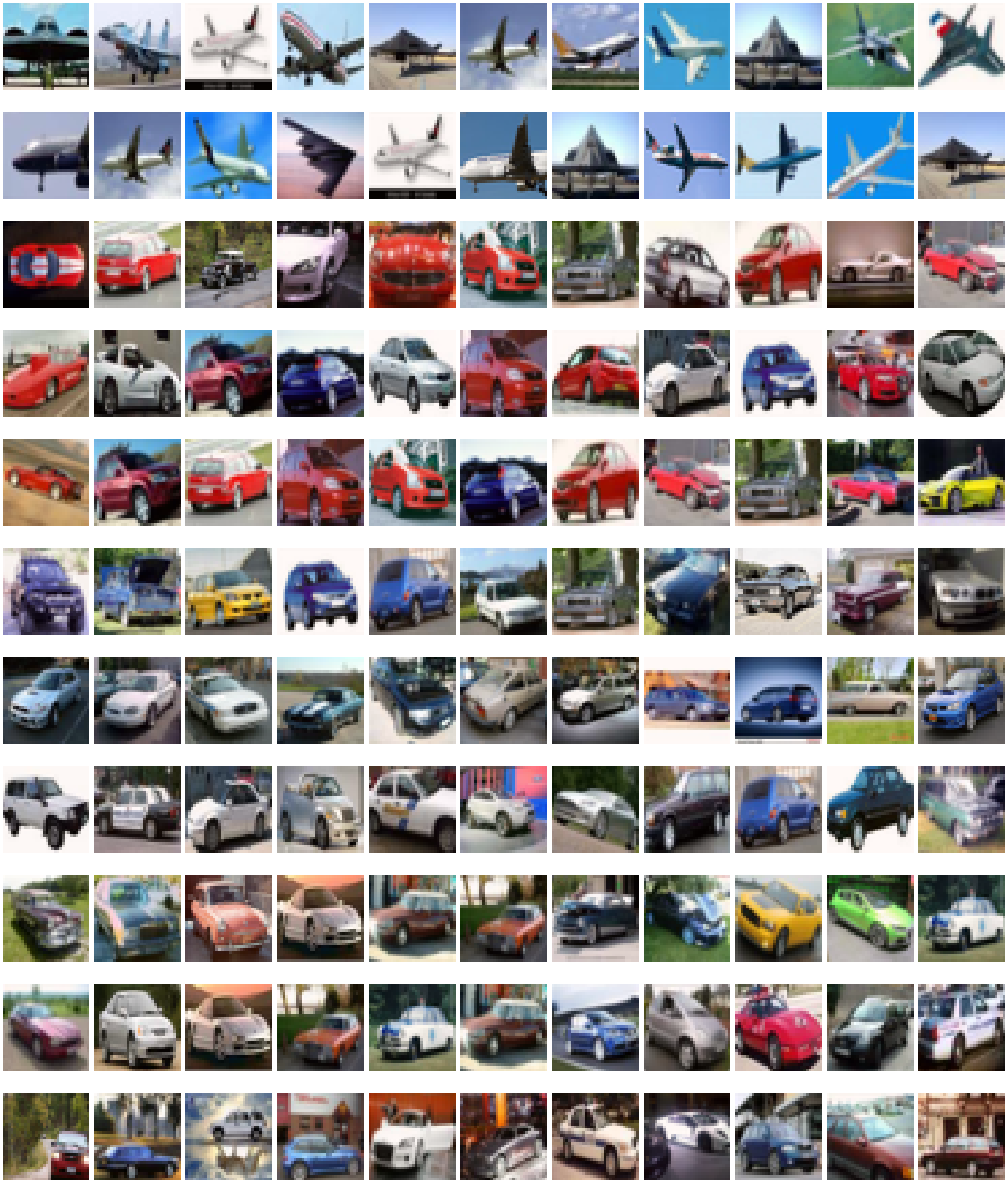

Fig. S39: Explanations for Cifar10 (set 4).

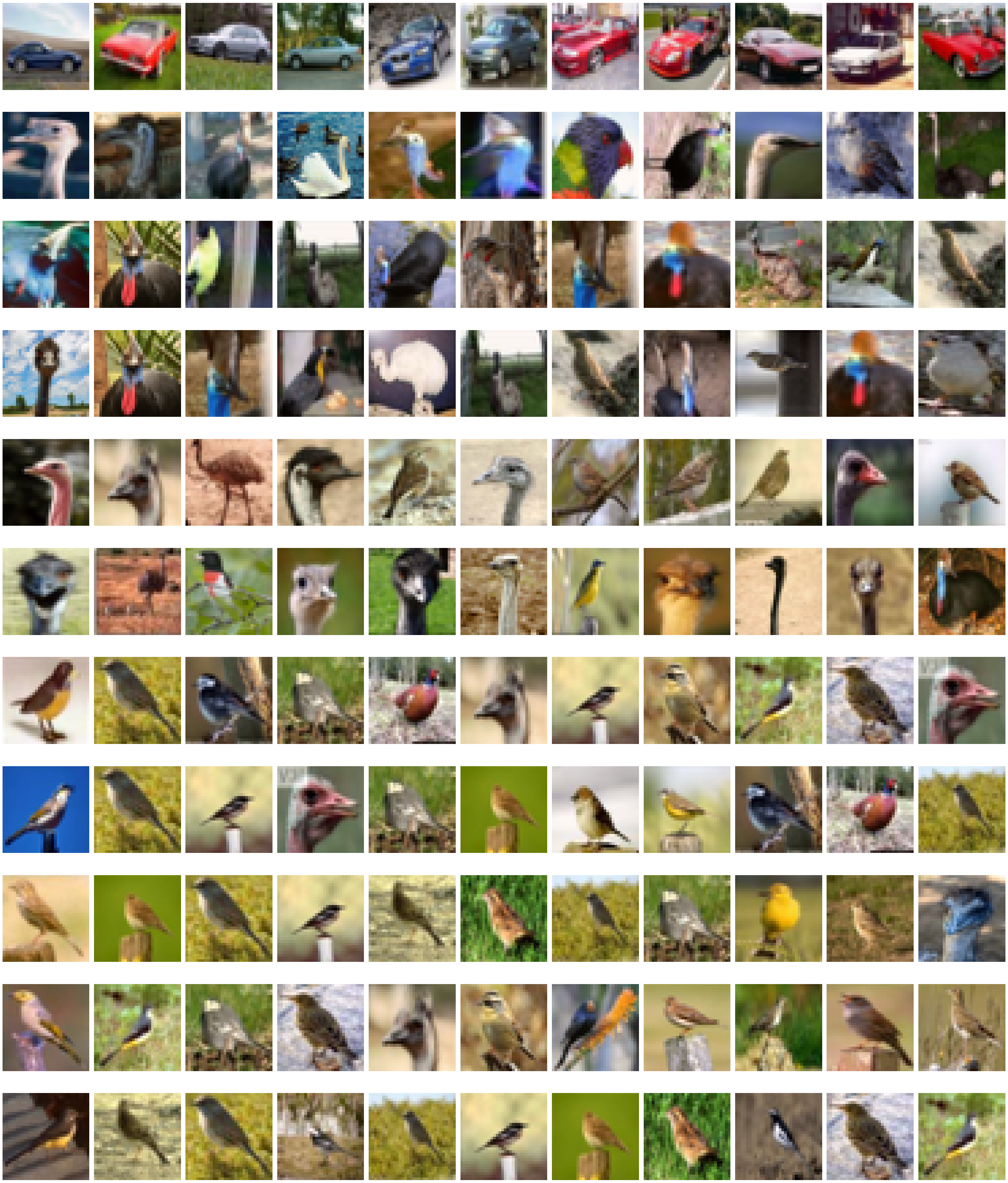

Fig. S40: Explanations for Cifar10 (set 5).

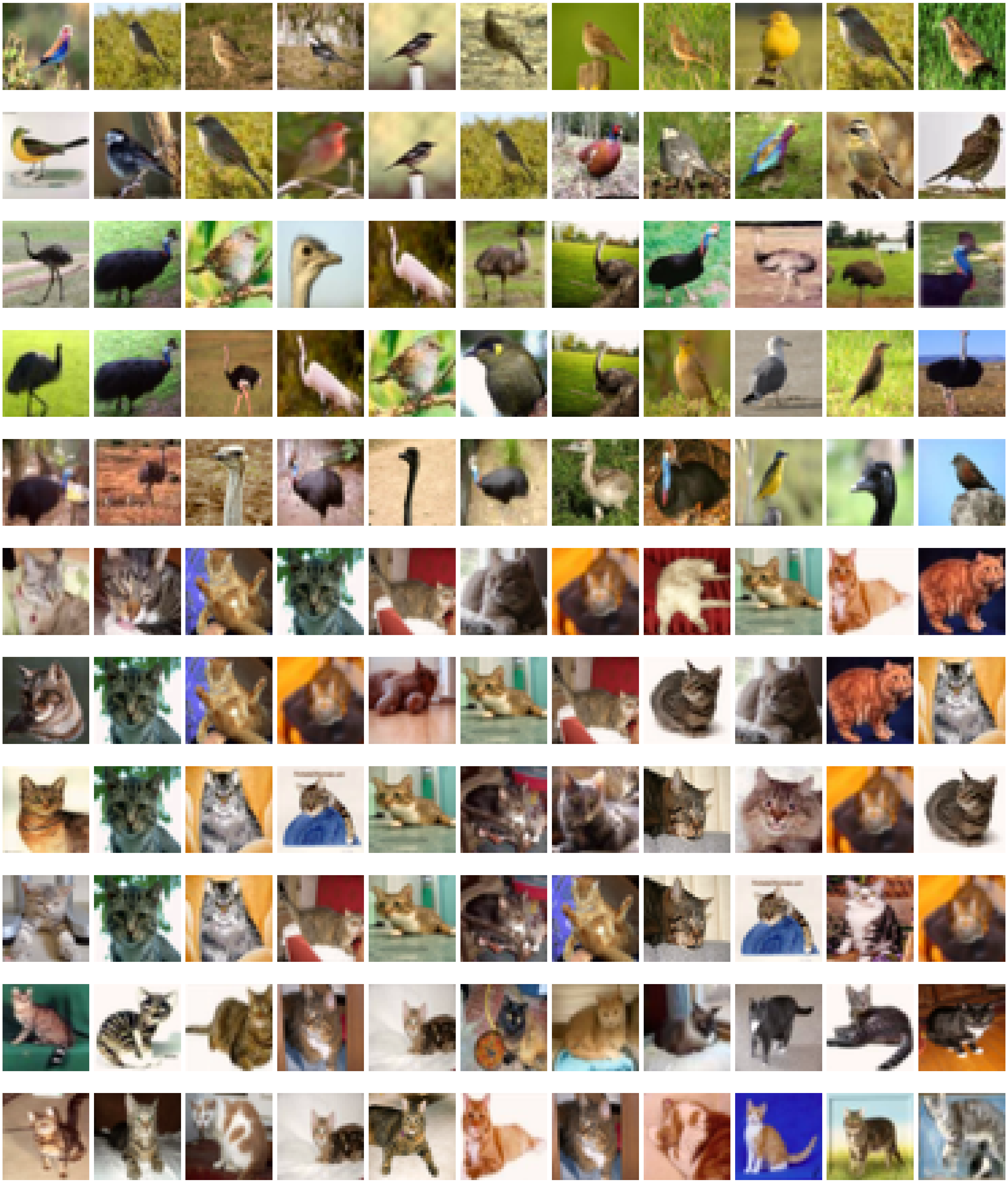

Fig. S41: Explanations for Cifar10 (set 6).

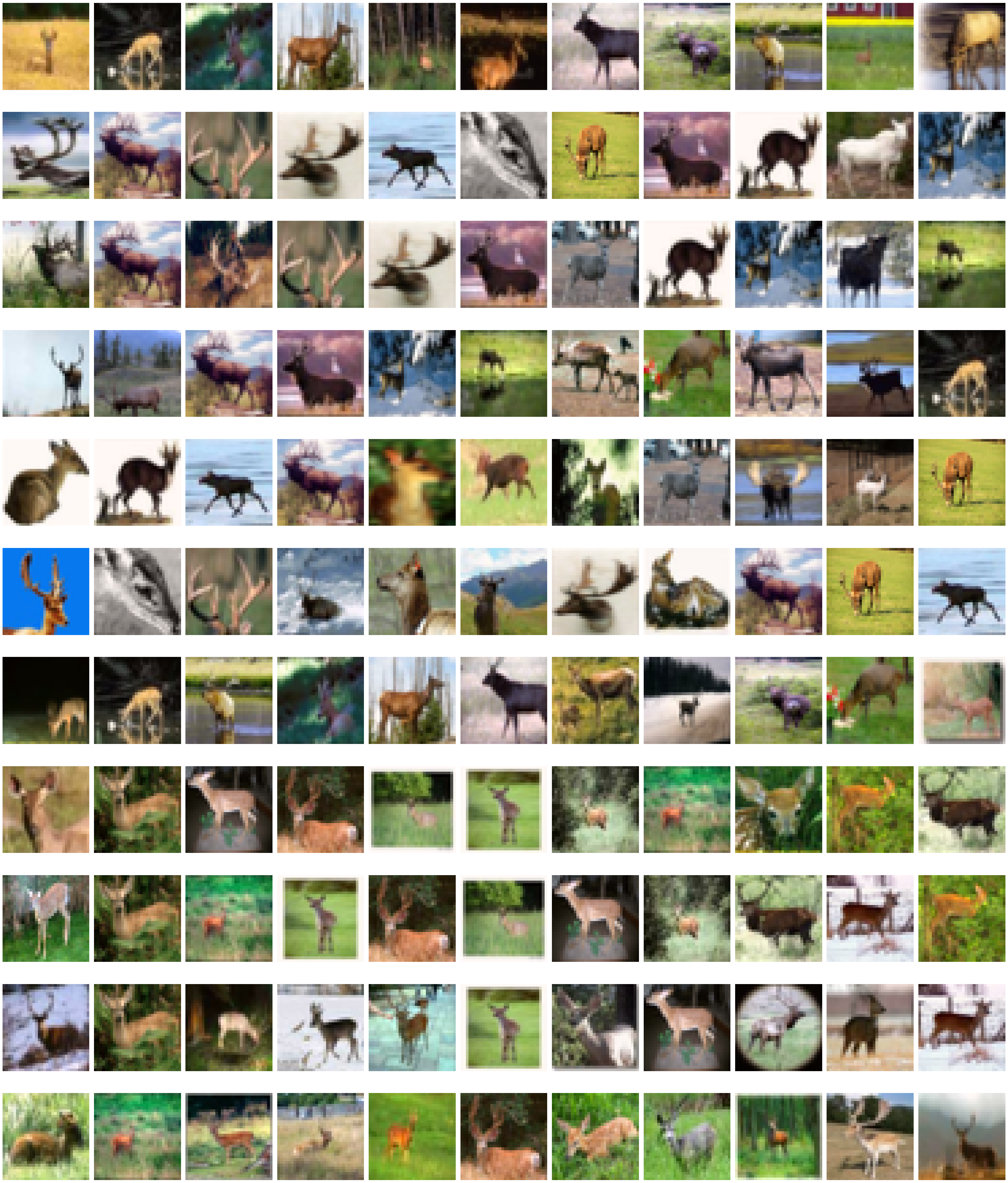

Fig. S42: Explanations for Cifar10 (set 7).

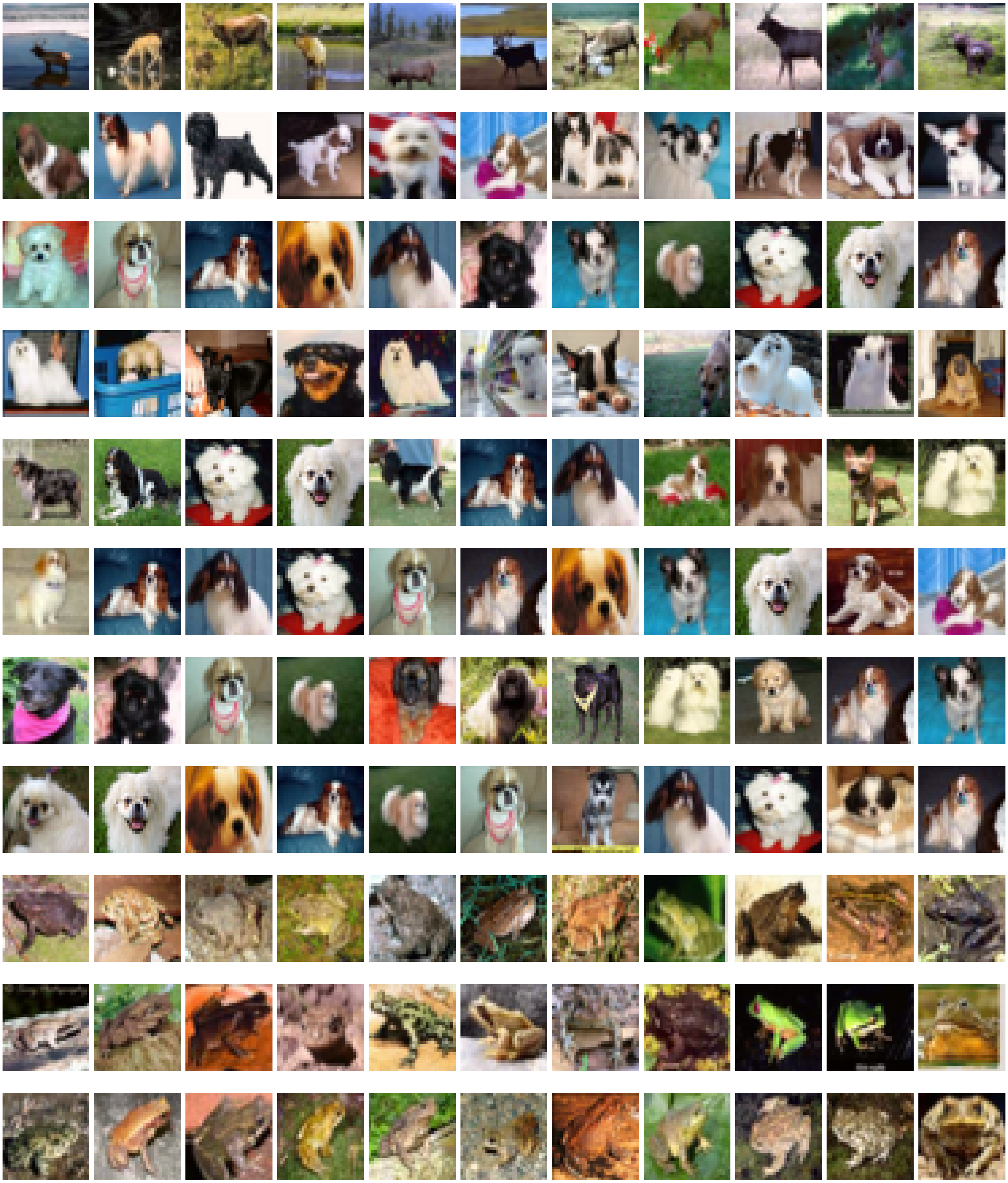

Fig. S43: Explanations for Cifar10 (set 8).

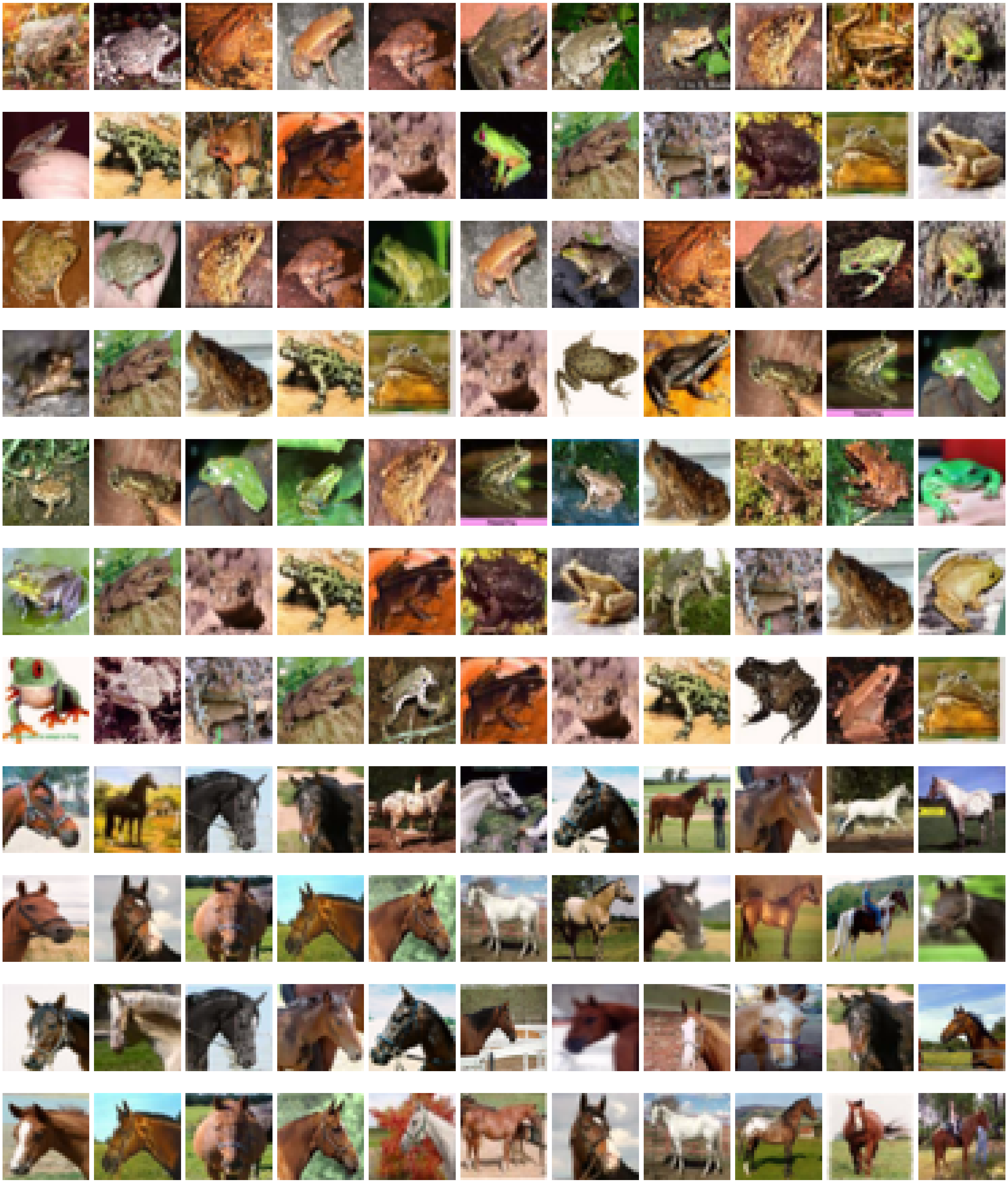

Fig. S44: Explanations for Cifar10 (set 9).

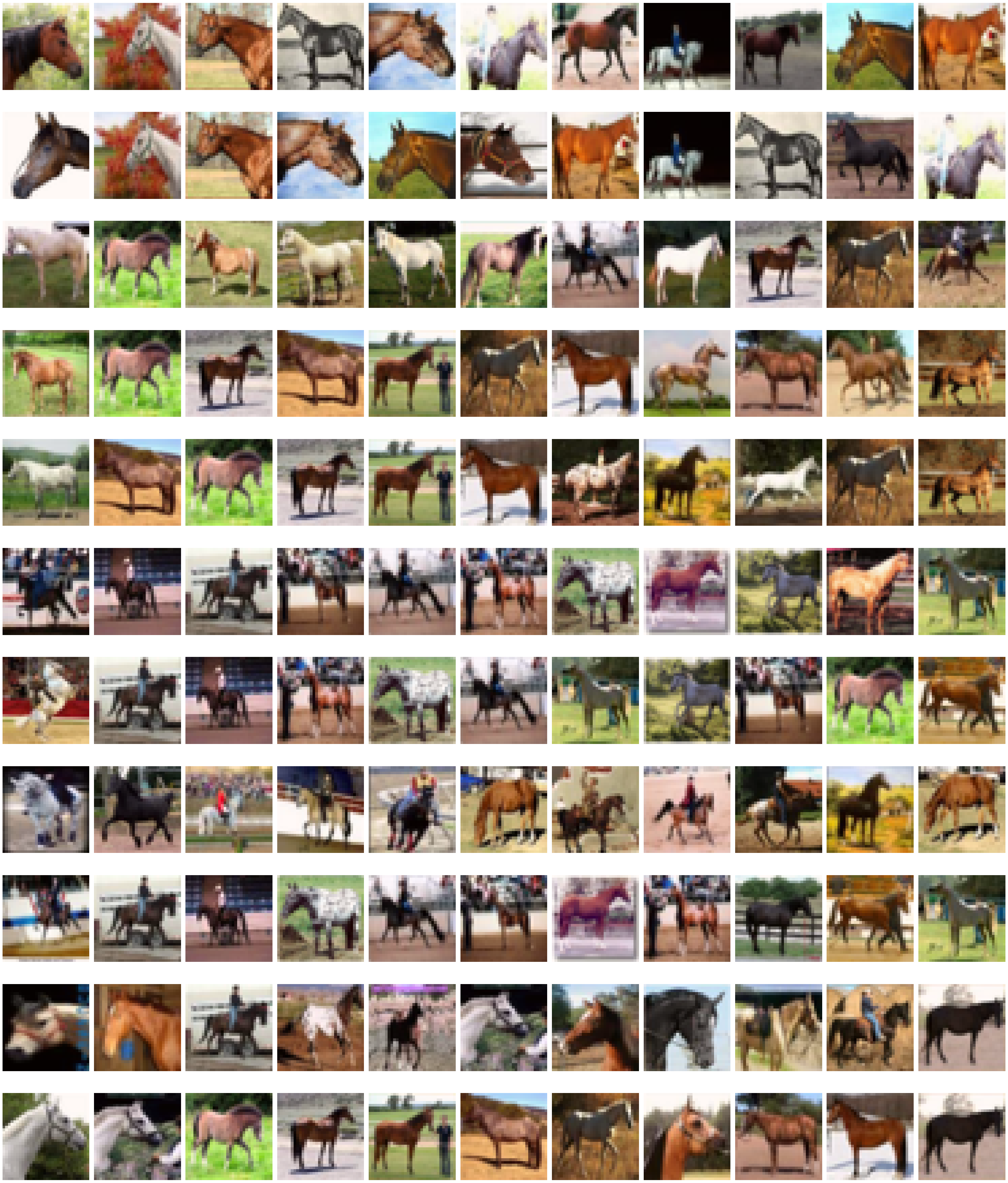

Fig. S45: Explanations for Cifar10 (set 10).

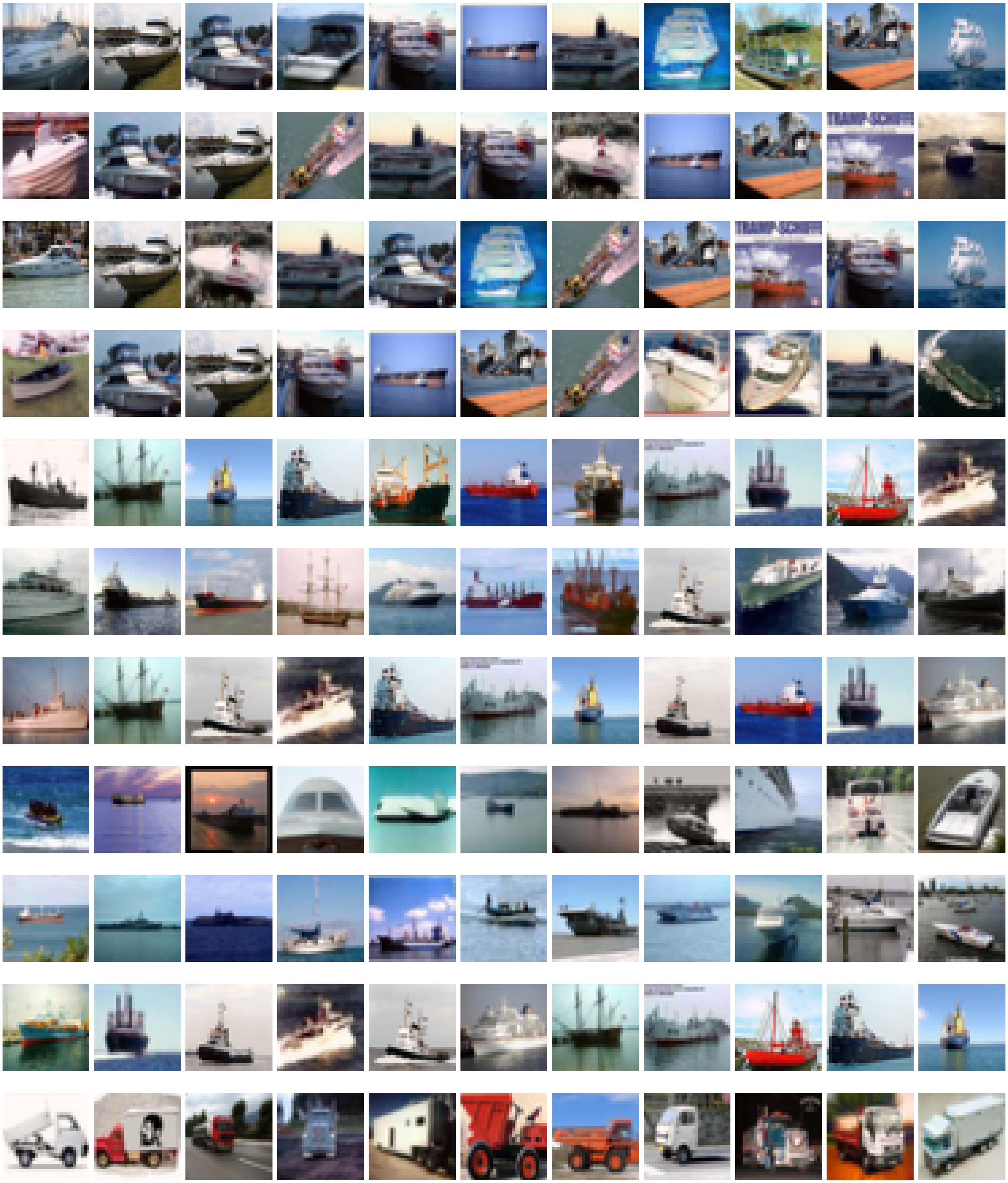

Fig. S46: Explanations for Cifar10 (set 11).

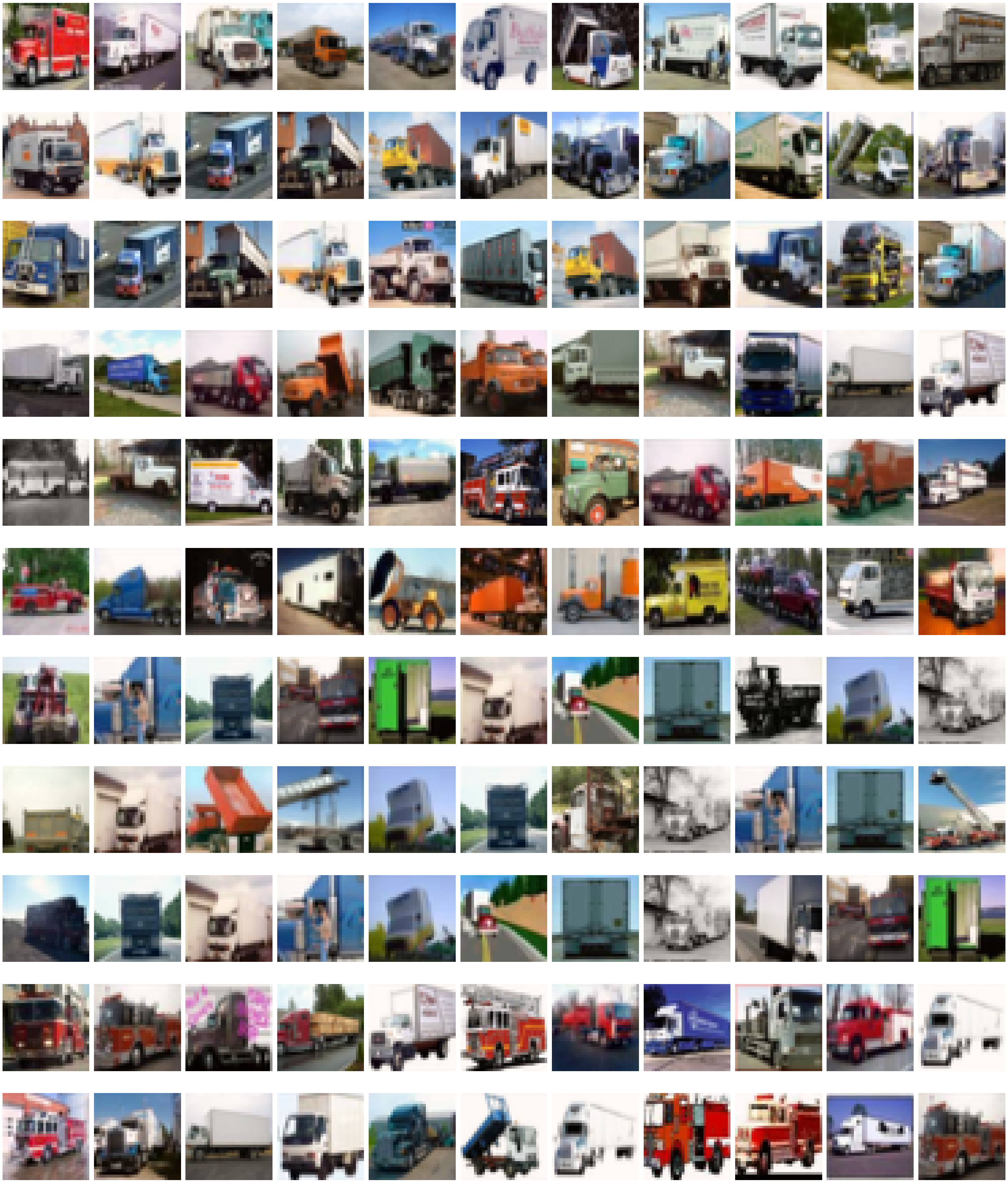

Fig. S47: Explanations for Cifar10 (set 12).

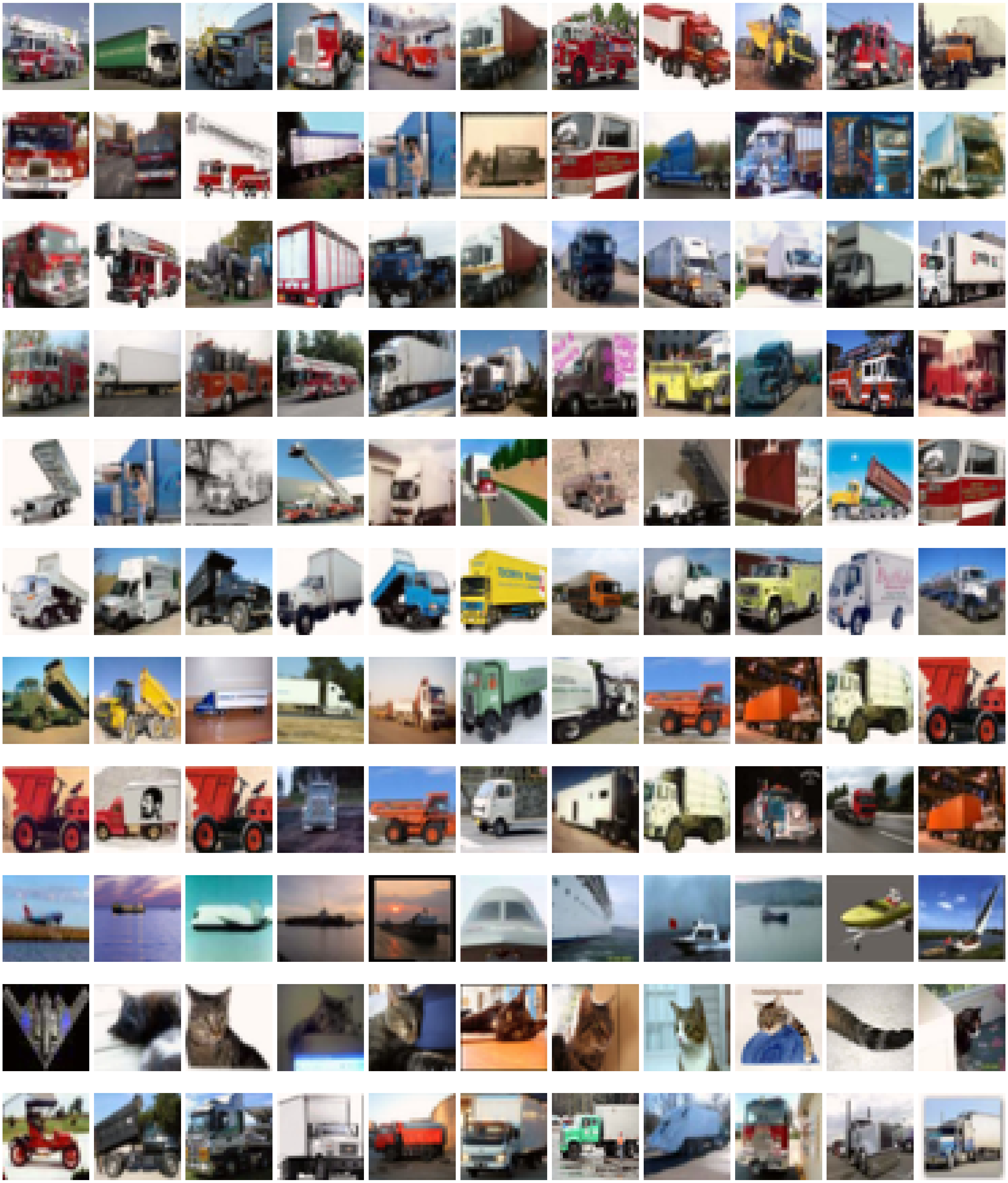

Fig. S48: Explanations for Cifar10 (set 13).

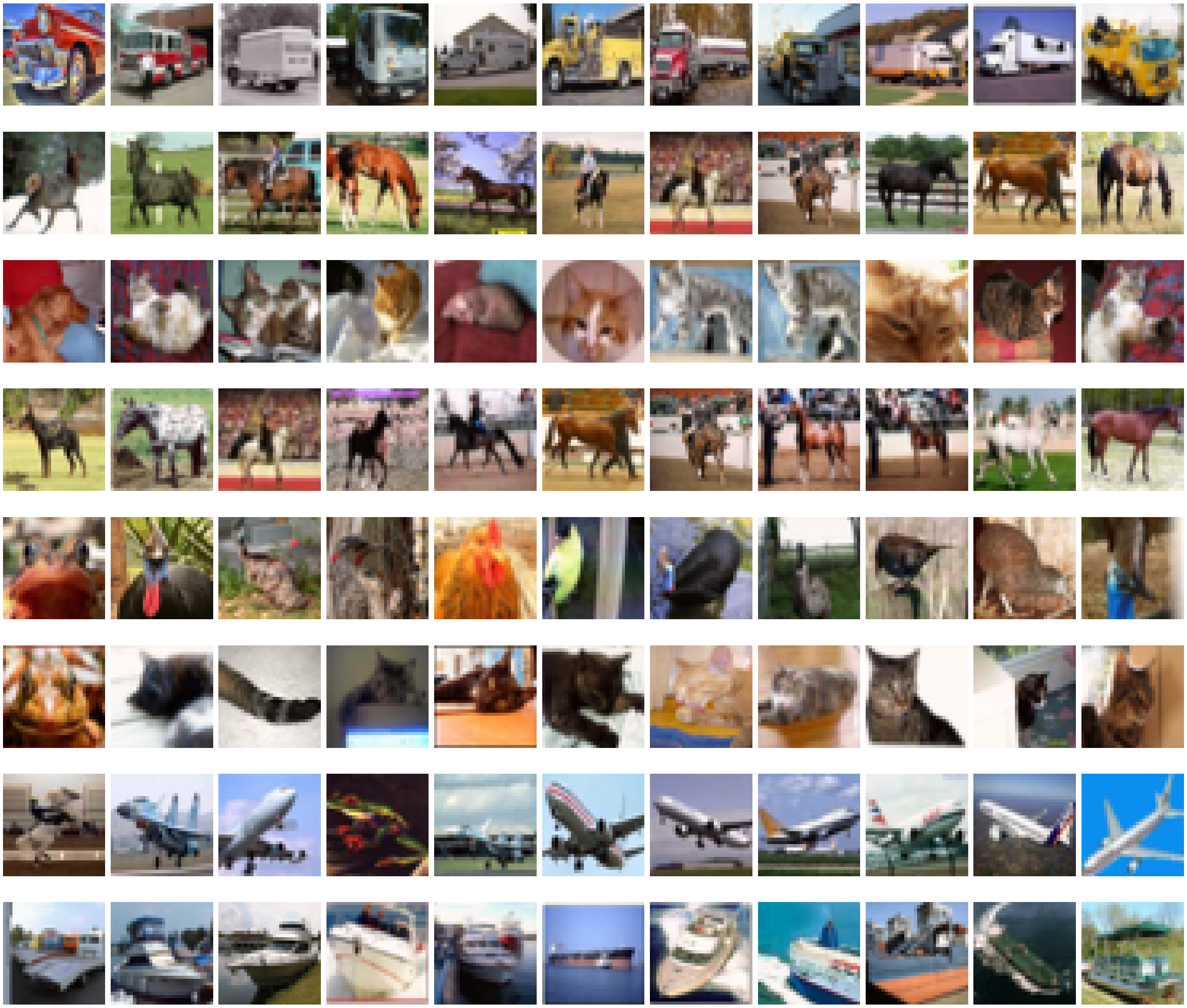

Fig. S49: Explanations for Cifar10 (set 14).

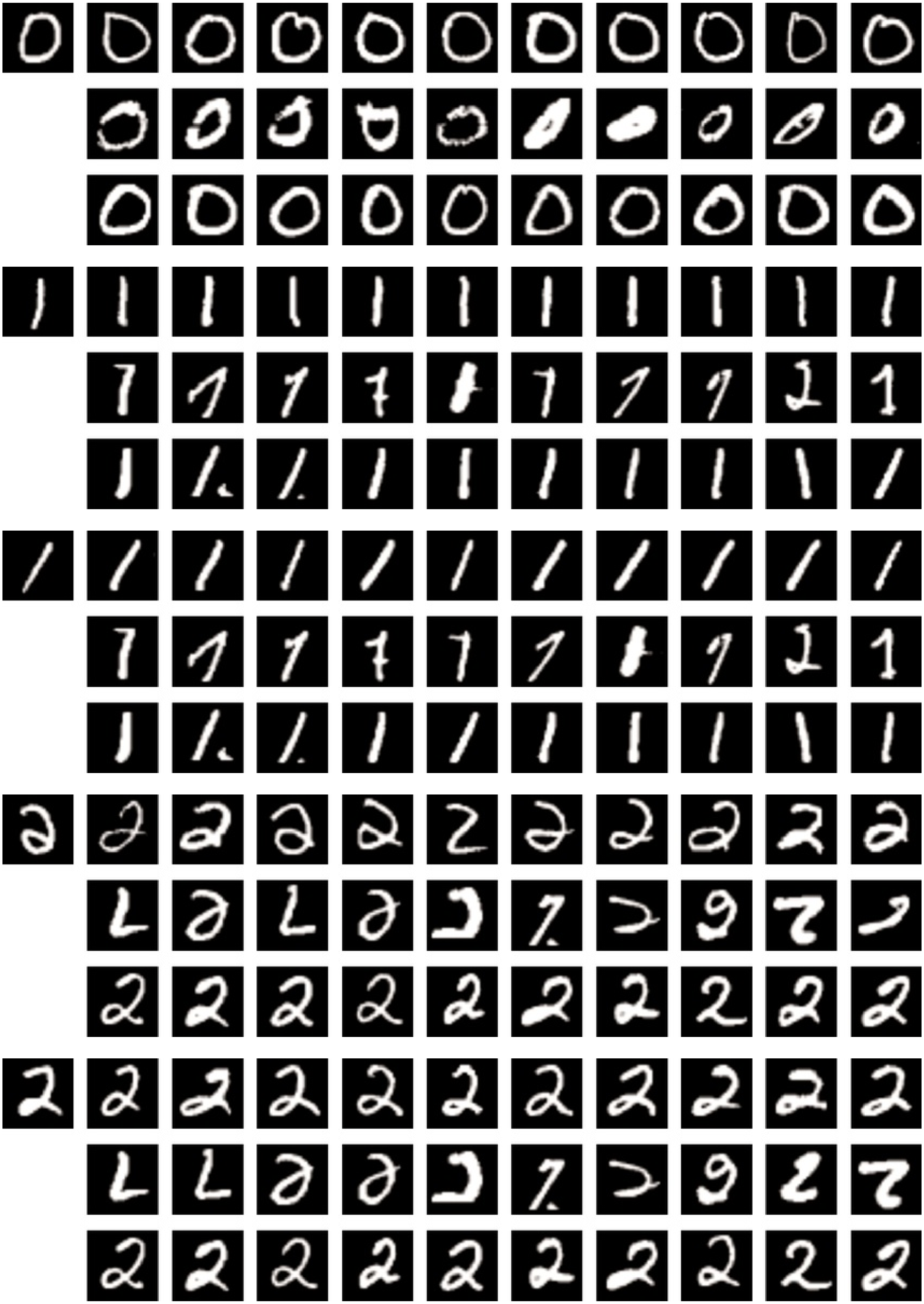

Fig. S50: Comparing the proposed GPEX with representer point selection [33] (set 1).

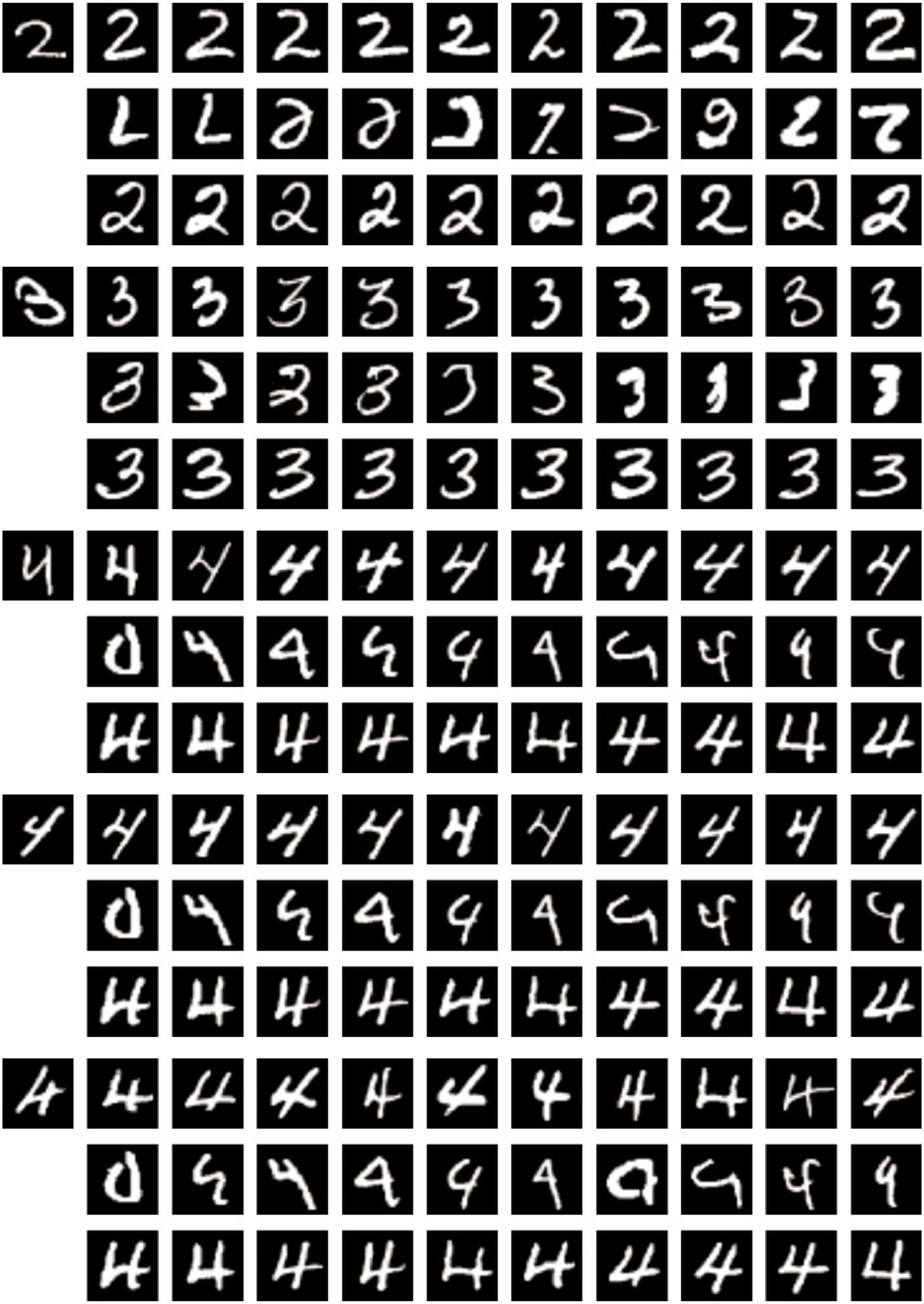

Fig. S51: Comparing the proposed GPEX with representer point selection [33] (set 2).

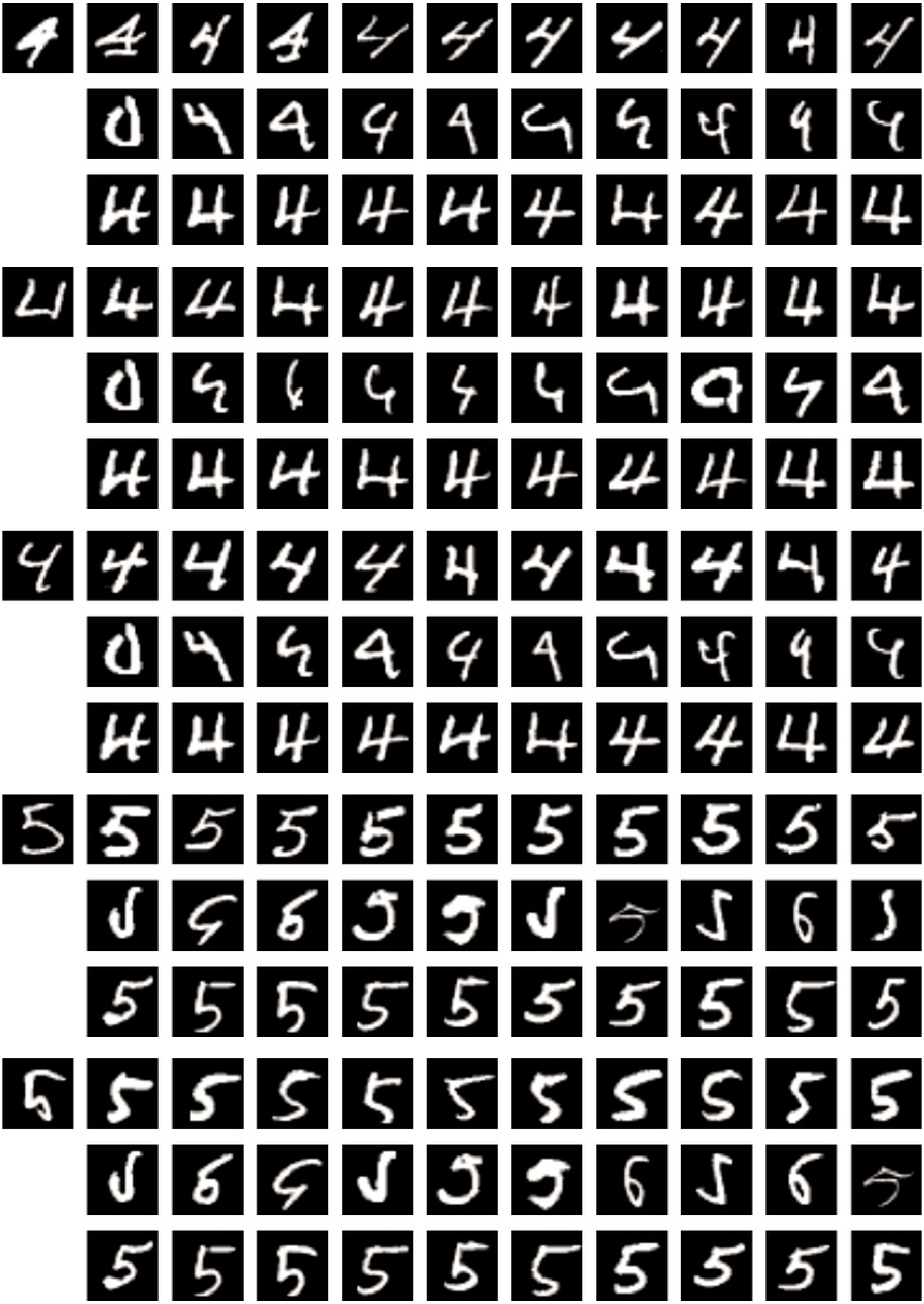

Fig. S52: Comparing the proposed GPEX with representer point selection [33] (set 3).

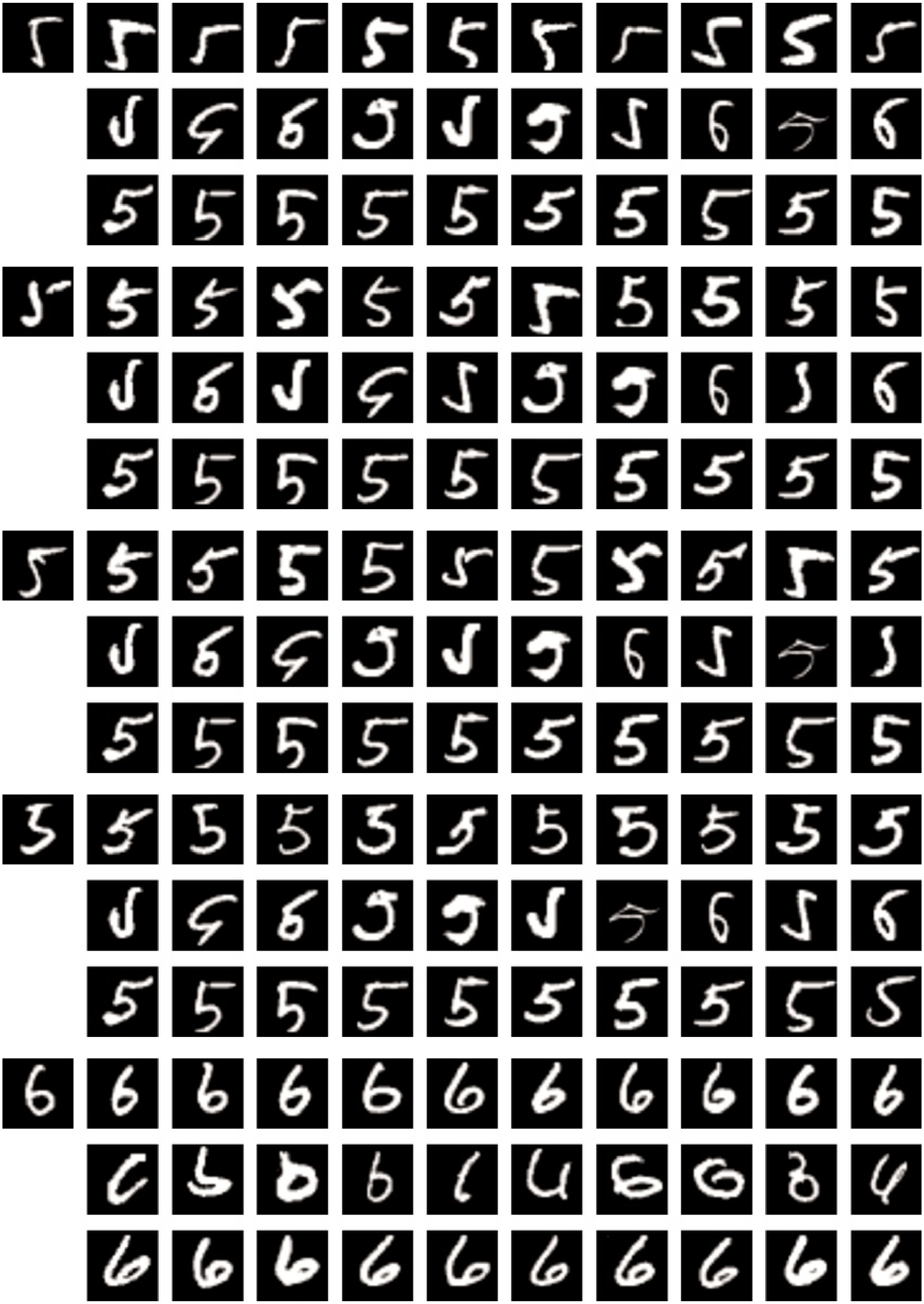

Fig. S53: Comparing the proposed GPEX with representer point selection [33] (set 4).

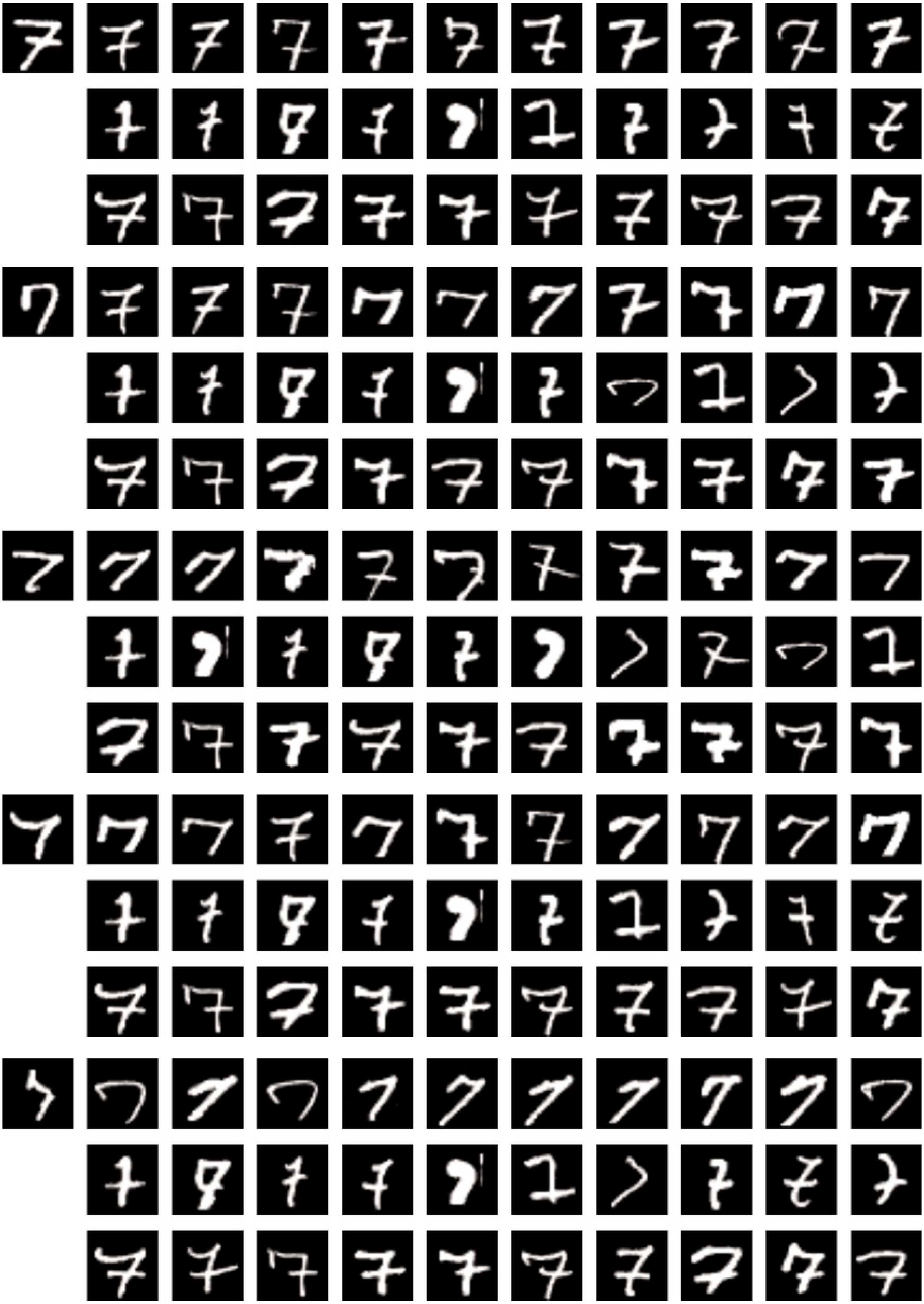

Fig. S54: Comparing the proposed GPEX with representer point selection [33] (set 5).

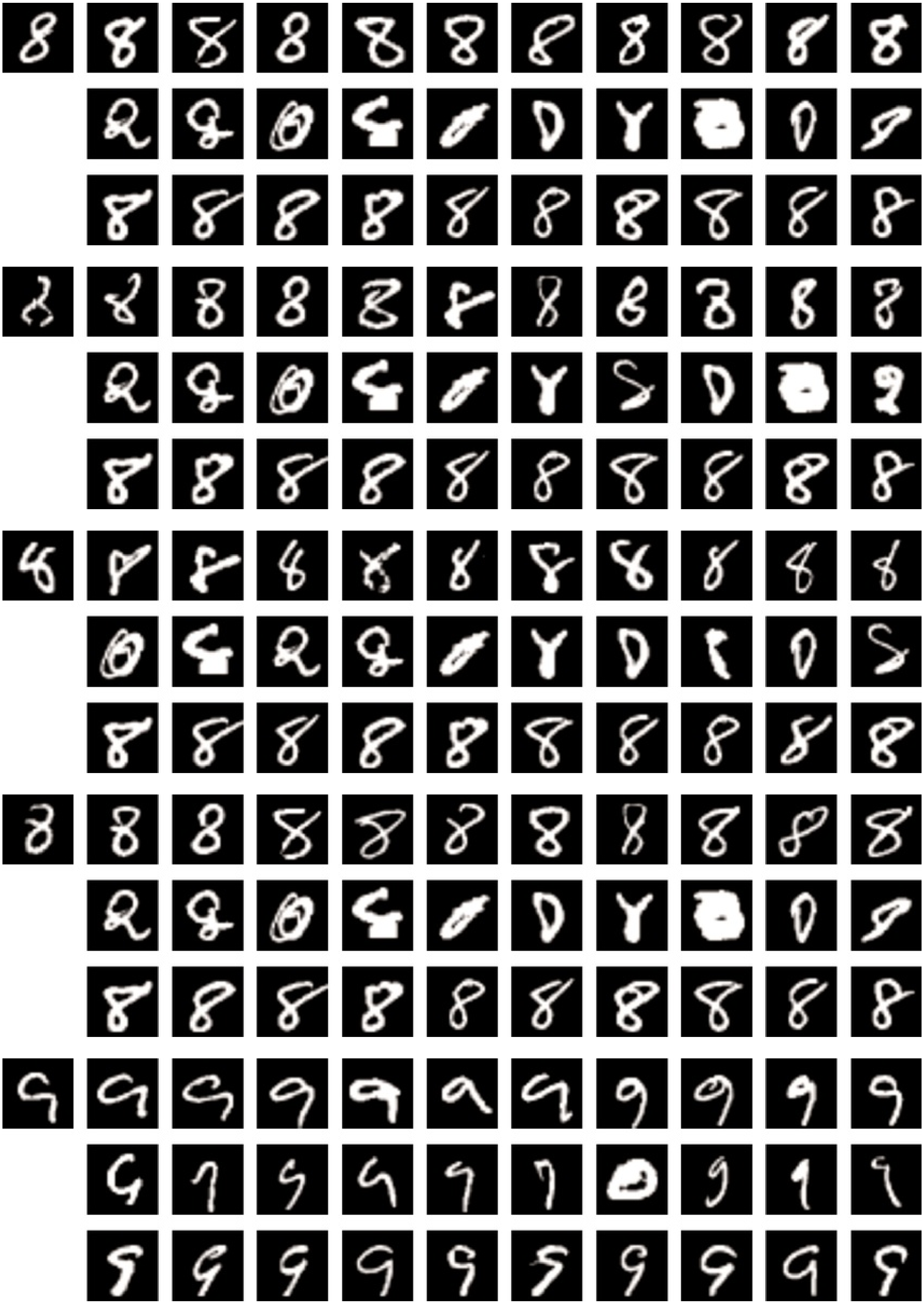

Fig. S55: Comparing the proposed GPEX with representer point selection [33] (set 6).

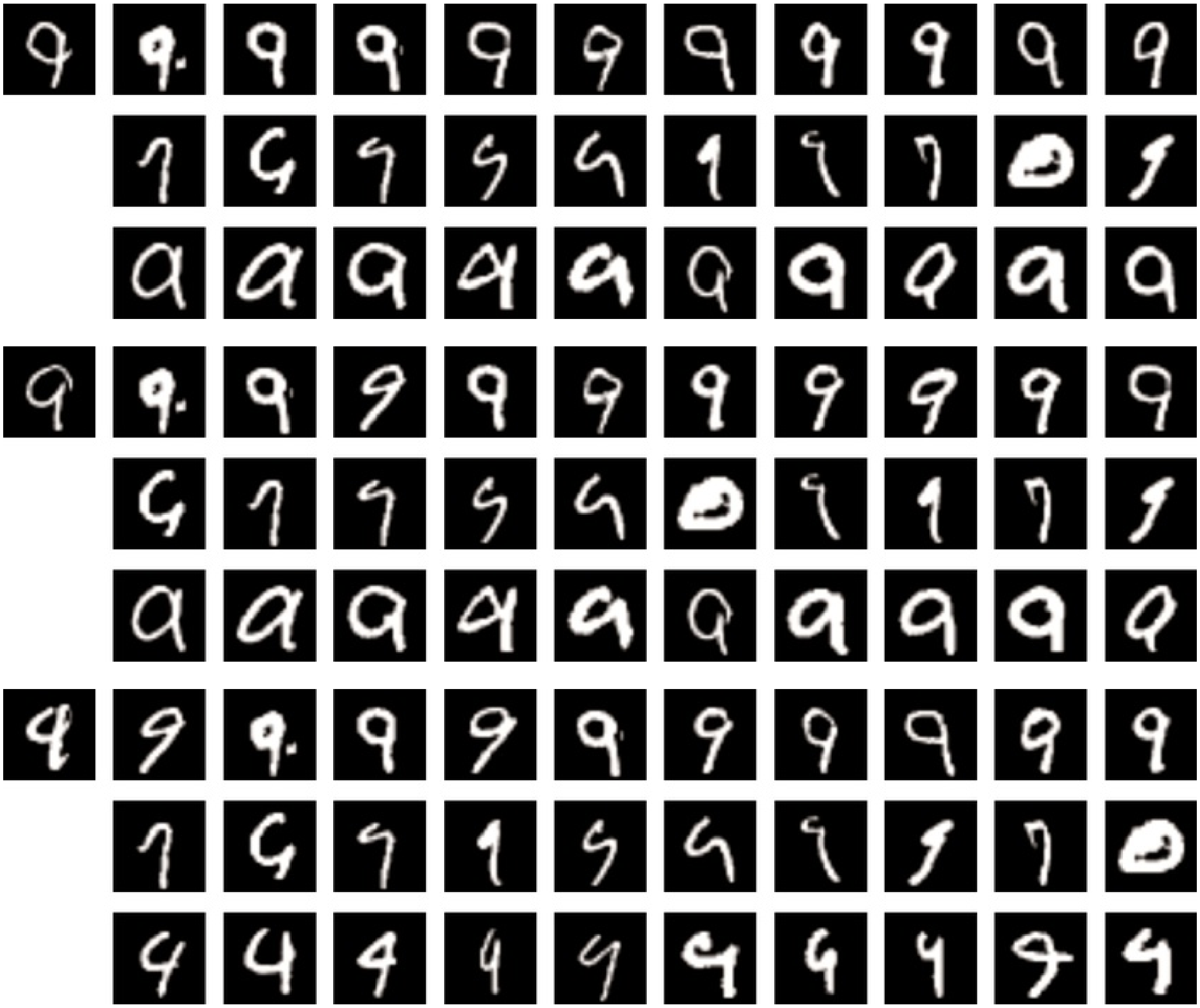

Fig. S56: Comparing the proposed GPEX with representer point selection [33] (set 7).

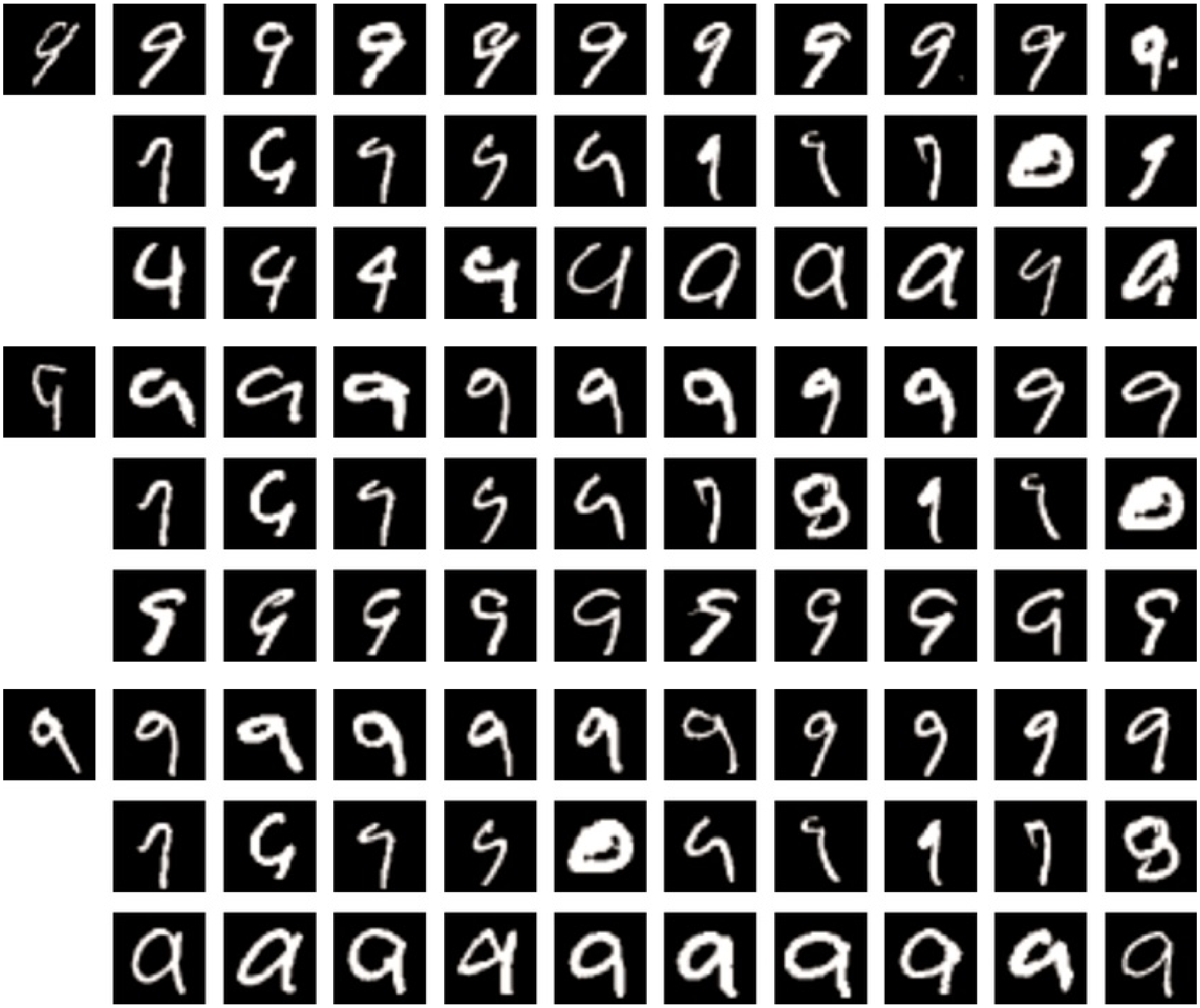

Fig. S57: Comparing the proposed GPEX with representer point selection [33] (set 8).

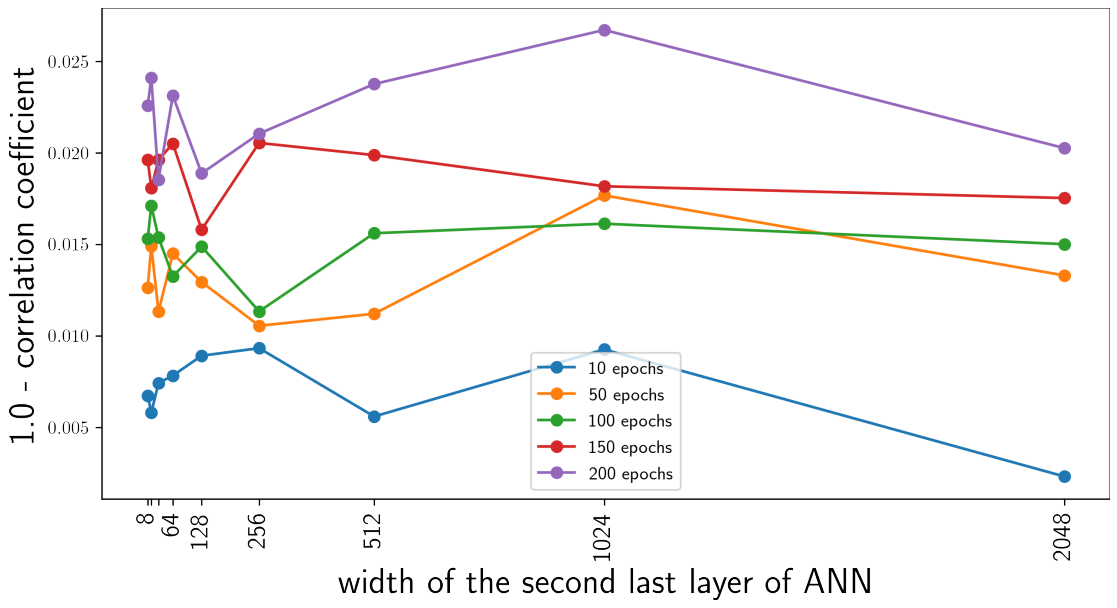

head 1

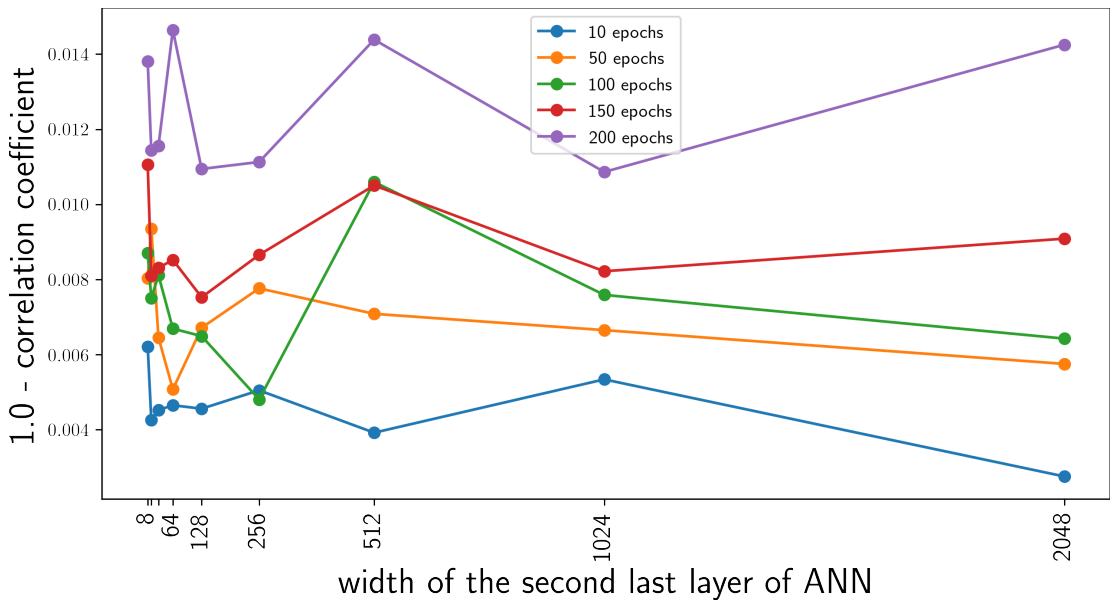

head 2

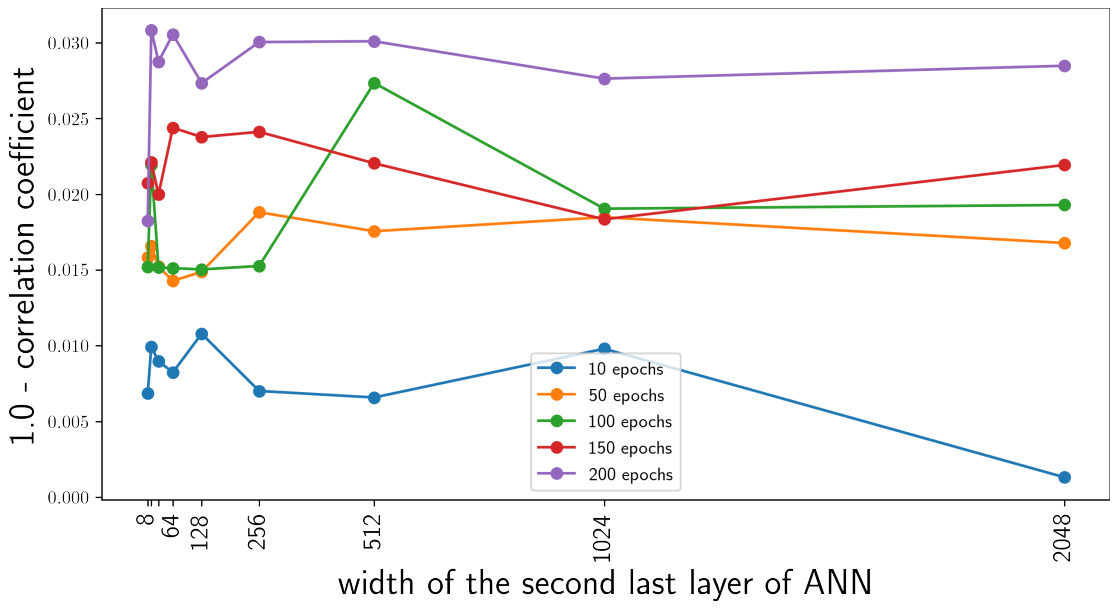

head 3

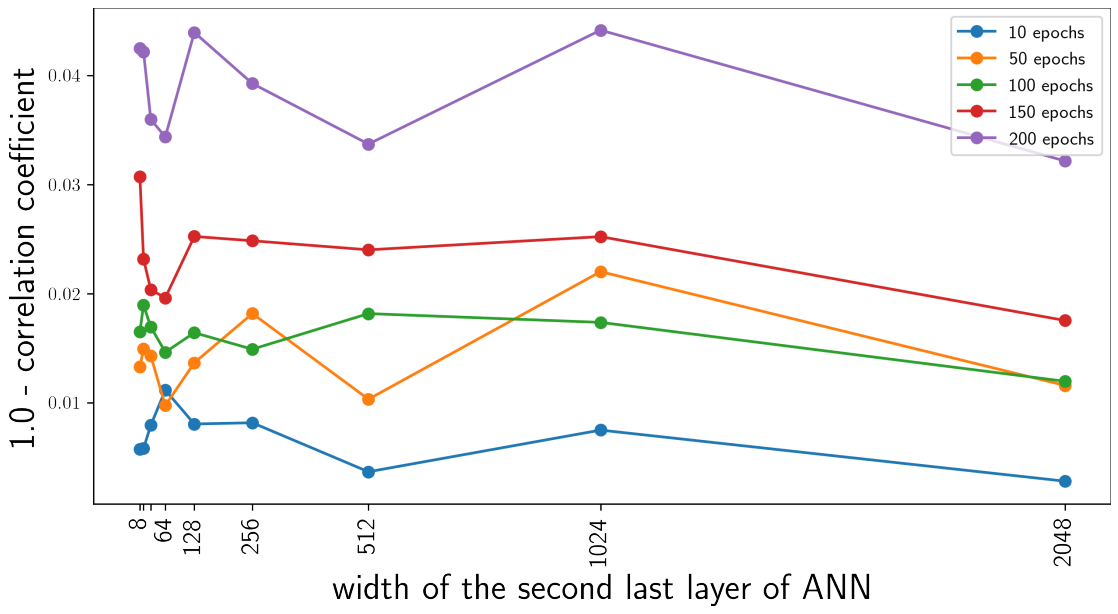

head 4

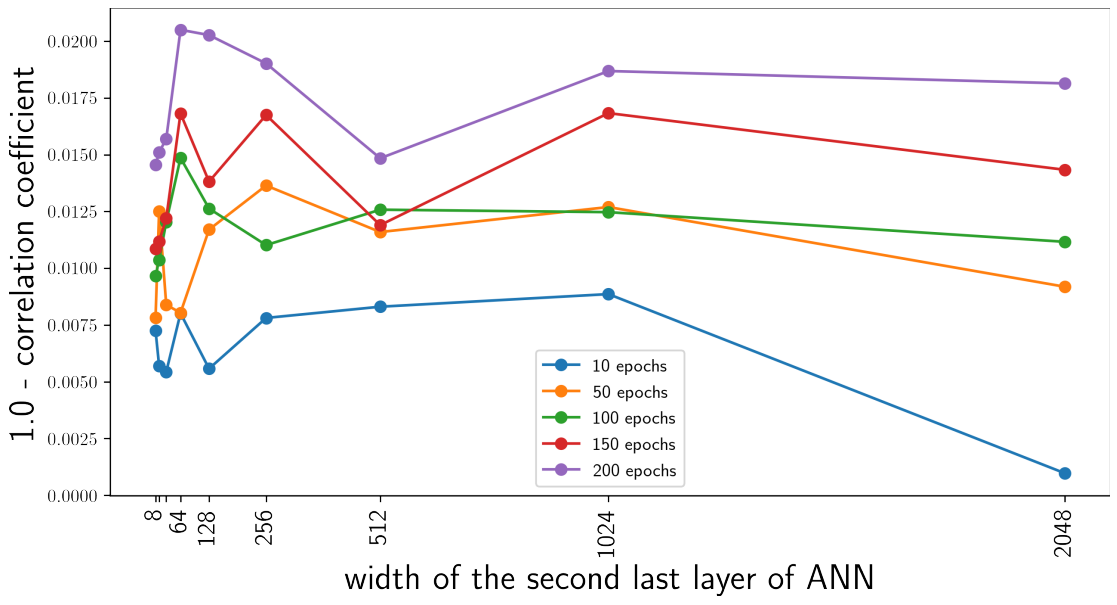

head 5

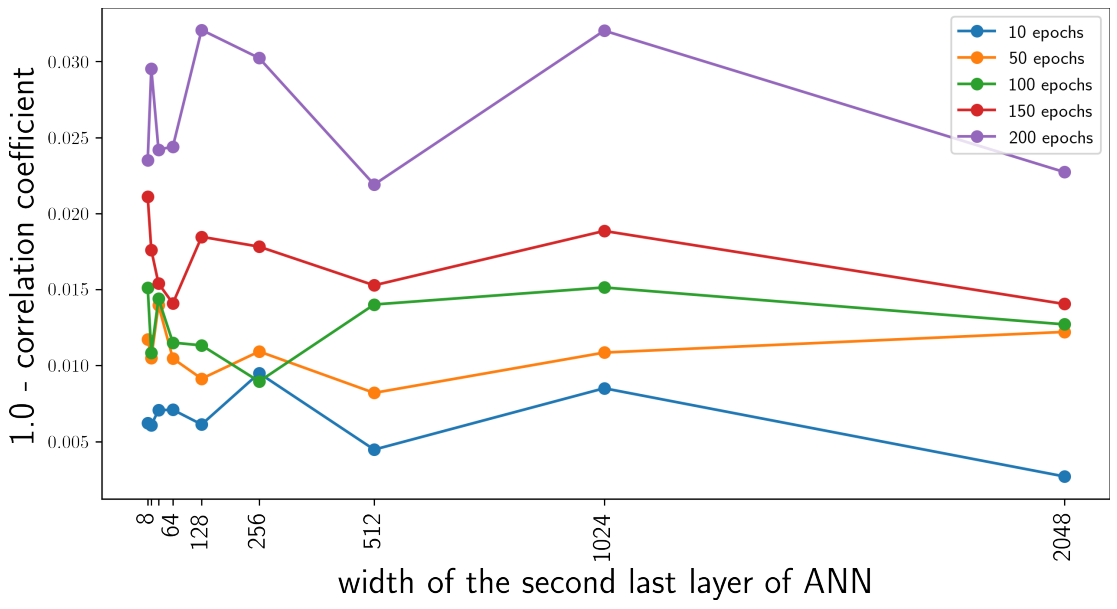

head 6

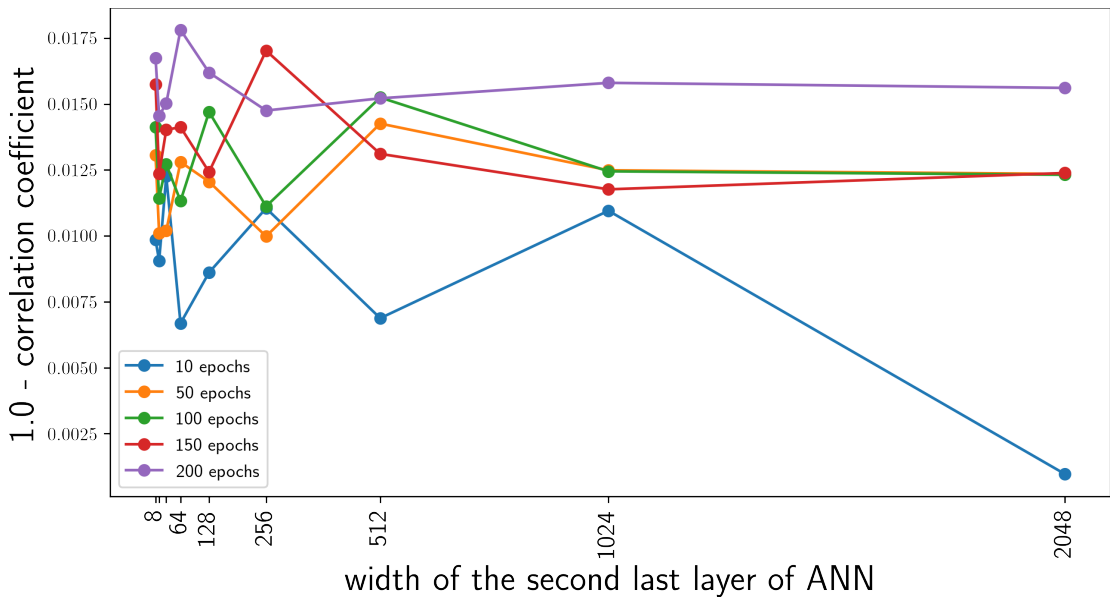

head 7

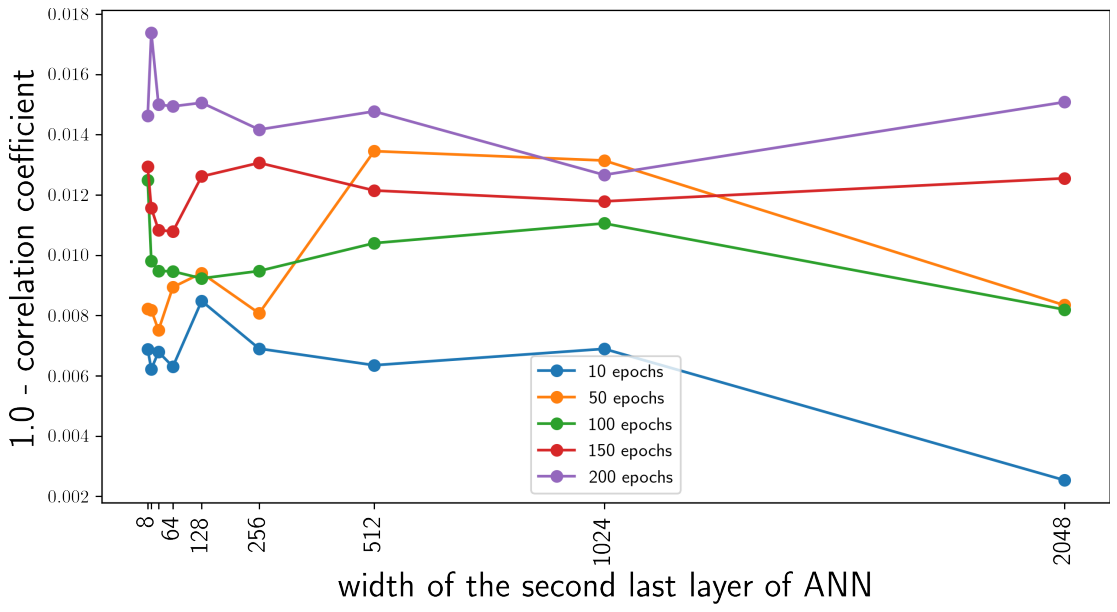

head 8

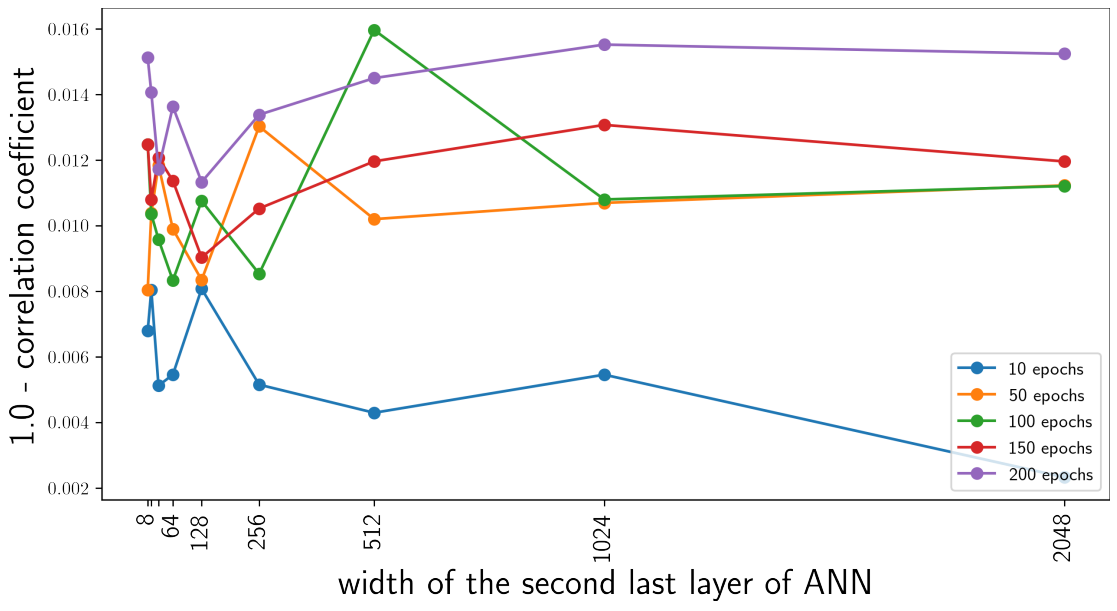

head 9

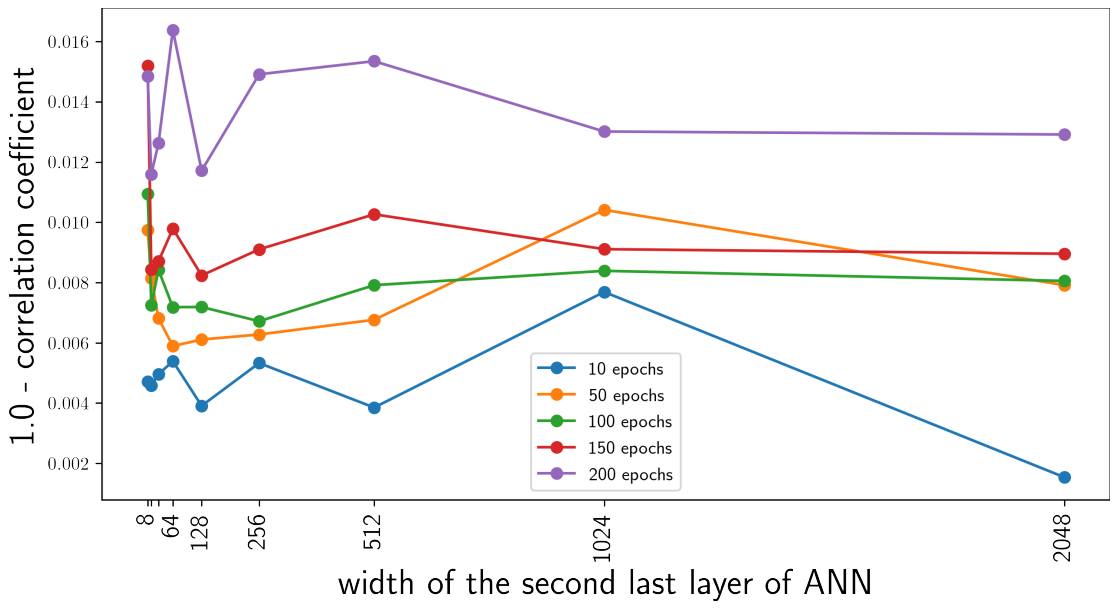

head 10

Fig. S58: Parameter analysis of Sec.S7.

| | Cifar10 [15] | MNIST [6] | Kather [12] | DogsWolves [30] |
|---|---|---|---|---|
| ANN accuracy | 95.43 | 99.56 | 96.80 | 80.50 |
| GPs accuracy | 92.26 | 99.41 | 93.60 | 78.75 |

TABLE S1: Accuracies of ANN classifiers versus the accuracies of the explainer GPs on four datasets.

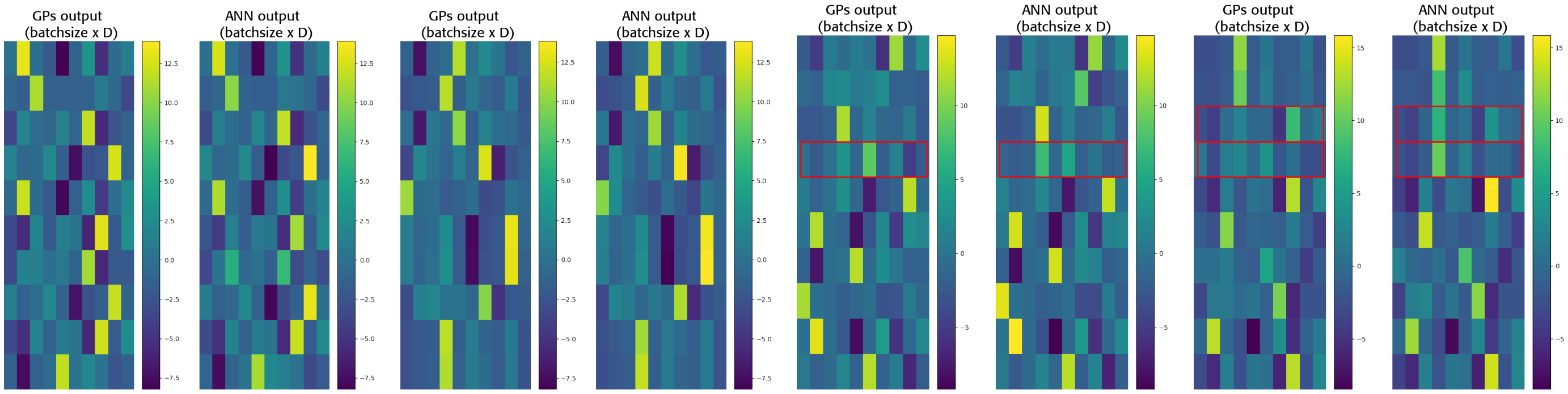

batch 1 batch 2 batch 3 batch 4

Fig. S59: Comparing GP and ANN outputs for four batches of Cifar10 dataset [33]. The red rectangles highlight the instnaces for which the predictions of GP and ANN (i.e. the class with maximum score) are different.

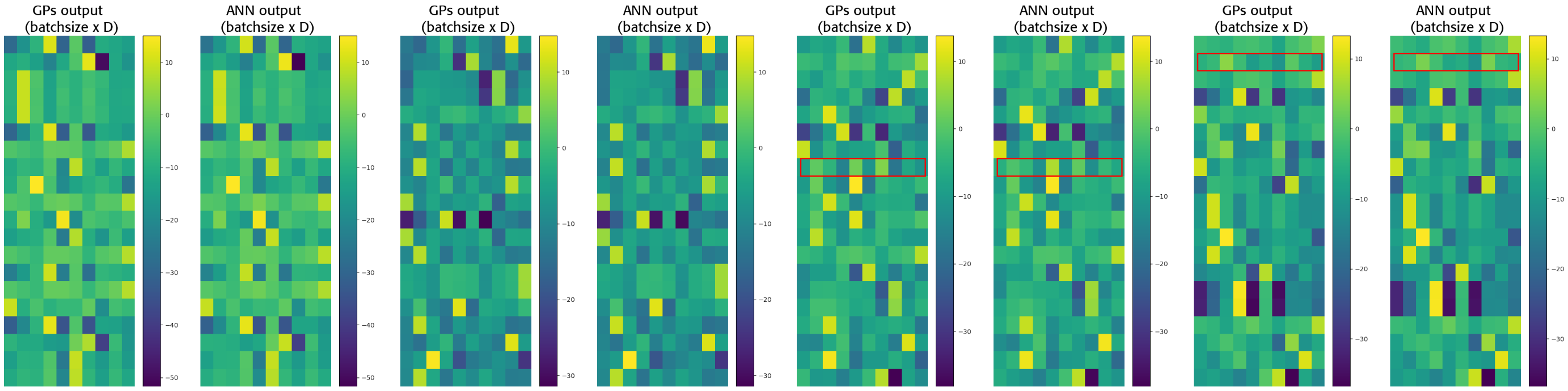

batch 1 batch 2 batch 3 batch 4

Fig. S60: Comparing GP and ANN outputs for four batches of MNIST dataset [32]. The red rectangles highlight the instnaces for which the predictions of GP and ANN (i.e. the class with maximum score) are different.

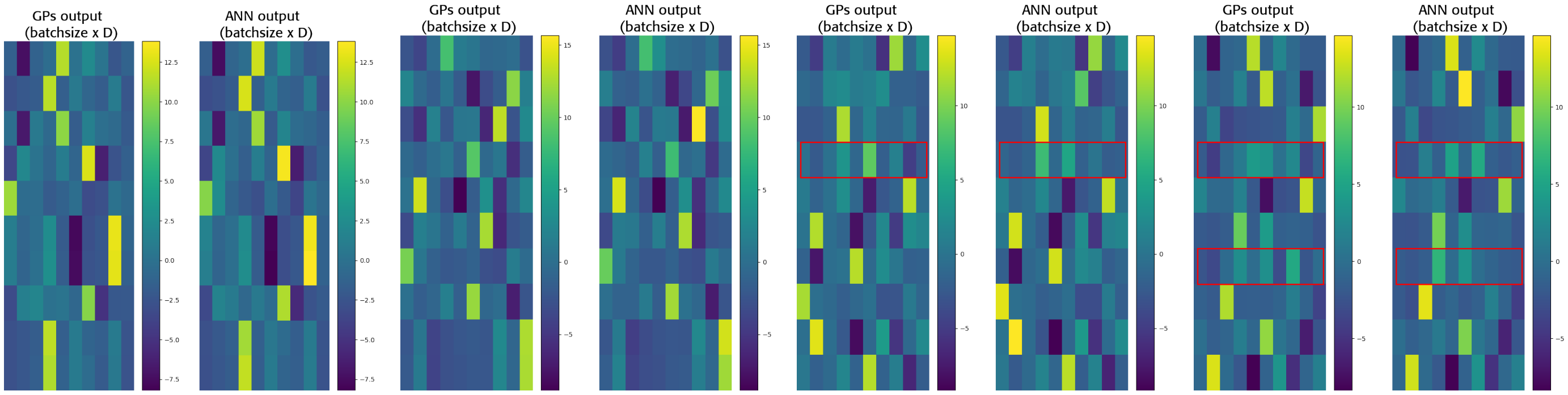

batch 1 batch 2 batch 3 batch 4

Fig. S61: Comparing GP and ANN outputs for four batches of Kather dataset [34]. The red rectangles highlight the instnaces for which the predictions of GP and ANN (i.e. the class with maximum score) are different.

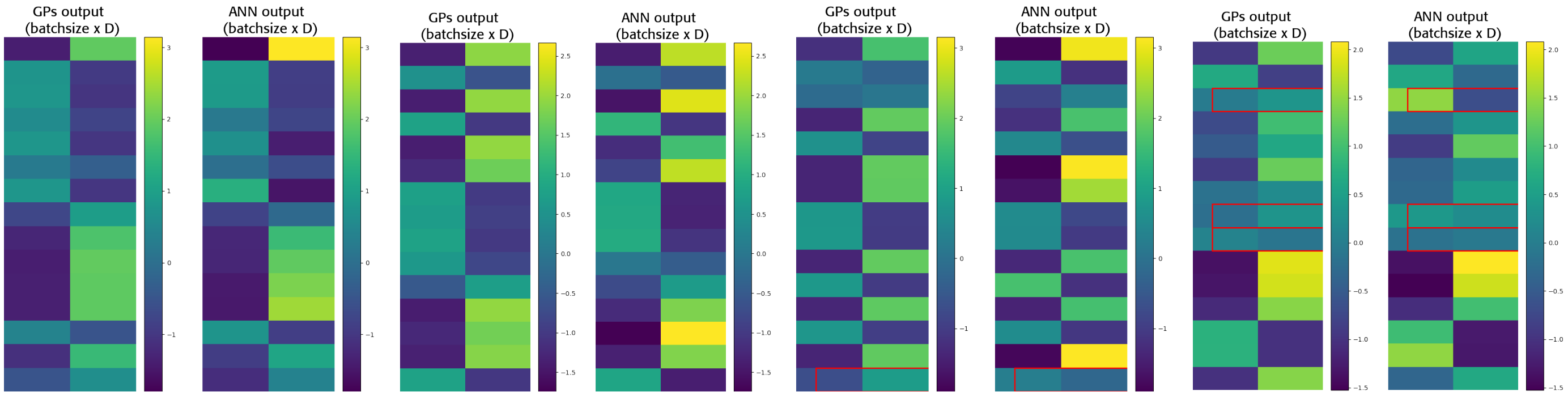

batch 1 batch 2 batch 3 batch 4

Fig. S62: Comparing GP and ANN outputs for four batches of DogsWolves dataset [35]. The red rectangles highlight the instnaces for which the predictions of GP and ANN (i.e. the class with maximum score) are different.

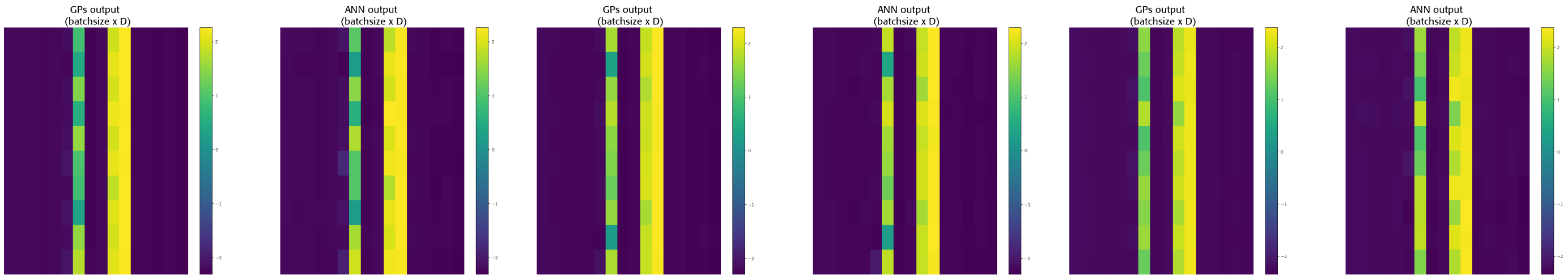

batch 1 batch 2 batch 3

Fig. S63: Comparing GP and ANN (attention submodule) outputs for 3 batches of Cifar10 dataset [15].

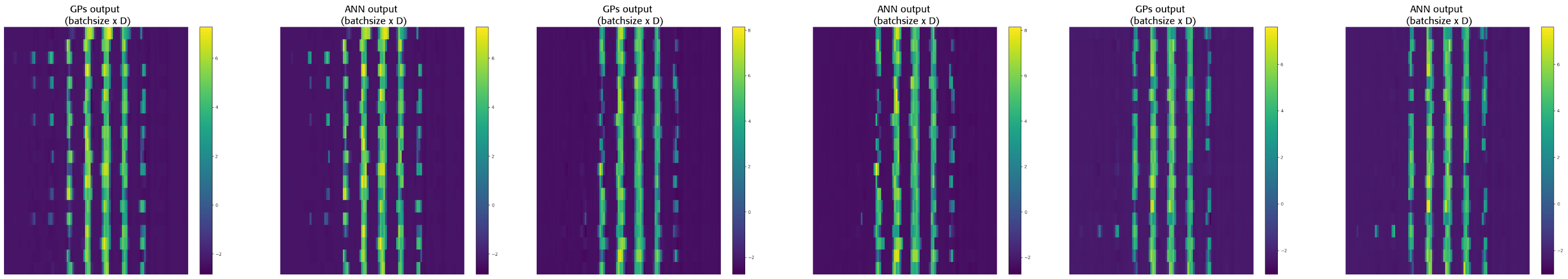

batch 1 batch 2 batch 3

Fig. S64: Comparing GP and ANN (attention submodule) outputs for 3 batches of MNIST dataset [6].

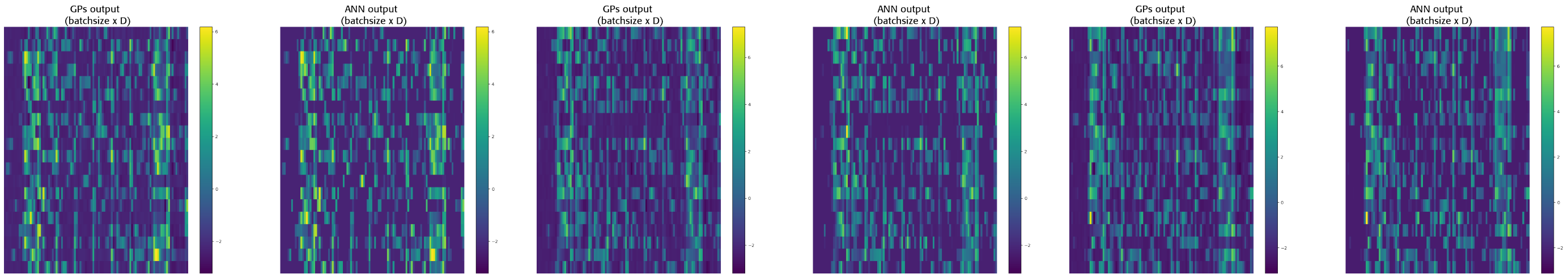

batch 1 batch 2 batch 3

Fig. S65: Comparing GP and ANN (attention submodule) outputs for 3 batches of Kather dataset [12].

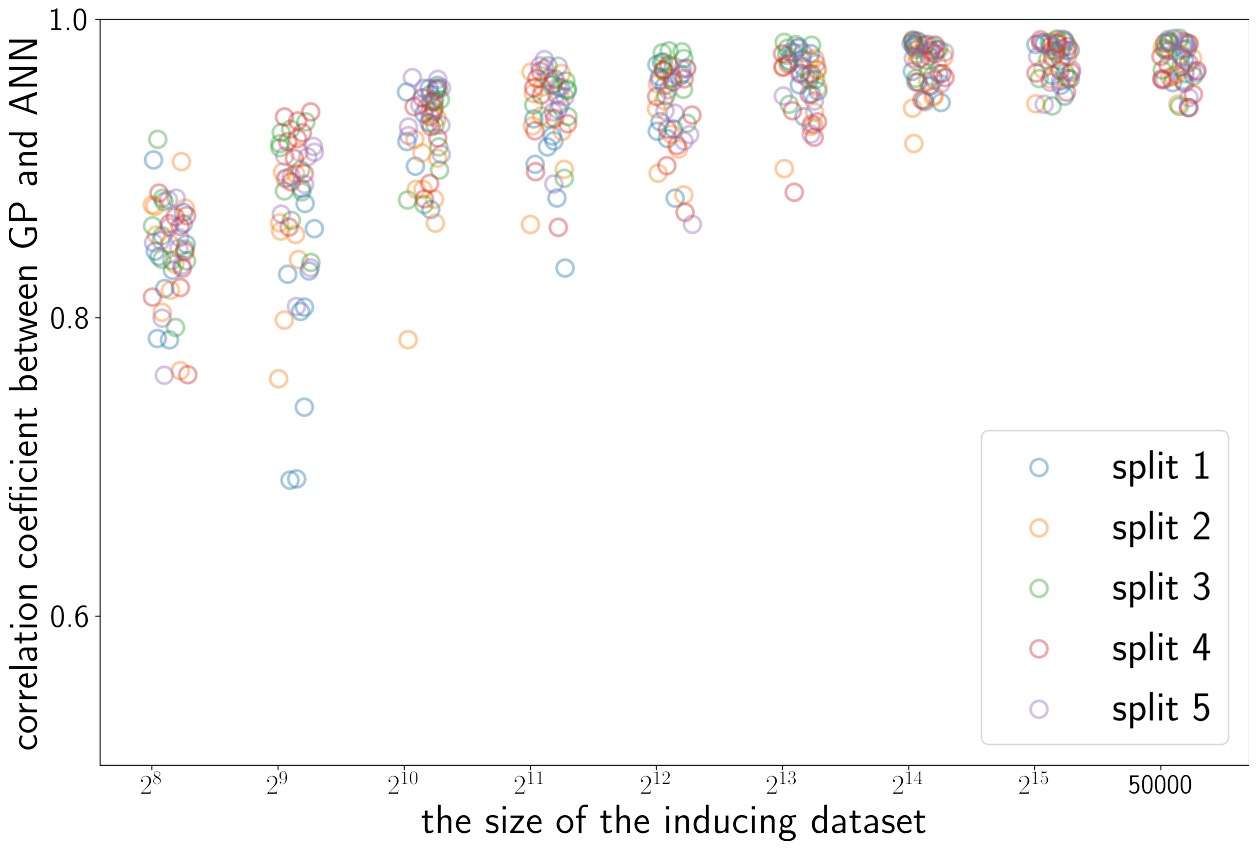

Fig. S66: Analyzing the effect of the size of inducing dataset.

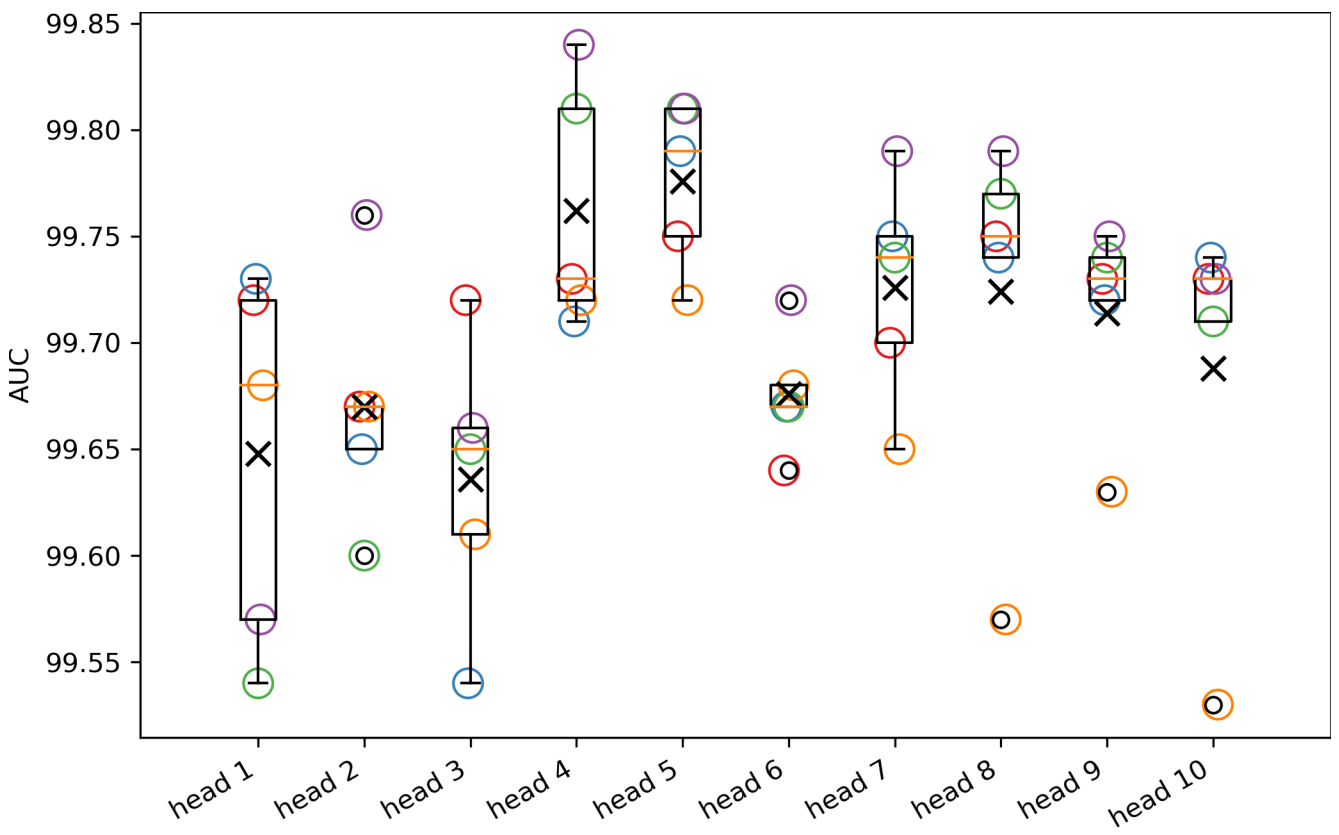

MNIST

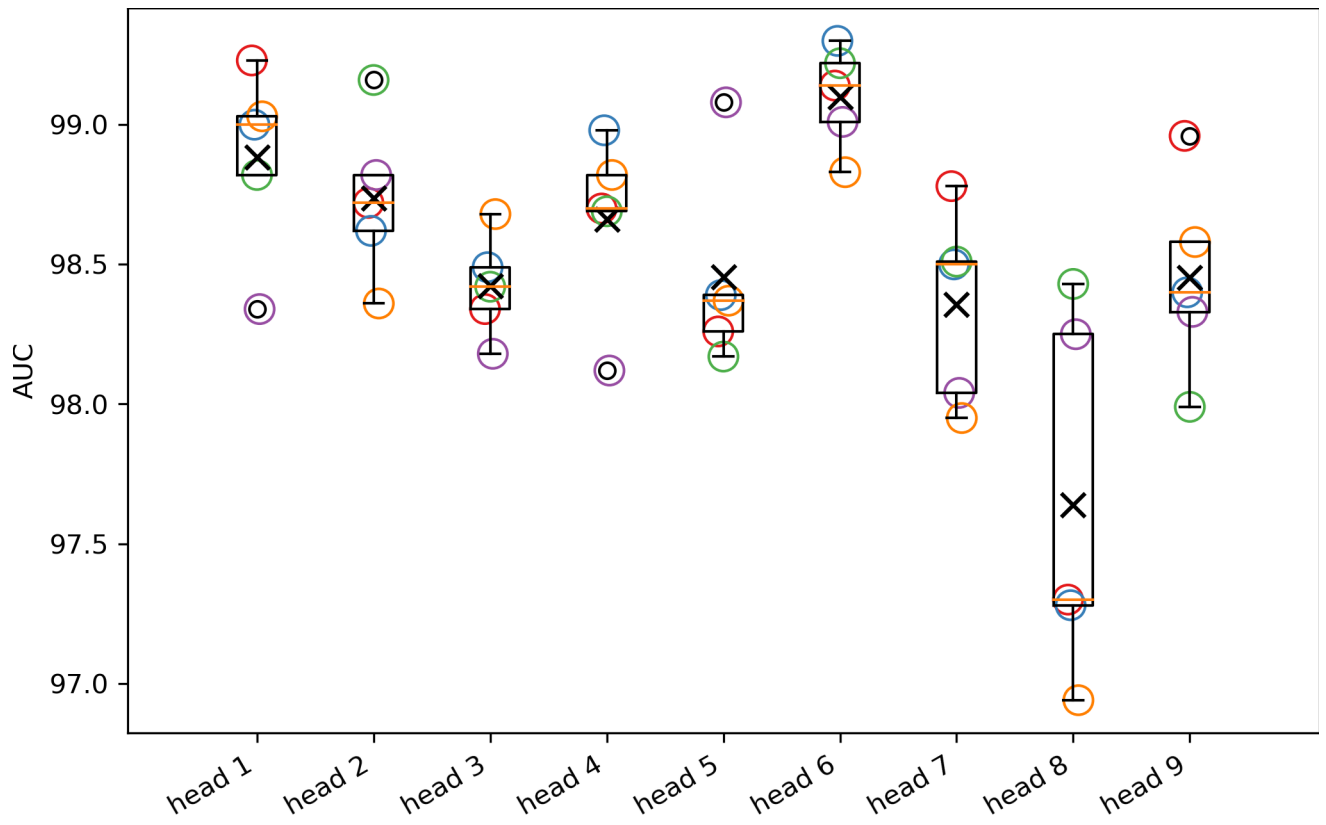

Kather

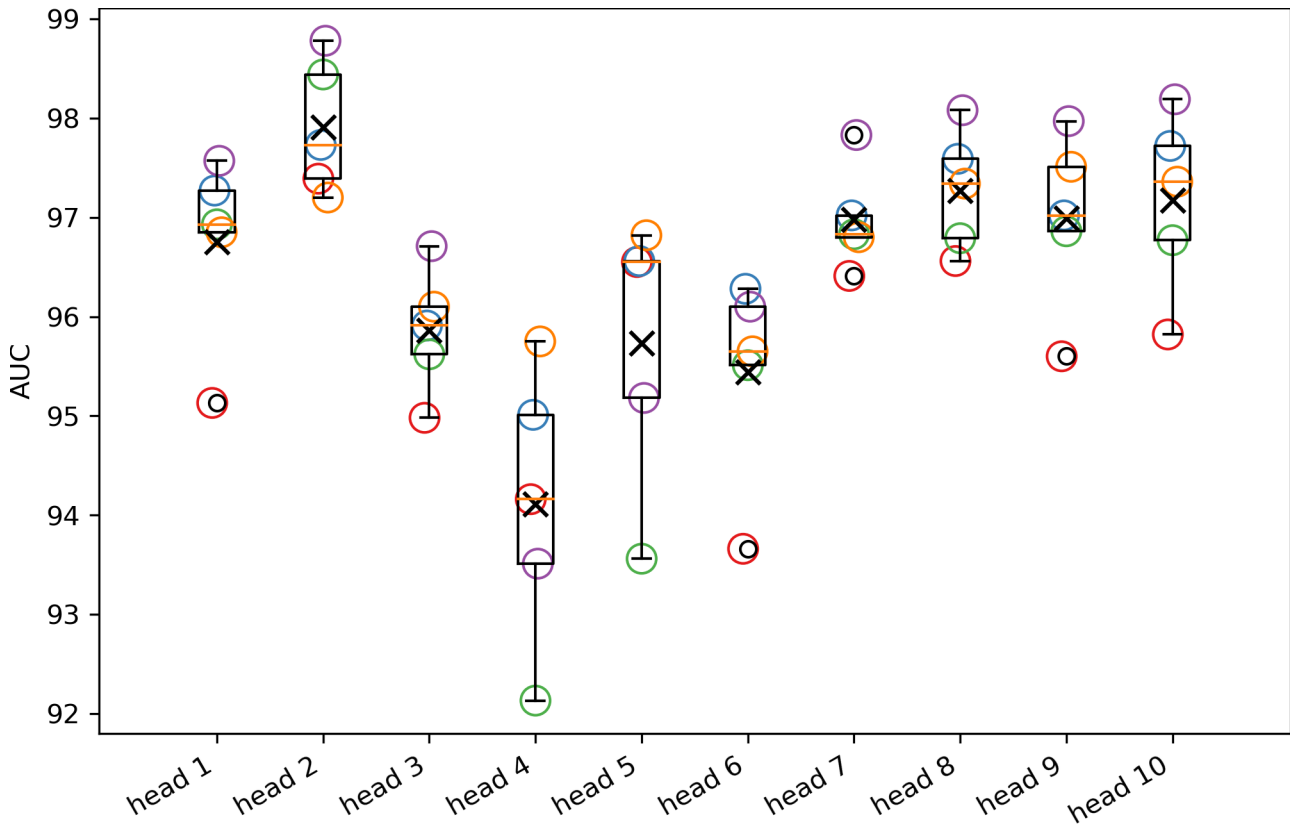

Cifar10

Fig. S67: Correlation coefficients for 5 splits.